\pdfoutput=1
\documentclass{article}
\PassOptionsToPackage{dvipsnames,table}{xcolor}
\usepackage{aibl_arxiv}

\usepackage{amsmath,amsfonts,bm}

\def\eqref#1{equation~\ref{#1}}

\def\1{\bm{1}}

\DeclareMathAlphabet{\mathsfit}{\encodingdefault}{\sfdefault}{m}{sl}
\SetMathAlphabet{\mathsfit}{bold}{\encodingdefault}{\sfdefault}{bx}{n}

\usepackage{hyperref}
\usepackage{url}
\usepackage{algorithm}
\usepackage{algpseudocode}

\usepackage{amsmath,amssymb}
\usepackage{graphicx}
\usepackage[dvipsnames,table]{xcolor}
\usepackage{colortbl}
\usepackage{booktabs}
\usepackage{multirow}
\usepackage{array}
\usepackage{makecell}
\usepackage{pifont}
\usepackage{titletoc}
\usepackage{needspace}
\graphicspath{{figures/}}

\newcommand{\cmark}{\ding{51}}
\newcommand{\xmark}{\ding{55}}
\newcommand{\yes}{\textcolor[HTML]{1E8449}{\cmark}}
\newcommand{\no}{\textcolor[HTML]{C62828}{\xmark}}
\newcommand{\venue}[1]{{\scriptsize\textcolor{gray}{#1}}}
\newcommand{\fitwidth}[1]{%
  \begingroup\setlength{\tabcolsep}{4pt}%
  \sbox0{#1}%
  \ifdim\wd0>\textwidth
    \resizebox{\textwidth}{!}{\usebox0}%
  \else\usebox0\fi
  \endgroup}
\newcommand{\fitnarrow}[1]{\resizebox{\ifdim\width>0.52\textwidth 0.52\textwidth\else \width\fi}{!}{#1}}

\newif\ifshowchanges
\showchangesfalse
\newcommand{\sh}[1]{{\leavevmode\ifshowchanges\color{blue}\fi#1}}
\begin{document}
\clubpenalty=10000
\widowpenalty=10000
\displaywidowpenalty=10000

\aiblmaketitle
  {Depth-to-RGB: Repurposing a Frozen Depth Estimator for Geometry-Guided Compositing}
  {Sanghyun Jo$^{1,2,\dagger}$, \ Chae Yeon Lim$^{1,2}$, \ Donghwan Lee$^{2}$, \ Sihyun Kim$^{2}$, \ Soo Ye Kim$^{3}$, \ Kyungsu Kim$^{2,\dagger}$}
  {$^{1}$OGQ \quad $^{2}$Seoul National University \quad $^{3}$Adobe Research\\[0.15em]
   {\small $^{\dagger}$Corresponding authors: \texttt{shjo.april@gmail.com}, \texttt{kyskim@snu.ac.kr}}}
  {Reference-based object compositing inserts or replaces an object using a background image, a reference image, and a 2D compositing mask. These inputs guide appearance and placement but leave the completed scene's geometry implicit, which can distort object structure or alter the surroundings. Our Depth-to-RGB (D2R) framework predicts composite depth for a scene not yet observed in the RGB inputs. It learns reference-conditioned corrections to a frozen depth estimator using encoder features of paired completed scenes as targets. The unchanged decoder maps the corrected representation to the intended scene's depth, which a separately trained renderer holds fixed during RGB synthesis. Under matched architecture and training, encoder-feature supervision reduces OOD Stage-1 AbsRel by 31.4\% relative to decoded-depth supervision. We also introduce AnyInsertion++ with paired in-distribution and category-disjoint splits to evaluate generalization beyond compositing training categories. The complete D2R system leads 12 open-source and 3 closed-source baselines in estimator-derived geometry and photometric quality on both paired splits. On category-disjoint data, D2R reduces AbsRel by 43.7\% and improves PSNR by 2.4\,dB over the matched RGB baseline. Across three unpaired benchmarks, D2R leads both identity metrics and reduces mean CLIP reference cosine distance by 55\% relative to the strongest baseline. Code and AnyInsertion++ are publicly available at the links below.}
  {\aiblresource{Project Page}{https://shjo-april.github.io/Depth2RGB/}
   \aiblresource{Github}{https://github.com/shjo-april/Depth2RGB}
   \aiblresource{Lab}{https://aibl.snu.ac.kr}}

\section{Introduction}
\label{sec:intro}

\begin{figure}[t]
\centering
\includegraphics[width=\textwidth]{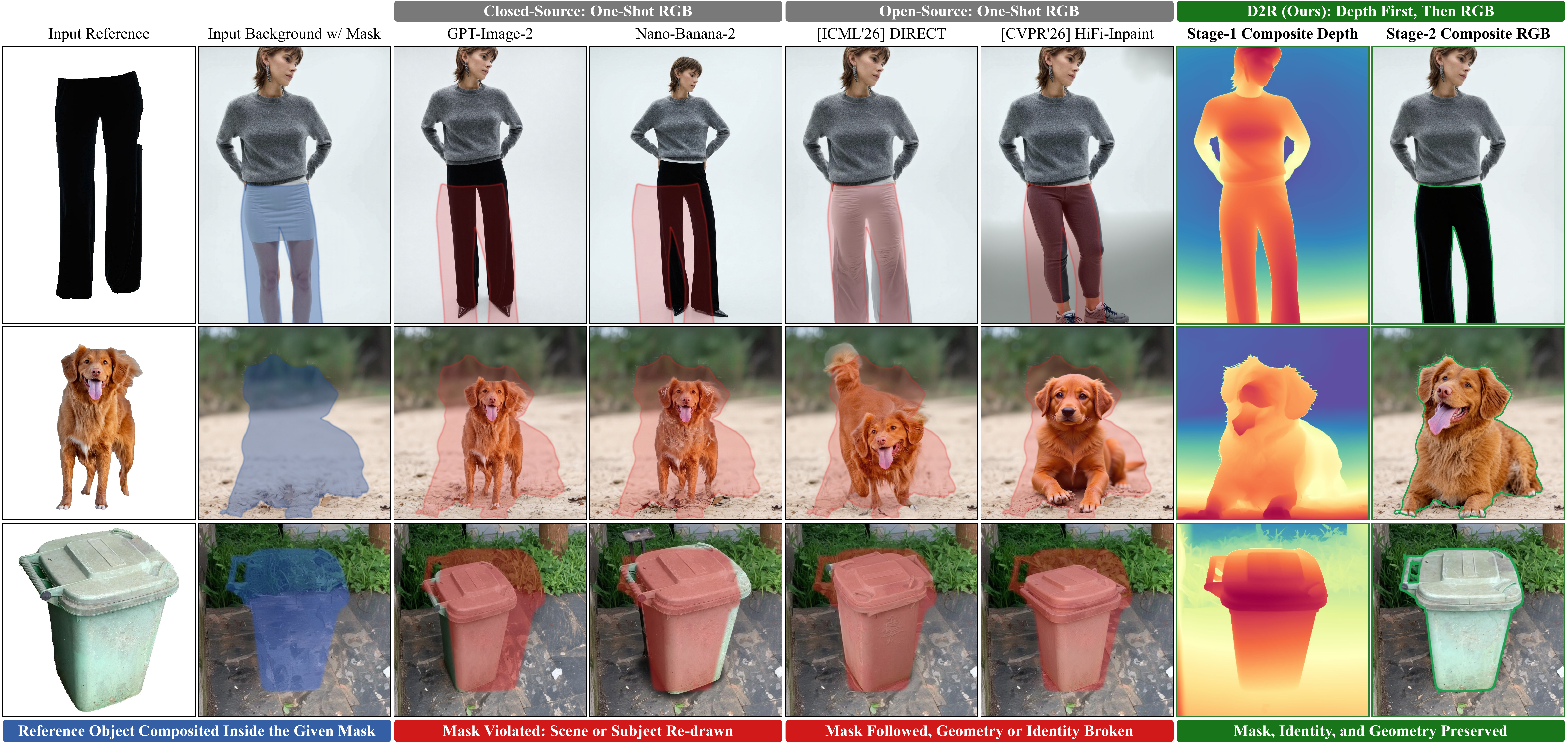}
\vspace{-13pt}
\caption{\textbf{Predicting Composite Depth Before RGB Synthesis.} Depth-to-RGB (D2R; Ours) predicts composite depth (seventh column) and holds it fixed while generating RGB (eighth). D2R preserves reference appearance while matching the target shape and pose; the open- and closed-source baselines instead miss the target geometry, alter object identity, or modify surrounding content.}
\label{fig:teaser}
\end{figure}

Object compositing must preserve a reference object's identity while adapting its shape and placement to a new scene. The background provides scene context, and a 2D mask indicates the target region without specifying the intended composite depth. Figure~\ref{fig:teaser} illustrates the challenge: transferring reference appearance can produce distorted object structure, incorrect placement, or changes to the surrounding scene. These errors motivate a geometric condition for the completed scene, guiding appearance generation with an explicit description of the object and its surroundings.

Most reference-guided object compositing methods infer geometry implicitly during RGB denoising. Some approaches with explicit geometric conditions use depth estimated from an input image \citep{chen2024mimicbrush,parihar2025zeroshot}, assembled foreground and background depth maps \citep{li2024bifrost}, or an RGB proxy rendered from a user-posed 3D object \citep{gong2026direct}. Depth estimated from an input image describes its observed geometry, which does not determine the reference object's depth and shape in the completed composite. Assembling maps relies on rules for placing and rescaling reference depth within the background, while rendering a proxy requires the object's target pose to be specified before RGB generation. Our goal is to predict post-composite depth from the background image, reference image, and 2D compositing mask, jointly estimating the geometry of the reference object and its surroundings.

We call this field \emph{composite depth}: the depth of the entire scene after compositing. Background depth omits the intended object, while reference depth describes it in another scene and viewpoint. Neither map directly describes the object's shape and placement within the target scene. Predicting this field requires combining reference appearance with scene context. We formulate this task as \emph{endogenous geometry provisioning}: inferring this geometric condition from the given inputs.

A pretrained depth estimator maps an observed image's features to depth. \emph{Depth-to-RGB} (D2R) freezes its encoder and decoder while learning reference-conditioned residual corrections to the background representation. Training aligns the resulting representation with encoder features extracted from paired completed scenes, enabling the unchanged decoder to predict composite depth. This geometric condition remains fixed throughout RGB synthesis.

To evaluate geometry and category generalization, we curate AnyInsertion++ from AnyInsertion~\citep{song2026insert}, expanding paired evaluation from 38 to 426 cases with a category-disjoint split. Paired evaluation compares estimated depth and RGB composites with paired targets. Additionally, we assess reference identity on TF-ICON~\citep{lu2023tficon}, AIComposer~\citep{li2025aicomposer}, and ComplexCompo~\citep{lu2026shine}, which collectively cover photographic, cartoon, and artistic backgrounds (Tab.~\ref{tab:unpaired}). These unpaired benchmarks test whether reference appearance is retained across different visual styles beyond the paired evaluation setting. Our contributions are:

{\setlength{\topsep}{2pt}\setlength{\partopsep}{0pt}
\begin{itemize}
\setlength{\itemsep}{1pt}\setlength{\parsep}{0pt}
\item \textbf{Problem.} We address the missing geometric condition in reference-based object compositing: predicting the completed scene's relative depth from a background image, reference image, and 2D compositing mask.
\item \textbf{Solution.} We propose \emph{Depth-to-RGB} (D2R). First, we learn reference-conditioned corrections to a frozen depth estimator using encoder features of paired completed scenes as targets. Under matched architecture and training, encoder-feature supervision reduces OOD Stage-1 AbsRel by 31.4\% relative to decoded-depth supervision. Second, we train an RGB renderer to use the resulting depth as a fixed geometric condition.
\item \textbf{Results.} The complete D2R system reduces category-disjoint AbsRel by 43.7\% and improves PSNR by 2.4\,dB over the matched RGB baseline. Across three unpaired benchmarks, D2R leads the 15 evaluated baselines on both identity metrics, reducing mean CLIP reference cosine distance by 55\% relative to the strongest baseline.
\end{itemize}}

\section{Related Work}
\label{sec:related}

\noindent\textbf{Implicit Geometry in RGB Compositing.} Paint-by-Example \citep{yang2023paint} and AnyDoor \citep{chen2024anydoor} transfer reference appearance; Personalize-Anything~\citep{feng2026personalize}, SHINE~\citep{lu2026shine}, and HiFi-Inpaint~\citep{liu2026hifiinpaint} improve identity through token substitution, sampling-time control, and high-frequency detail cues (\emph{e.g.}, edges), respectively. Such 2D structural cues can preserve local shape but do not specify scene-wide composite depth. D2R predicts this geometric condition before RGB generation. Insert-Anything \citep{song2026insert} provides our direct baseline because its in-context canvas and LoRA \citep{hu2022lora} support a matched evaluation of adding depth prediction and supervision.

\begin{table}[t]
\centering
\caption{\textbf{Geometry for Reference-Based Object Compositing.} Standard inputs are background and reference images plus a 2D compositing mask. Fixed geometry denotes depth maps or RGB proxies supplied before RGB generation.}
\label{tab:related}
{\setlength{\tabcolsep}{3pt}
\begin{tabular}{@{}l l c c c@{}}
\toprule
Method & Geometry Condition & \makecell{Standard\\Inputs Only} & \makecell{Predicts\\Composite\\Depth} & \makecell{Fixed Geometry\\Before RGB} \\
\midrule
Insert-Anything \citep{song2026insert} \venue{AAAI'26} & Implicit in RGB & \yes & \no & \no \\
Personalize-Anything \citep{feng2026personalize} \venue{AAAI'26} & Implicit in RGB & \yes & \no & \no \\
SHINE \citep{lu2026shine} \venue{ICLR'26} & Implicit in RGB & \yes & \no & \no \\
HiFi-Inpaint \citep{liu2026hifiinpaint} \venue{CVPR'26} & Implicit in RGB & \yes & \no & \no \\
\midrule
MimicBrush \citep{chen2024mimicbrush} \venue{NeurIPS'24} & Input Background Depth & \yes & \no & \yes \\
BIFR\"OST \citep{li2024bifrost} \venue{NeurIPS'24} & Rule-Assembled Depth & \no & \no & \yes \\
DIRECT \citep{gong2026direct} \venue{ICML'26} & Proxy-Rendered RGB & \no & \no & \yes \\
\midrule
\rowcolor{ForestGreen!12}
\textbf{D2R (Ours)} & Predicted Composite Depth & \yes & \yes & \yes \\
\bottomrule
\end{tabular}}
\end{table}

\noindent\textbf{Geometry Before RGB Synthesis.} \sh{MimicBrush \citep{chen2024mimicbrush} optionally uses input-background depth. BIFR\"OST~\citep{li2024bifrost} assembles reference and scene depth in its instruction branch; we evaluate its mask-based branch. DIRECT~\citep{gong2026direct} uses a posed 3D proxy. Person-In-Situ~\citep{masuda2026personinsitu} predicts completed-scene depth for human insertion using rendered body depth; 3DIS~\citep{zhou20253dis} generates scene depth from instance layouts before RGB. D2R's distinction is reference-conditioned prediction of completed-scene features within a frozen depth estimator, using standard compositing inputs (Tab.~\ref{tab:related}). Appendix~\ref{sec:appx_related} compares the inputs, prediction targets, and learned components.}

\noindent\textbf{Geometry During Sampling.} TRACE~\citep{jo2026trace} finds instance structure in intermediate diffusion states, motivating our focus on when geometry becomes available. FreeInpaint \citep{gong2026freeinpaint} optimizes noise and intermediate latents; closed-loop control \citep{zhao2026openloop} adjusts conditioning using feedback from generated outputs. Both control methods adjust RGB generation at inference, whereas D2R predicts composite depth once and holds it fixed throughout RGB sampling. Section~\ref{sec:ablation} compares D2R with compositing-specific optimization and joint-denoising baselines.

\noindent\textbf{Encoder Adaptation and RGB Conditioning.} ViT-Adapter~\citep{chen2023vitadapter} and ViT-CoMer~\citep{xia2024vitcomer} augment encoders for dense prediction; auxiliary branches and canvas patches condition RGB generators~\citep{zhang2023adding,mou2024t2iadapter,tan2025ominicontrol}. VGGT-World~\citep{sun2026vggtworld} predicts future geometry features from video observations and decodes them through frozen geometry components. D2R instead predicts the representation of a requested static composite from a background, reference, and compositing mask. Appendix~\ref{sec:appx_representation_comparison} compares these targets and the connections to joint RGB-depth modeling.

\section{Method}
\label{sec:method}

\begin{figure}[t]
\centering
\includegraphics[width=\textwidth]{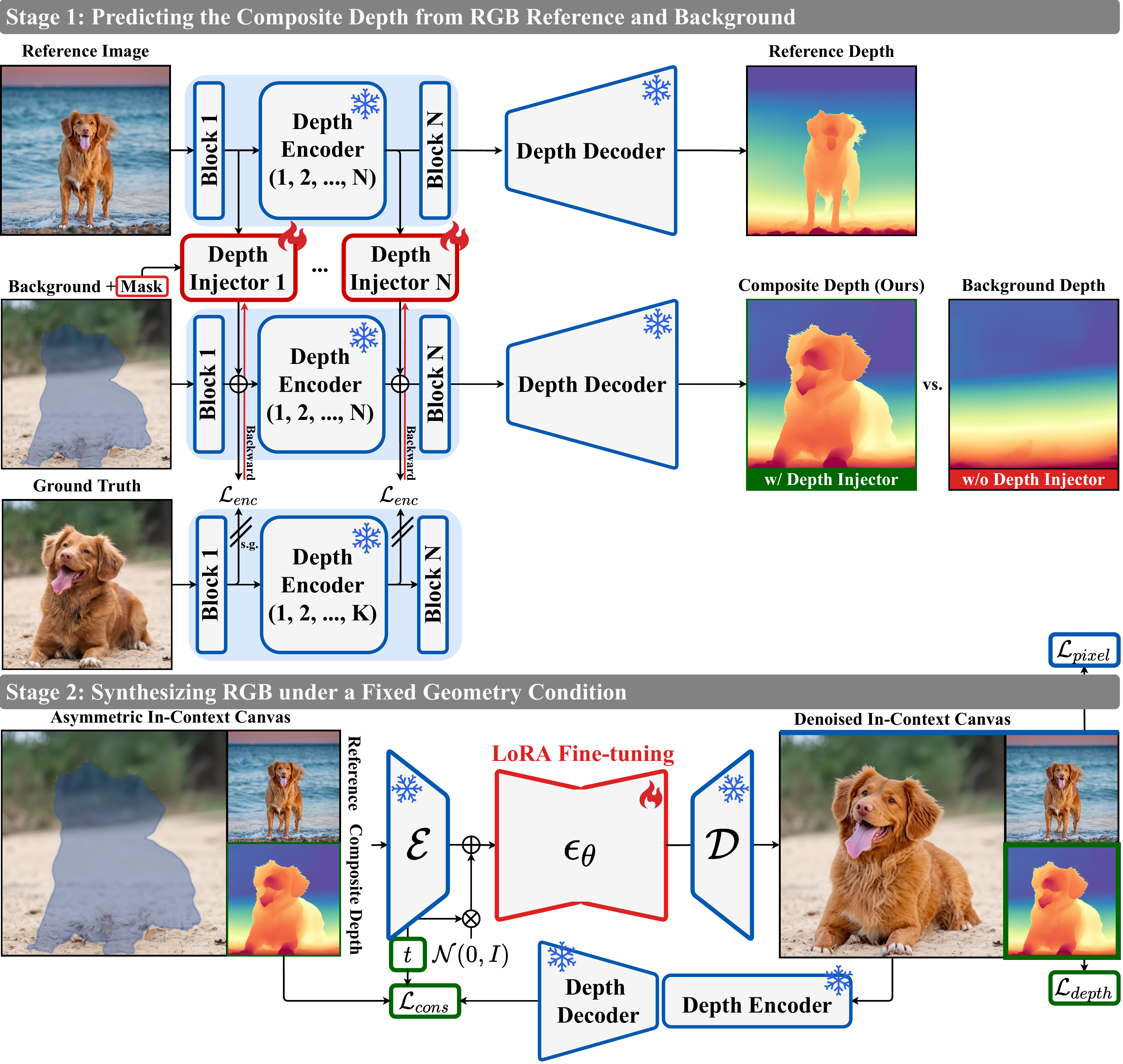}
\vspace{-13pt}
\caption{\textbf{Overview of Depth-to-RGB (D2R).} Stage~1 (Sec.~\ref{sec:stage1}) predicts composite depth through reference-conditioned corrections to background features in a frozen depth estimator. Stage~2 (Sec.~\ref{sec:stage2}) synthesizes RGB with this depth held fixed throughout sampling. \sh{Training corrupts a separate canvas containing paired RGB, the reference, and target depth $D$. Depth denoising and RGB-depth consistency always target $D$, preventing errors in $D_{\mathrm{in}}$ from becoming supervision.}}
\label{fig:overview}
\end{figure}

Let $X=(I_{\mathrm{bg}},I_{\mathrm{ref}},M)$ contain the background, reference, and 2D compositing mask, with $M=1$ on the target region. Reference cropping uses its dataset-provided object mask (App.~\ref{sec:appx_canvas}). For a paired composite $I$, a frozen estimator $F$ supplies $D=F(I)$, termed \emph{ground-truth (GT) composite depth}. This target is estimator-derived relative depth, not sensor ground truth. Stage~1 predicts $\hat D=\Phi_\phi(X)$ with trainable injector parameters $\phi$. D2R uses the following factorization:
\begin{equation}
p(I\mid X)=\int p(I\mid D,X)\,p(D\mid X)\,\mathrm{d}D,
\qquad \hat{D}=\Phi_\phi(X).
\label{eq:egp}
\end{equation}
D2R approximates $p(D\mid X)$ with one prediction $\hat D$ and generates RGB from $p_\theta(I\mid\hat D,X)$, where $\theta$ denotes trainable LoRA parameters (Fig.~\ref{fig:overview}). Appendix~\ref{sec:appx_notation} defines the sampling notation.

\subsection{Stage 1: Predicting Composite Depth from Reference and Background}
\label{sec:stage1}

Stage~1 uses a paired composite's encoder features as targets because the pretrained decoder already maps them to composite depth. We freeze the encoder $E$ and decoder $G$ of $F=G\circ E$ and train mask-conditioned injectors. At encoder blocks $\mathcal K$ read by $G$, each injector attends to background and reference features and adds a residual $\delta^\ell$:
\begin{equation}
\begin{split}
\tilde{h}^{\ell}_{\mathrm{bg}}&=B_\ell(\tilde{h}^{\ell-1}_{\mathrm{bg}})
+\mathbf{1}[\ell\in\mathcal{K}]\,\delta^\ell,
\qquad \hat{D}=G\bigl(\{\tilde{h}^{\ell}_{\mathrm{bg}}\}_{\ell\in\mathcal{K}}\bigr),\\
\mathcal{L}_{\mathrm{enc}}&=\frac{1}{|\mathcal{K}|}\sum_{\ell\in\mathcal{K}}
\operatorname{mean}_{b}\!\left[\frac{\operatorname{mean}\bigl|\tilde{h}^{\ell}_{\mathrm{bg},b}-\mathrm{sg}(h^{\ell}_{\mathrm{gt},b})\bigr|}{\mathrm{sg}(\sigma_b^\ell)+\varepsilon_{\mathrm{enc}}}\right].
\end{split}
\label{eq:stage1}
\end{equation}
Here $B_\ell$ is a frozen encoder block and $\tilde h^\ell_{\mathrm{bg}}$ denotes corrected features initialized from masked-background tokens at $\ell=0$. The indicator selects $\mathcal K$, and $\mathrm{sg}$ stops gradients. A training-only pass through $E$ supplies targets $h^\ell_{\mathrm{gt}}$. For sample $b$, $\sigma_b^\ell$ is their standard deviation over channels and spatial tokens; the inner error mean uses the same dimensions and $\varepsilon_{\mathrm{enc}}=10^{-6}$ stabilizes division. Appendix~\ref{sec:appx_supervision_comparison} compares feature targets with decoded-depth supervision and their combination.

Corrections span the full token grid and propagate through subsequent frozen blocks, allowing object and background geometry to be inferred jointly. Zero-initialized output projections initially recover the frozen estimator on the same masked-background input. Each depth backbone uses the same injector design with adapted attachment points and separately trained parameters. Appendix~\ref{sec:appx_stage1} specifies attention, gates, and architecture-specific connections; App.~\ref{sec:appx_s1} gives training details.

\subsection{Stage 2: Synthesizing RGB with Fixed Composite Depth}
\label{sec:stage2}

Following canvas-based conditioning \citep{song2026insert,tan2025ominicontrol}, Stage~2 arranges the masked background, reference image, and composite depth as composite, reference, and depth patches on one input canvas. At inference, RGB is generated only in the masked region of the composite patch; the rest of the canvas remains fixed as conditioning input (App.~\ref{sec:appx_canvas}).

\noindent\textbf{Training and Auxiliary Supervision.} Auxiliary depth prediction trains the shared denoiser to represent composite geometry. A frozen autoencoder encodes the input canvas and a supervision canvas whose three patches contain the paired composite, reference image, and target depth $D$; only denoiser LoRA \citep{hu2022lora} is trained. The input depth patch receives $\hat D$ or $D$ with equal probability, exposing the renderer to prediction errors while retaining the same supervision. We add noise to the full supervision canvas during training and minimize $\mathcal{L}_{\mathrm{total}}=\mathcal{L}_{\mathrm{pixel}}+\mathcal{L}_{\mathrm{depth}}+\mathcal{L}_{\mathrm{cons}}$. The RGB loss supervises the composite and reference patches; the depth loss supervises the depth patch. At inference, only the masked region of the composite patch is generated, while the supplied depth patch remains fixed. An accurate depth patch does not ensure that the generated RGB image matches the target geometry. We therefore apply the frozen estimator $F$ to the one-step clean RGB estimate $\hat{I}_{\mathrm{comp}}$ and compare its depth with the paired target $D$:
\begin{equation}
\mathcal{L}_{\mathrm{cons}}=(1-t)\,\bigl\|A(F(\hat{I}_{\mathrm{comp}}))-\mathrm{sg}(A(D))\bigr\|_1.
\label{eq:cons}
\end{equation}
$A$ subtracts each map's median and divides by its mean absolute deviation from that median, using statistics over the entire composite patch. The loss averages over valid pixels and always targets $D$, including when the input is $\hat D$. With $t=1$ denoting noise and $t=0$ clean data, $1-t$ downweights noisy early estimates. Gradients flow through the frozen RGB decoder and depth estimator $F$ to update only the LoRA parameters. Appendix~\ref{sec:appx_stage2} defines depth normalization and RGB decoding; App.~\ref{sec:appx_notation} gives one-step clean-latent estimates for rectified flow and variance-preserving diffusion.

\noindent\textbf{Inference.} Stage~1 predicts $\hat D$ once, then Stage~2 synthesizes the masked RGB region while restoring the depth condition after every update. The default pipeline takes 22.40\,s per image on RTX PRO 6000 and preserves the background image's resolution (App.~\ref{sec:appx_cost}).

\section{Experiments}
\label{sec:exp}

\noindent\textbf{Datasets.} AnyInsertion++ contains 52.9K curated training samples from AnyInsertion \citep{song2026insert} and 426 paired evaluation cases: 287 in-distribution (ID) and 139 from categories absent from compositing training (OOD). Thumbnail-hash and confirmed perceptual-hash checks find no train/test image matches. For reference-identity evaluation without paired targets, TF-ICON \citep{lu2023tficon}, AIComposer \citep{li2025aicomposer}, and ComplexCompo \citep{lu2026shine} provide 332, 272, and 300 cases. Target regions use object masks in AnyInsertion++ and coarse box masks in the unpaired benchmarks. Appendix~\ref{sec:appx_bench} details curation, overlap checks, and evaluation protocols.

\noindent\textbf{Implementation Details.} Stage~1 uses a width-640 injector with frozen Depth Anything V2 \mbox{ViT-L}~\citep{yang2024depthv2}. Stage~2 adapts FLUX.1-Fill with Redux\footnote{Official model cards: \href{https://huggingface.co/black-forest-labs/FLUX.1-Fill-dev}{FLUX.1-Fill} and \href{https://huggingface.co/black-forest-labs/FLUX.1-Redux-dev}{FLUX.1-Redux}.}, following the LoRA and sampling setup of Insert-Anything~\citep{song2026insert}. All trained variants use final checkpoints after 50K iterations for Stage~1 and 100K for Stage~2. Insert-Anything$^*$ is retrained on the same AnyInsertion++ data. Other baselines use released checkpoints or public interfaces. Settings and seed checks are in App.~\ref{sec:appx_render_training}; paired uncertainty and scorer robustness are in App.~\ref{sec:appx_metrics}.

\noindent\textbf{Metrics.} Full-frame AbsRel, $\delta_1$, and RMSE compare depths estimated from GT and generated RGB after background-based scale and shift alignment. PSNR and LPIPS~\citep{zhang2018unreasonable} measure photometric quality against GT composites. CLIP-I and DINO-I use IsoCLIP~\citep{magistri2026isoclip} and DINOv3 \mbox{ViT-L}~\citep{simeoni2025dinov3}, respectively, to measure cosine similarity between generated object crops and GT crops (paired) or reference crops (unpaired; App.~\ref{sec:appx_identity_metrics}).

\begin{table}[t]
\centering
\caption{In-Distribution Results on AnyInsertion++.}
\label{tab:main_id}
\fitwidth{\setlength{\tabcolsep}{3pt}%
\begin{tabular}{llccccccc}
\toprule
\multirow{2}{*}{Method} & \multirow{2}{*}{Backbone} & \multicolumn{3}{c}{Geometric} & \multicolumn{2}{c}{Photometric} & \multicolumn{2}{c}{Identity} \\
\cmidrule(lr){3-5}\cmidrule(lr){6-7}\cmidrule(lr){8-9}
& & AbsRel$\downarrow$ & $\delta_1\uparrow$ & RMSE$\downarrow$ & PSNR$\uparrow$ & LPIPS$\downarrow$ & CLIP-I$\uparrow$ & DINO-I$\uparrow$ \\
\midrule
Paint-by-Example \citep{yang2023paint} \venue{CVPR'23} & SD1.5 & 0.112 & 0.885 & 0.085 & 17.7 & 0.305 & 0.682 & 0.427 \\
TF-ICON \citep{lu2023tficon} \venue{ICCV'23} & SD2.1 & 0.205 & 0.779 & 0.118 & 17.2 & 0.414 & 0.545 & 0.262 \\
AnyDoor \citep{chen2024anydoor} \venue{CVPR'24} & SD2.1 & 0.173 & 0.841 & 0.091 & 17.5 & 0.370 & 0.770 & 0.634 \\
MimicBrush \citep{chen2024mimicbrush} \venue{NeurIPS'24} & SD1.5 & 0.066 & 0.950 & 0.048 & 19.5 & 0.218 & 0.771 & 0.623 \\
BIFR\"OST \citep{li2024bifrost} \venue{NeurIPS'24} & SD2.1 & 0.099 & 0.635 & 0.056 & 19.4 & 0.314 & 0.764 & 0.649 \\
DreamFuse \citep{huang2025dreamfuse} \venue{ICCV'25} & FLUX.1 & 0.127 & 0.887 & 0.077 & 17.8 & 0.334 & 0.738 & 0.612 \\
UniCombine \citep{wang2025unicombine} \venue{ICCV'25} & FLUX.1 & 0.073 & 0.950 & 0.046 & 17.6 & 0.289 & 0.754 & 0.654 \\
Personalize-Anything \citep{feng2026personalize} \venue{AAAI'26} & FLUX.1 & 0.177 & 0.835 & 0.104 & 17.5 & 0.459 & 0.706 & 0.554 \\
Insert-Anything$^{*}$ \citep{song2026insert} \venue{AAAI'26} & FLUX.1 & 0.062 & 0.963 & 0.036 & 21.2 & 0.213 & 0.887 & 0.909 \\
SHINE \citep{lu2026shine} \venue{ICLR'26} & FLUX.1 & 0.162 & 0.840 & 0.092 & 17.2 & 0.331 & 0.652 & 0.493 \\
HiFi-Inpaint \citep{liu2026hifiinpaint} \venue{CVPR'26} & FLUX.1 & 0.104 & 0.905 & 0.076 & 19.0 & 0.262 & 0.757 & 0.661 \\
DIRECT \citep{gong2026direct} \venue{ICML'26} & FLUX.1 & 0.088 & 0.922 & 0.069 & 19.4 & 0.249 & 0.832 & 0.759 \\
\midrule
GPT-Image-1 \venue{Tested 2026-08} & Proprietary & 0.468 & 0.568 & 0.162 & 12.6 & 0.607 & 0.765 & 0.767 \\
GPT-Image-2 \venue{Tested 2026-08} & Proprietary & 0.206 & 0.796 & 0.113 & 15.0 & 0.432 & 0.872 & 0.888 \\
Nano Banana 2 \venue{Tested 2026-08} & Proprietary & 0.175 & 0.844 & 0.094 & 16.1 & 0.365 & 0.901 & 0.874 \\
\midrule
\rowcolor{ForestGreen!12}
\textbf{D2R (Predicted Depth)} & FLUX.1 & \textbf{0.048} & \textbf{0.978} & \textbf{0.028} & \textbf{22.2} & \textbf{0.172} & \textbf{0.927} & \textbf{0.931} \\
\textcolor{gray}{D2R (Ground-Truth Depth)} & \textcolor{gray}{FLUX.1} & \textcolor{gray}{0.044} & \textcolor{gray}{0.982} & \textcolor{gray}{0.023} & \textcolor{gray}{22.7} & \textcolor{gray}{0.167} & \textcolor{gray}{0.931} & \textcolor{gray}{0.938} \\
\bottomrule
\end{tabular}}
\end{table}
\begin{table}[t]
\centering
\caption{Category-Disjoint Results on AnyInsertion++.}
\label{tab:main_ood}
\fitwidth{\setlength{\tabcolsep}{3pt}%
\begin{tabular}{llccccccc}
\toprule
\multirow{2}{*}{Method} & \multirow{2}{*}{Backbone} & \multicolumn{3}{c}{Geometric} & \multicolumn{2}{c}{Photometric} & \multicolumn{2}{c}{Identity} \\
\cmidrule(lr){3-5}\cmidrule(lr){6-7}\cmidrule(lr){8-9}
& & AbsRel$\downarrow$ & $\delta_1\uparrow$ & RMSE$\downarrow$ & PSNR$\uparrow$ & LPIPS$\downarrow$ & CLIP-I$\uparrow$ & DINO-I$\uparrow$ \\
\midrule
Paint-by-Example \citep{yang2023paint} \venue{CVPR'23} & SD1.5 & 0.136 & 0.871 & 0.074 & 17.6 & 0.258 & 0.676 & 0.401 \\
TF-ICON \citep{lu2023tficon} \venue{ICCV'23} & SD2.1 & 0.382 & 0.633 & 0.135 & 16.9 & 0.366 & 0.542 & 0.217 \\
AnyDoor \citep{chen2024anydoor} \venue{CVPR'24} & SD2.1 & 0.258 & 0.752 & 0.096 & 17.2 & 0.308 & 0.714 & 0.503 \\
MimicBrush \citep{chen2024mimicbrush} \venue{NeurIPS'24} & SD1.5 & 0.090 & 0.926 & 0.049 & 20.2 & 0.174 & 0.795 & 0.669 \\
BIFR\"OST \citep{li2024bifrost} \venue{NeurIPS'24} & SD2.1 & 0.120 & 0.905 & 0.052 & 19.4 & 0.243 & 0.731 & 0.565 \\
DreamFuse \citep{huang2025dreamfuse} \venue{ICCV'25} & FLUX.1 & 0.227 & 0.799 & 0.089 & 17.3 & 0.301 & 0.746 & 0.675 \\
UniCombine \citep{wang2025unicombine} \venue{ICCV'25} & FLUX.1 & 0.190 & 0.839 & 0.079 & 19.3 & 0.226 & 0.796 & 0.764 \\
Personalize-Anything \citep{feng2026personalize} \venue{AAAI'26} & FLUX.1 & 0.188 & 0.830 & 0.074 & 19.7 & 0.322 & 0.704 & 0.602 \\
Insert-Anything$^{*}$ \citep{song2026insert} \venue{AAAI'26} & FLUX.1 & 0.119 & 0.914 & 0.048 & 21.6 & 0.176 & 0.863 & 0.887 \\
SHINE \citep{lu2026shine} \venue{ICLR'26} & FLUX.1 & 0.218 & 0.827 & 0.088 & 17.5 & 0.295 & 0.647 & 0.443 \\
HiFi-Inpaint \citep{liu2026hifiinpaint} \venue{CVPR'26} & FLUX.1 & 0.133 & 0.893 & 0.067 & 18.7 & 0.235 & 0.746 & 0.665 \\
DIRECT \citep{gong2026direct} \venue{ICML'26} & FLUX.1 & 0.137 & 0.877 & 0.073 & 19.4 & 0.221 & 0.828 & 0.749 \\
\midrule
GPT-Image-1 \venue{Tested 2026-08} & Proprietary & 0.492 & 0.591 & 0.149 & 13.2 & 0.511 & 0.786 & 0.782 \\
GPT-Image-2 \venue{Tested 2026-08} & Proprietary & 0.300 & 0.735 & 0.116 & 14.8 & 0.407 & 0.902 & 0.912 \\
Nano Banana 2 \venue{Tested 2026-08} & Proprietary & 0.237 & 0.812 & 0.091 & 16.5 & 0.328 & 0.922 & 0.905 \\
\midrule
\rowcolor{ForestGreen!12}
\textbf{D2R (Predicted Depth)} & FLUX.1 & \textbf{0.067} & \textbf{0.970} & \textbf{0.030} & \textbf{24.0} & \textbf{0.117} & \textbf{0.923} & \textbf{0.916} \\
\textcolor{gray}{D2R (Ground-Truth Depth)} & \textcolor{gray}{FLUX.1} & \textcolor{gray}{0.059} & \textcolor{gray}{0.977} & \textcolor{gray}{0.024} & \textcolor{gray}{24.8} & \textcolor{gray}{0.108} & \textcolor{gray}{0.927} & \textcolor{gray}{0.929} \\
\bottomrule
\end{tabular}}
\end{table}

\noindent\textbf{Paired and Unpaired Performance.} D2R outperforms all 15 open- and closed-source baselines on every reported geometric, photometric, and identity metric across both paired splits (Tabs.~\ref{tab:main_id} and~\ref{tab:main_ood}). Relative to the strongest baseline for each metric, it reduces AbsRel by at least 22\% and LPIPS by at least 19\%, while improving PSNR by at least 1.0\,dB. D2R also leads both identity metrics on all three unpaired benchmarks, reducing mean reference cosine distance by 55.4\% with CLIP and 22.6\% with DINO over Insert-Anything$^*$ (Tab.~\ref{tab:unpaired}; App.~\ref{sec:appx_unpaired_identity}). These gains occur \emph{without an identity-specific loss}. \sh{We relate them to reference structure: objects with similar semantics and colors can differ in shape and part arrangement, which D2R better preserves in the displayed examples (Fig.~\ref{fig:teaser}). Composite depth supplies these structural cues while RGB supervision learns appearance. Their joint contribution is consistent with higher embedding similarity.}

\noindent\textbf{Geometry from Standard Compositing Inputs.} On the paired splits, replacing Stage-1 predictions with GT composite depth lowers rendered-image AbsRel by about 8 to 12\% and raises PSNR by 0.5 to 0.8\,dB. The predicted condition already supports leading performance on unseen categories without a completed target image, measured depth, or a user-posed 3D proxy at inference.

Unpaired benchmarks further assess reference identity across real-world photographs and stylized scenes, extending evaluation beyond paired image comparisons. Surprisingly, the same pipeline also produces geometry-aware composites in cartoon and artistic scenes. A physical depth sensor viewing a flat picture or screen measures that surface, rather than distances within the depicted scene. A learned depth estimator instead infers relative depth from RGB cues. D2R repurposes this capability to predict the intended composite's geometry and guide RGB synthesis. Appendix~\ref{sec:appx_qual} discusses this observation through the cartoon and illustrated examples in Figs.~\ref{fig:qual_tficon} and~\ref{fig:qual_aicomposer}.

\begin{table}[t]
\centering
\caption{Identity on Unpaired Compositing Benchmarks.}
\label{tab:unpaired}
\fitwidth{\setlength{\tabcolsep}{4pt}%
\begin{tabular}{llcccccc}
\toprule
\multirow{2}{*}{Method} & \multirow{2}{*}{Backbone} & \multicolumn{2}{c}{TF-ICON} & \multicolumn{2}{c}{AIComposer} & \multicolumn{2}{c}{ComplexCompo} \\
\cmidrule(lr){3-4}\cmidrule(lr){5-6}\cmidrule(lr){7-8}
& & CLIP-I$\uparrow$ & DINO-I$\uparrow$ & CLIP-I$\uparrow$ & DINO-I$\uparrow$ & CLIP-I$\uparrow$ & DINO-I$\uparrow$ \\
\midrule
Paint-by-Example \citep{yang2023paint} \venue{CVPR'23} & SD1.5 & 0.689 & 0.342 & 0.696 & 0.387 & 0.528 & 0.223 \\
TF-ICON \citep{lu2023tficon} \venue{ICCV'23} & SD2.1 & 0.573 & 0.258 & 0.654 & 0.374 & 0.489 & 0.199 \\
AnyDoor \citep{chen2024anydoor} \venue{CVPR'24} & SD2.1 & 0.752 & 0.628 & 0.777 & 0.625 & 0.718 & 0.645 \\
MimicBrush \citep{chen2024mimicbrush} \venue{NeurIPS'24} & SD1.5 & 0.753 & 0.537 & 0.767 & 0.575 & 0.645 & 0.415 \\
BIFR\"OST \citep{li2024bifrost} \venue{NeurIPS'24} & SD2.1 & 0.730 & 0.557 & 0.736 & 0.548 & 0.685 & 0.565 \\
DreamFuse \citep{huang2025dreamfuse} \venue{ICCV'25} & FLUX.1 & 0.652 & 0.410 & 0.718 & 0.597 & 0.650 & 0.569 \\
UniCombine \citep{wang2025unicombine} \venue{ICCV'25} & FLUX.1 & 0.697 & 0.555 & 0.747 & 0.641 & 0.482 & 0.297 \\
Personalize-Anything \citep{feng2026personalize} \venue{AAAI'26} & FLUX.1 & 0.739 & 0.542 & 0.808 & 0.716 & 0.668 & 0.615 \\
Insert-Anything$^{*}$ \citep{song2026insert} \venue{AAAI'26} & FLUX.1 & 0.760 & 0.679 & 0.816 & 0.794 & 0.778 & 0.724 \\
SHINE \citep{lu2026shine} \venue{ICLR'26} & FLUX.1 & 0.402 & 0.034 & 0.469 & 0.155 & 0.557 & 0.376 \\
HiFi-Inpaint \citep{liu2026hifiinpaint} \venue{CVPR'26} & FLUX.1 & 0.468 & 0.116 & 0.583 & 0.352 & 0.449 & 0.219 \\
DIRECT \citep{gong2026direct} \venue{ICML'26} & FLUX.1 & 0.714 & 0.529 & 0.742 & 0.595 & 0.742 & 0.683 \\
\midrule
GPT-Image-1 \venue{Tested 2026-08} & Proprietary & 0.612 & 0.416 & 0.645 & 0.567 & 0.469 & 0.241 \\
GPT-Image-2 \venue{Tested 2026-08} & Proprietary & 0.625 & 0.450 & 0.669 & 0.608 & 0.456 & 0.178 \\
Nano Banana 2 \venue{Tested 2026-08} & Proprietary & 0.591 & 0.345 & 0.632 & 0.487 & 0.413 & 0.150 \\
\midrule
\rowcolor{ForestGreen!12}
\textbf{D2R (Predicted Depth)} & FLUX.1 & \textbf{0.902} & \textbf{0.783} & \textbf{0.915} & \textbf{0.813} & \textbf{0.895} & \textbf{0.781} \\
\bottomrule
\end{tabular}}
\end{table}

\section{Ablation Study and Discussion}
\label{sec:ablation}

\begin{table}[!ht]
\centering
\caption{Effect of the Supervision Target with a Matched Depth Predictor.}
\label{tab:supervision_final}
\fitwidth{\setlength{\tabcolsep}{6pt}%
\begin{tabular}{lccccc}
\toprule
\multirow{2}{*}{Supervision Target} & \multirow{2}{*}{\makecell{Decoder\\Backpropagation}} & \multicolumn{2}{c}{AbsRel $\downarrow$} & \multicolumn{2}{c}{$\delta_1 \uparrow$} \\
\cmidrule(lr){3-4}\cmidrule(lr){5-6}
 & & ID & OOD & ID & OOD \\
\midrule
\rowcolor{ForestGreen!12}
\textbf{Encoder Feature Only} & \no & \textbf{0.0462} & \textbf{0.0751} & \textbf{0.977} & \textbf{0.965} \\
Decoded Depth Only & \yes & 0.0669 & 0.1094 & 0.955 & 0.931 \\
\bottomrule
\end{tabular}}
\end{table}

\noindent\textbf{Learning the Completed Scene's Depth Features.} \sh{Table~\ref{tab:supervision_final} changes only the supervision target: both variants train the same injector while keeping the depth encoder and decoder frozen. The paired composite supplies features that the unchanged decoder already maps to the intended depth. Matching these features therefore teaches reference-conditioned prediction within the pretrained representation, whereas depth-only supervision constrains the decoded map. Feature targets lower ID and OOD Stage-1 AbsRel by 30.9\% and 31.4\%, while avoiding decoder backpropagation. With the same trained RGB renderer, they also lower OOD rendered-image AbsRel from 0.089 to 0.067, connecting the supervision choice to final RGB geometry. Adding decoded-depth loss to feature matching improves ID by about 5\% but worsens OOD by about 3\% and reduces the largest single-GPU batch by about 30\% (App.~\ref{sec:appx_supervision_comparison}).}

\noindent\textbf{Feature Correction Versus Diffusion-Based Depth Generation.} Following the depth-generation approach in 3DIS~\citep{zhou20253dis}, we build a compositing-specific diffusion baseline with FLUX.1-Fill and Redux. It denoises composite-depth maps using the same Stage-1 training pairs. D2R reduces OOD AbsRel from 0.1740 to 0.0751 (56.8\%) with one feed-forward pass instead of 28 denoising steps. This compares depth-prediction strategies; Tab.~\ref{tab:supervision_final} isolates supervision within the same predictor (App.~\ref{sec:appx_depth_priors}).

\begin{table}[!ht]
\centering
\caption{Geometric and Photometric Performance Across Geometry Strategies.}
\label{tab:ladder}
\fitwidth{\setlength{\tabcolsep}{5pt}%
\begin{tabular}{lcccc}
\toprule
\multirow{2}{*}{Geometry Provisioning} & \multicolumn{2}{c}{Out-of-Distribution} & \multicolumn{2}{c}{In-Distribution} \\
\cmidrule(lr){2-3}\cmidrule(lr){4-5}
 & AbsRel $\downarrow$ & PSNR $\uparrow$ & AbsRel $\downarrow$ & PSNR $\uparrow$ \\
\midrule
(a) Empty Depth Patch (Matched RGB Baseline) & 0.119 & 21.6 & 0.062 & 21.2 \\
(b) Heuristic Depth Fusion & 0.098 & 20.3 & 0.076 & 18.9 \\
(c) Latent Optimization Toward Predicted Depth & 0.224 & 16.9 & 0.139 & 17.3 \\
(d) Joint Denoising from Background Depth & 0.080 & 22.2 & 0.065 & 20.9 \\
\rowcolor{ForestGreen!12}
\textbf{(e) Sequential Synthesis with Fixed Depth (D2R; Ours)} & \textbf{0.067} & \textbf{24.0} & \textbf{0.048} & \textbf{22.2} \\
\bottomrule
\end{tabular}}
\end{table}

\noindent\textbf{Predicting Geometry Before RGB Synthesis.} The matched RGB-only baseline in Tab.~\ref{tab:ladder}(a) shares D2R's data and setup but infers object structure during appearance generation, reaching OOD AbsRel of 0.119 and PSNR of 21.6\,dB. Table~\ref{tab:ladder}(b) lowers AbsRel to 0.098 but reduces PSNR to 20.3\,dB. Rescaling reference depth into the background supplies observed geometry without predicting the object's shape in its new scene, so this cue can conflict with the intended composite.

Latent optimization in Tab.~\ref{tab:ladder}(c) already targets Stage-1 depth, yet worsens both scores to 0.224 and 16.9\,dB. It adjusts latents using depth estimated from provisional RGB images without training the renderer to follow that condition. Such depth-driven updates can be unreliable before object structure emerges and do not explicitly enforce reference appearance. Joint denoising in Tab.~\ref{tab:ladder}(d) reaches OOD AbsRel of 0.080 but raises ID error from 0.062 to 0.065: starting from background depth leaves composite geometry to be resolved alongside RGB. In Tab.~\ref{tab:ladder}(e), D2R instead predicts that geometry first and trains RGB synthesis under a fixed condition, improving both metrics on both splits. These results compare complete implementations: geometry construction and training also differ, so they do not isolate generation order (App.~\ref{sec:appx_s2}).

\begin{table}[t]
\centering
\caption{Effect of Depth Conditioning and Auxiliary Supervision.}
\label{tab:loss}
\fitwidth{\setlength{\tabcolsep}{4pt}%
\begin{tabular}{lcccccccc}
\toprule
 & \multirow{2}{*}{$\mathcal{L}_{\mathrm{pixel}}$} & \multirow{2}{*}{$\mathcal{L}_{\mathrm{depth}}$} & \multirow{2}{*}{$\mathcal{L}_{\mathrm{cons}}$} & \multicolumn{3}{c}{Geometric} & \multicolumn{2}{c}{Photometric} \\
\cmidrule(lr){5-7}\cmidrule(lr){8-9}
 & & & & AbsRel $\downarrow$ & $\delta_1 \uparrow$ & RMSE $\downarrow$ & PSNR $\uparrow$ & LPIPS $\downarrow$ \\
\midrule
(a) & \yes & \no\,{\footnotesize(w/o depth input)} & \no & 0.119 & 0.914 & 0.048 & 21.6 & 0.176 \\
(b) & \yes & \no\,{\footnotesize(w/ depth input)} & \no & 0.108 & 0.931 & 0.046 & 21.9 & 0.168 \\
(c) & \yes & \yes & \no & 0.081 & 0.958 & 0.038 & 23.1 & 0.127 \\
(d) & \yes & \no & \yes\,{\footnotesize$(1{-}t)$} & 0.084 & 0.947 & 0.043 & 22.1 & 0.158 \\
(e) & \yes & \yes & \yes\,{\footnotesize(uniform)} & 0.075 & 0.950 & 0.041 & 22.8 & 0.135 \\
\rowcolor{ForestGreen!12}
\textbf{(f)} & \yes & \yes & \yes\,{\footnotesize$(1{-}t)$} & \textbf{0.067} & \textbf{0.970} & \textbf{0.030} & \textbf{24.0} & \textbf{0.117} \\
\bottomrule
\end{tabular}}
\end{table}

\noindent\textbf{Learning to Use the Depth Condition.} Table~\ref{tab:loss} first tests depth input under RGB-only supervision: OOD AbsRel falls from 0.119 to 0.108 (a to b). With depth supply fixed, depth-patch denoising and RGB-depth consistency reduce error further to 0.081 and 0.084, respectively (b to c/d). The former supervises the depth patch; the latter compares geometry estimated from generated RGB with the paired target. Combining them improves on either objective alone (c/d to f). With both enabled, replacing uniform consistency weighting with $1-t$ lowers AbsRel from 0.075 to 0.067 and raises PSNR from 22.8 to 24.0\,dB (e to f), reducing the influence of noisy early RGB estimates. All conditioned variants use Stage-1 predictions at evaluation (App.~\ref{sec:appx_render_ablations}).

\noindent\textbf{Learning from Accurate and Predicted Depth.} Stage~2 must handle errors in the depth supplied at inference. Table~\ref{tab:mix} varies training inputs with GT supervision fixed and scores generated RGB geometry. With GT-only training, OOD AbsRel rises from 0.062 under GT depth to 0.081 under predictions. Predicted-only training improves the latter to 0.069 but weakens performance with GT depth. Blending both sources can average mismatched object boundaries. Selection instead pairs either intact map with the same GT target. This exposes the renderer to prediction errors without using them as supervision targets. Selection gives the lowest error under both input types: OOD AbsRel of 0.059 with GT depth and 0.067 with predictions (App.~\ref{sec:appx_stage2}).

\begin{table}[t]
\centering
\caption{Training the Renderer to Handle Depth Prediction Errors.}
\label{tab:mix}
\fitwidth{\setlength{\tabcolsep}{4pt}%
\begin{tabular}{lcccc}
\toprule
\multirow{2}{*}{Depth Supply Strategy} & \multicolumn{2}{c}{\makecell{Ground-Truth Depth\\at Inference}} & \multicolumn{2}{c}{\makecell{Predicted Depth\\at Inference}} \\
\cmidrule(lr){2-3}\cmidrule(lr){4-5}
 & ID AbsRel $\downarrow$ & OOD AbsRel $\downarrow$ & ID AbsRel $\downarrow$ & OOD AbsRel $\downarrow$ \\
\midrule
Ground-Truth Depth Only & 0.049 & 0.062 & 0.056 & 0.081 \\
Predicted Depth Only & 0.052 & 0.066 & 0.053 & 0.069 \\
\midrule
Blending (\emph{i.e.}, mixup \citep{zhang2018mixup}) & 0.051 & 0.064 & 0.055 & 0.070 \\
\rowcolor{ForestGreen!12}
Selection, $p=0.5$ & \textbf{0.044} & \textbf{0.059} & \textbf{0.048} & \textbf{0.067} \\
\bottomrule
\end{tabular}}
\end{table}

\begin{table}[!t]
\centering
\caption{Depth-Conditioned RGB Synthesis Across Generators.}
\label{tab:generator}
\fitwidth{\setlength{\tabcolsep}{4pt}%
\begin{tabular}{lcccccccc}
\toprule
\multirow{2}{*}{Diffusion Backbone} & \multicolumn{3}{c}{AbsRel $\downarrow$} & \multicolumn{2}{c}{$\delta_1 \uparrow$} & \multicolumn{3}{c}{PSNR (dB) $\uparrow$} \\
\cmidrule(lr){2-4}\cmidrule(lr){5-6}\cmidrule(lr){7-9}
 & Pretrained & + D2R & $\Delta$ & Pretrained & + D2R & Pretrained & + D2R & $\Delta$ \\
\midrule
SD2-Inpainting \citep{rombach2022high} \venue{CVPR'22} & 0.158 & \cellcolor{ForestGreen!12}\textbf{0.110} & {\footnotesize\textcolor{gray}{$-$31\%}} & 0.830 & \cellcolor{ForestGreen!12}\textbf{0.925} & 18.10 & \cellcolor{ForestGreen!12}\textbf{19.57} & {\footnotesize\textcolor{gray}{$+$1.47}} \\
SDXL-Inpainting \citep{podell2024sdxl} \venue{ICLR'24} & 0.155 & \cellcolor{ForestGreen!12}\textbf{0.105} & {\footnotesize\textcolor{gray}{$-$32\%}} & 0.832 & \cellcolor{ForestGreen!12}\textbf{0.930} & 18.00 & \cellcolor{ForestGreen!12}\textbf{20.13} & {\footnotesize\textcolor{gray}{$+$2.13}} \\
Qwen-Inpainting \citep{wu2025qwenimage} \venue{arXiv'25} & 0.100 & \cellcolor{ForestGreen!12}\textbf{0.082} & {\footnotesize\textcolor{gray}{$-$18\%}} & 0.920 & \cellcolor{ForestGreen!12}\textbf{0.954} & 20.47 & \cellcolor{ForestGreen!12}\textbf{21.55} & {\footnotesize\textcolor{gray}{$+$1.08}} \\
FLUX.1-Fill & 0.088 & \cellcolor{ForestGreen!12}\textbf{0.067} & {\footnotesize\textcolor{gray}{$-$23\%}} & 0.942 & \cellcolor{ForestGreen!12}\textbf{0.970} & 21.97 & \cellcolor{ForestGreen!12}\textbf{24.02} & {\footnotesize\textcolor{gray}{$+$2.05}} \\
\bottomrule
\end{tabular}}
\end{table}

\noindent\textbf{Correcting Object and Scene Features Together.} The mask localizes the compositing region, but relative depth also depends on its surroundings. Restricting correction to masked tokens raises average Stage-1 AbsRel by about 2.5\%, with errors increasing inside and outside the mask. Sharing one injector across four layers raises ID and OOD error by about 10\% and 3\%. These results favor scene-wide correction with separate layer parameters, letting reference information influence both object and context features (App.~\ref{sec:appx_stage1}). Injection also improves all six depth backbones against their own targets, supporting reuse across estimators without ranking absolute accuracy (Tab.~\ref{tab:depth_full}).

\noindent\textbf{Adapting Different RGB Generators.} We supply identical Stage-1 depth to pretrained and adapted SD2~\citep{rombach2022high}, SDXL~\citep{podell2024sdxl}, Qwen~\citep{wu2025qwenimage}, and FLUX.1-Fill. Adapted configurations reduce OOD AbsRel by 18\% to 32\% and increase PSNR by 1.08 to 2.13\,dB (Tab.~\ref{tab:generator}). Because both versions receive the same depth, the gains reflect improved RGB generation under that condition. This supports reuse of the depth interface across four generators (App.~\ref{sec:appx_renderer_results}).

\noindent\textbf{Consistency Across Cases and Object Regions.} D2R improves OOD full-frame AbsRel over Insert-Anything$^*$ on 103 of 139 cases, with gains inside objects and along their boundaries. Appendix~\ref{sec:appx_paired_uncertainty} reports paired and image-cluster uncertainty and sensitivity to the largest gains. Five default training runs per stage show about 1\% average relative variation across geometric, photometric, and identity metrics (App.~\ref{sec:appx_render_training}).

\noindent\textbf{Generalization Beyond Rigid Objects.} To test whether gains extend to objects whose shape can change, we evaluate the annotated rigid and non-rigid groups separately. D2R improves OOD AbsRel and PSNR over Insert-Anything$^*$ in both groups; non-rigid AbsRel falls from 0.133 to 0.071. This evidence covers 20 non-rigid OOD cases (App.~\ref{sec:appx_bench}, Tab.~\ref{tab:rigid}).

\noindent\textbf{Agreement Across Depth Scorers.} We hold generated RGB fixed and change only the depth scorer. D2R ranks first among sixteen methods on both splits under all five scorers, including Depth Pro and Metric-Anything, which supply no training targets to the default model (App.~\ref{sec:appx_metrics}).

\noindent\textbf{Computational Cost.} On RTX PRO 6000, the complete Stage-1 predictor takes 56.5\,ms and 5.21\,GB peak inference memory. Relative to 22.34\,s for FLUX RGB synthesis, this adds about 0.25\% latency. Predicted depth is reused throughout all 28 RGB sampling steps. The injector adds 272.4M parameters. The lower-resolution SD2 configuration takes 0.87\,s overall, trading geometric and photometric quality for speed (Tab.~\ref{tab:generator}; App.~\ref{sec:appx_cost}).

\noindent\textbf{Limitations.} D2R keeps its predicted depth fixed during RGB synthesis, so errors in object shape or placement can persist in the final image. Even with plausible geometry, an object may retain reference colors or textures that do not match a grayscale or illustrated background. Depth describes spatial structure but does not specify the color and texture changes needed to match the background. Appendix~\ref{sec:appx_limits} provides examples and further discussion.

\section{Conclusion}
\label{sec:conclusion}

\sh{D2R repurposes a frozen depth estimator by learning to predict encoder features of an intended composite from its background and reference. Completed-scene feature targets improve both the intermediate depth and final RGB geometry under the same renderer, linking the prediction objective to downstream compositing. The complete pipeline also improves photometric quality and reference identity.} The depth predictor also outperforms our diffusion-based composite-depth baseline without iterative denoising. Feature correction adapts multiple depth estimators, while the resulting depth condition supports different RGB generators. These results extend depth estimation from observed images to intended composites, providing a reusable geometric condition for RGB synthesis without a depth sensor or user-posed 3D proxy. We believe this interface can support future harmonization and relighting that adapt object color and illumination to the surrounding scene while preserving the intended object shape and its spatial relationships with nearby surfaces.

\section*{Acknowledgements}
This work was partly supported by the KHIDI grant funded by the Korean government (MOHW) [No.RS-2025-02307233, No.RS-2026-25613012], the NRF or IITP grants funded by the Korean government (MSIT) [No.05-26-04-0094, No.RS-2026-25472075, No.RS-2025-02305581, No.RS-2025-25442338, and No.RS-2021-II211343], the ITIP grant funded by the Korean government (MOTIR) [No.RS-2026-25549946], the Research grant from SNU, and the Strategic Hub grant for International Research Collaboration of SNU. Kyungsu Kim is affiliated with the School of Transdisciplinary Innovations, Department of Biomedical Science, Interdisciplinary Program in AI (IPAI) \& Graduate School of AI, Medical Research Center, and AI Institute at SNU.

\bibliography{main}
\bibliographystyle{plainnat}

\clearpage
\appendix
\clubpenalty=10000
\widowpenalty=10000
\displaywidowpenalty=10000
\brokenpenalty=10000
\makeatletter\setlength{\@fptop}{0pt}\makeatother

\renewcommand{\thesection}{\Alph{section}}

\makeatletter
\@addtoreset{table}{section}
\@addtoreset{figure}{section}
\makeatother
\renewcommand{\thetable}{\thesection\arabic{table}}
\renewcommand{\thefigure}{\thesection\arabic{figure}}
\def\theHtable{\thesection.\arabic{table}}
\def\theHfigure{\thesection.\arabic{figure}}
\setcounter{table}{0}
\setcounter{figure}{0}

\makeatletter
\@addtoreset{equation}{section}
\makeatother
\renewcommand{\theequation}{\thesection\arabic{equation}}
\def\theHequation{\thesection.\arabic{equation}}
\setcounter{equation}{0}

\phantomsection
{\huge\centering\bfseries Appendix\par}
\addcontentsline{toc}{section}{Appendix}
\vspace{1.5em}
{\large\centering\bfseries Contents\par}
\vspace{1em}

\startcontents[appendix]
\begingroup
\small
\setlength{\parskip}{3.5pt}
\titlecontents{section}[1.7em]
  {\addvspace{0.4em}\bfseries}
  {\contentslabel{1.7em}}{}
  {\titlerule*[0.7pc]{.}\contentspage}
\titlecontents{subsection}[4.1em]
  {\addvspace{0.1em}}
  {\contentslabel{2.4em}}{}
  {\titlerule*[0.7pc]{.}\contentspage}
\titlecontents{subsubsection}[7.2em]
  {\addvspace{0em}}
  {\contentslabel{3.1em}}{}
  {\titlerule*[0.7pc]{.}\contentspage}
\printcontents[appendix]{}{1}{\setcounter{tocdepth}{3}}
\endgroup
\clearpage

\section{Extended Related Work}
\label{sec:appx_related}

\subsection{{Geometric Conditions for Reference-Based Compositing}}
\label{sec:appx_compositing_comparison}

\begin{table}[!ht]
\centering
\caption{{Geometry Conditions in Reference-Based Compositing.}}
\label{tab:related_perwork}
\fitwidth{\setlength{\tabcolsep}{2pt}%
\begin{tabular}{lccccc}
\toprule
Method & \makecell{Composite-Scene\\Depth} & \makecell{Predicted,\\Not Assembled} & \makecell{Scene-Wide\\Prediction} & \makecell{2D Inputs\\Alone} & \makecell{Fixed Geometry\\Before RGB} \\
\midrule
\multicolumn{6}{l}{\textbf{(a) Implicit Composite Depth (RGB Only)}} \\
Paint-by-Example \citep{yang2023paint} \venue{CVPR'23}     & \no  & \no  & \no  & \yes & \no  \\
AnyDoor \citep{chen2024anydoor} \venue{CVPR'24}              & \no  & \no  & \no  & \yes & \no  \\
Insert-Anything \citep{song2026insert} \venue{AAAI'26}      & \no  & \no  & \no  & \yes & \no  \\
Personalize-Anything \citep{feng2026personalize} \venue{AAAI'26} & \no  & \no  & \no  & \yes & \no  \\
SHINE \citep{lu2026shine} \venue{ICLR'26}                & \no  & \no  & \no  & \yes & \no  \\
InsertFuse \citep{li2026insertfuse} \venue{arXiv'26}          & \no  & \no  & \no  & \yes & \no  \\
HiFi-Inpaint \citep{liu2026hifiinpaint} \venue{CVPR'26}         & \no  & \no  & \no  & \yes & \no  \\
\midrule
\multicolumn{6}{l}{\textbf{(b-1) Depth from the Existing Scene}} \\
MimicBrush \citep{chen2024mimicbrush} \venue{NeurIPS'24}        & \no  & \no  & \no  & \yes & \yes \\
\midrule
\multicolumn{6}{l}{\textbf{(b-2) Geometry from Additional Inputs or Assembly Rules}} \\
BIFR\"OST (Instruction Branch) \citep{li2024bifrost} \venue{NeurIPS'24}         & \yes & \no  & \no  & \no  & \yes \\
DIRECT \citep{gong2026direct} \venue{ICML'26}               & \no & \no  & \no  & \no  & \yes \\
SpatialHand \citep{wang2026spatialhand} \venue{ICLR'26}          & \yes & \no  & \no  & \no  & \yes \\
\midrule
\multicolumn{6}{l}{{\textbf{(c) Predicted Composite-Scene Depth}}} \\
Person-In-Situ (Two-Stage Configuration) \citep{masuda2026personinsitu} \venue{CAVW'26} & \yes & \yes & \yes & \no & \yes \\
\rowcolor{ForestGreen!12}
\textbf{D2R (Ours)}                  & \yes & \yes & \yes & \yes & \yes \\
\bottomrule
\end{tabular}}
\end{table}

\noindent{\textbf{Comparison Criteria.} Table~\ref{tab:related_perwork} extends Tab.~\ref{tab:related} for reference-based compositing and specialized insertion tasks. The first three criteria ask whether depth describes the intended composite, is predicted rather than assembled, and is predicted jointly across the scene. A full-frame map assembled from separate depths does not satisfy the third criterion. The fourth tests use of standard compositing inputs alone, including dataset-provided reference masks (App.~\ref{sec:appx_canvas}). The final criterion asks whether a geometric condition is fixed before RGB synthesis; this includes depth maps and DIRECT's rendered RGB proxy. Person-In-Situ is classified by its two-stage configuration; its rendered body-depth condition is additional to the standard compositing inputs.}

\noindent\textbf{RGB Compositing Without Explicit Composite Depth.} Paint-by-Example~\citep{yang2023paint} and AnyDoor~\citep{chen2024anydoor} preserve reference identity through image features. Insert-Anything~\citep{song2026insert} places the reference and target on one canvas and adapts an inpainting transformer with LoRA~\citep{hu2022lora}. It provides our matched baseline because D2R retains this rendering interface while adding depth prediction and supervision. Other mechanisms include token substitution in Personalize-Anything~\citep{feng2026personalize}, iterative identity control in SHINE~\citep{lu2026shine}, and high-frequency feature preservation in HiFi-Inpaint~\citep{liu2026hifiinpaint}. InsertFuse~\citep{li2026insertfuse} maps the insertion mask and its distance transform to visual-token features, specifying the editing region in 2D. Related appearance-conditioned methods include ObjectStitch~\citep{song2023objectstitch}, IMPRINT~\citep{song2024imprint}, ObjectDrop~\citep{winter2024objectdrop}, ObjectMate~\citep{winter2025objectmate}, Add-it~\citep{tewel2025addit}, CareCom~\citep{chen2026carecom}, and A$^2$-Edit~\citep{zheng2026a2edit}. Their features can encode shape and geometry implicitly. D2R makes scene-wide relative depth an explicit condition before RGB synthesis.

\noindent{\textbf{Separating Object Structure and Appearance.} PAIR Diffusion~\citep{goel2024pair} controls object-level editing with segmentation-based structure and image-encoder appearance features. Refacade~\citep{huang2026refacade} learns to remove source texture while retaining object geometry, then uses shuffled reference patches to transfer texture. These methods separate structure from appearance through different conditioning mechanisms. D2R instead predicts the completed scene's depth, including the inserted object's structure and spatial relationship to the background.}

\noindent\textbf{Depth from the Existing Scene.} MimicBrush~\citep{chen2024mimicbrush} transfers reference features through a second U-Net and optionally uses depth estimated from the input background before masking the compositing region. Table~\ref{tab:related} describes this depth-enabled configuration. Its depth map provides the pre-edit geometry, including pixels later masked. Zero-shot depth-aware editing~\citep{parihar2025zeroshot} decomposes supplied depth into layers that a user transforms, requiring depth beyond the three compositing inputs. Both methods start from existing geometry; neither predicts completed-scene depth from the standard compositing inputs.

\noindent\textbf{Geometry from Additional Inputs or Assembly Rules.} In its instruction-driven branch, BIFR\"OST~\citep{li2024bifrost} predicts an object box and depth value from language, then rescales reference depth into the background map. Table~\ref{tab:related_perwork} classifies this branch, whereas our quantitative comparisons use mask-based inpainting (App.~\ref{sec:appx_baselines}). Table~\ref{tab:ladder} separately tests depth fusion in D2R.

{Explicit pose and placement controls provide another way to specify the intended arrangement. DIRECT~\citep{gong2026direct} uses a rendered RGB proxy to specify object pose; SpatialHand~\citep{wang2026spatialhand} combines background depth with a user-selected placement depth in the insertion mask and a numerical orientation condition for 6DoF object insertion. Its depth assembly preserves nearer background occluders. FreeInsert~\citep{zhang2025freeinsert} and the dual-branch GeoEdit~\citep{he2026geoedit} reconstruct and manipulate an object already present. D2R uses only standard compositing inputs.}

\noindent{\textbf{Geometric Editing and Cross-Image Compositing.} FreeFine~\citep{zhu2025freefine} separates object transformation, source-region inpainting, and target-region refinement without training. Its cross-image extension refines coarse RGB composites assembled from source objects and source/target masks, so its scope also includes compositing. The distinction from D2R is how the target structure is obtained: FreeFine transforms and refines RGB content, whereas D2R learns to predict scene-wide composite depth before RGB generation.}

\noindent{\textbf{Depth-First Human Compositing.} Person-In-Situ~\citep{masuda2026personinsitu} predicts a scene's depth with the reference person inserted, then generates the RGB composite. Its two-stage configuration uses rendered 3D body depth, estimated scene depth, and a body-derived box mask alongside scene and reference images. It fine-tunes diffusion models on depth and RGB targets constructed from video frames. D2R instead uses a general monocular depth prior, learning reference-conditioned feature corrections while keeping the estimator frozen. Encoder features from paired completed scenes supervise the injector. This design requires no human-body representation and supports the diverse object types in our evaluation. The distinction concerns the inputs and learned representation, rather than the depth-before-RGB ordering.}

{Our paired evaluation covers 107 distinct object tags (87 ID and 20 OOD; Tab.~\ref{tab:rigid_tags}), with leading aggregate results on both splits. Among the 426 cases, 4 carry the person tag (0.9\%), whereas 49 carry accessory or garment tags (11.5\%). These counts use the literal \texttt{person\_person}, \texttt{accessory\_*}, and \texttt{garment\_*} tags, without manually regrouping other objects. The unpaired benchmarks do not provide reference-category tags, so identifying a person-only subset would require manual annotation. We therefore provide a conceptual comparison with Person-In-Situ.}

\subsection{{Scene Synthesis and Geometry Control in Related Tasks}}
\label{sec:appx_scene_comparison}

\noindent{\textbf{Geometry-Guided Scene Synthesis.} Compose-and-Conquer~\citep{lee2024compose} conditions RGB synthesis on foreground/background depth and exemplar semantics. 3DIS~\citep{zhou20253dis} predicts scene depth from instance layouts before rendering RGB. These methods establish depth-guided synthesis as a prior design choice. Table~\ref{tab:appx_scene_synthesis} distinguishes their geometry inputs and adaptation from D2R, which predicts composite depth from an existing background, reference object, and 2D compositing mask.}

\begin{table}[!ht]
\centering
\caption{{Geometry Inputs and Adaptation in Scene Synthesis.}}
\label{tab:appx_scene_synthesis}
{\setlength{\tabcolsep}{3pt}\renewcommand{\arraystretch}{1.15}
{\begin{tabular}{@{}>{\raggedright\arraybackslash}p{0.17\linewidth}>{\raggedright\arraybackslash}p{0.25\linewidth}>{\raggedright\arraybackslash}p{0.24\linewidth}>{\raggedright\arraybackslash}p{0.29\linewidth}@{}}
\toprule
Method & Task and Inputs & Geometry Condition & Learned Components \\
\midrule
Compose-and-Conquer \citep{lee2024compose} \venue{ICLR'24} & Composable synthesis from text, foreground/background depth, and exemplar semantics & Supplied foreground/background depth maps & Local/global fusers and cloned diffusion blocks; original diffusion backbone frozen \\
\addlinespace
3DIS \citep{zhou20253dis} \venue{ICLR'25} & Multi-instance generation from text and instance layouts & Scene depth generated before RGB & Fine-tuned depth generator and layout adapter; pretrained depth control and detail rendering without additional training \\
\addlinespace
\rowcolor{ForestGreen!12}
\textbf{D2R (Ours)} & Reference-based compositing from background, reference, and 2D compositing mask & Composite depth predicted before RGB & Depth-feature injector with frozen estimator; renderer LoRA \\
\bottomrule
\end{tabular}}}
\end{table}

\noindent{\textbf{Geometry from Explicit Object Transformations.} Diffusion Handles~\citep{pandey2024diffusionhandles} lifts diffusion activations into 3D using estimated source depth, applies a user-specified transform, and reprojects the activations to guide RGB sampling. The effects-sensitive GeoEdit~\citep{zhang2026effectssensitive} constructs transformed object images and masks, using mesh reconstruction for rotation, then performs in-context RGB inpainting with a trained FLUX.1-Fill model. It is distinct from the dual-branch GeoEdit~\citep{he2026geoedit} discussed above. These approaches use explicit transformations to specify the intended arrangement. D2R predicts that arrangement's relative depth from a separate reference, background, and compositing mask, without requiring a 3D transform.}

\noindent{\textbf{Depth-Conditioned Multi-View Canvases.} SIGNeRF~\citep{dihlmann2024signerf} edits NeRF scenes through depth-conditioned diffusion over a multi-view reference sheet. Geometry comes from the existing 3D scene and a selected region or external proxy mesh; the edited views are then used to update the NeRF. Its canvas organizes multiple views of supplied geometry. D2R's canvas instead combines a reference, masked background, and predicted composite depth for a single output image. The shared use of a canvas therefore does not imply the same source of geometric conditioning.}

\noindent\textbf{Geometry Control During Sampling.} FreeInpaint~\citep{gong2026freeinpaint} optimizes initial noise and intermediate latents without updating model weights. Closed-loop test-time control~\citep{zhao2026openloop} adjusts conditioning using feedback from generated outputs. These methods modify RGB generation at inference, whereas D2R predicts composite depth once and holds it fixed throughout RGB sampling. Their published tasks differ from reference-preserving compositing, so we compare compositing-specific alternatives in Tab.~\ref{tab:ladder}: Latent Optimization toward Stage-1 depth and Joint Denoising of depth and RGB initialized with background depth. Appendix~\ref{sec:appx_s2} specifies both baselines. These results concern the evaluated implementations, rather than iterative control in general.

\subsection{{Representation Prediction and Joint RGB-Depth Modeling}}
\label{sec:appx_representation_comparison}

\noindent{\textbf{Predictive Features and Joint RGB-Depth Models.} Table~\ref{tab:appx_geometry_models} separates representation choice from task. VGGT-World~\citep{sun2026vggtworld} predicts future geometry features from video observations and decodes them through frozen geometry components. D2R instead predicts features of a requested static composite from a background, reference object, and compositing mask. JointDiT~\citep{kwon2025jointdit} and Modality Forcing~\citep{duisterhof2026modalityforcing} support both joint RGB-depth modeling and depth-conditioned RGB generation. To test the joint-generation alternative in reference-based compositing, we implement a joint depth/RGB denoising baseline using D2R's data and rendering setup (Tab.~\ref{tab:ladder}; App.~\ref{sec:appx_render_ablations}). This is a task-specific comparison, rather than a reproduction or ranking of those models. D2R focuses on predicting the required depth condition from the compositing inputs before RGB generation.}

\begin{table}[!ht]
\centering
\caption{{Prediction Targets and Adaptation Across Geometry Models.}}
\label{tab:appx_geometry_models}
{\setlength{\tabcolsep}{3pt}\renewcommand{\arraystretch}{1.15}
{\begin{tabular}{@{}>{\raggedright\arraybackslash}p{0.17\linewidth}>{\raggedright\arraybackslash}p{0.25\linewidth}>{\raggedright\arraybackslash}p{0.24\linewidth}>{\raggedright\arraybackslash}p{0.29\linewidth}@{}}
\toprule
Method & Conditioning and Task & Prediction Target & Learned Components \\
\midrule
VGGT-World \citep{sun2026vggtworld} \venue{ECCV'26} & Video history for future geometry forecasting & Future geometry tokens, decoded into depth and point maps & Temporal flow transformer; VGGT encoder, remaining blocks, and 3D heads frozen \\
\addlinespace
JointDiT \citep{kwon2025jointdit} \venue{ICCV'25} & Text, with optional RGB or depth for conditional generation & Joint or conditional RGB/depth latents & Depth-branch LoRA and cross-branch modules; pretrained FLUX backbone frozen \\
\addlinespace
Modality Forcing \citep{duisterhof2026modalityforcing} \venue{arXiv'26} & Text, with optional RGB or depth for conditional generation & RGB latents and pixel-space depth & Post-trained DiT with modality-specific modules and noise levels; supports sparse depth supervision \\
\addlinespace
\rowcolor{ForestGreen!12}
\textbf{D2R (Ours)} & Background, reference, and 2D compositing mask for compositing & Completed-scene encoder features, decoded into composite depth & Reference-conditioned injector; depth encoder and decoder frozen, followed by renderer LoRA \\
\bottomrule
\end{tabular}}}
\end{table}

\noindent{\textbf{Feature Injection During RGB Sampling.} Plug-and-Play Diffusion Features~\citep{tumanyan2023plugandplay} preserves a guidance image's structure by injecting its spatial features and self-attention information during diffusion sampling, without training. D2R learns residual corrections inside a frozen depth estimator, using encoder features of paired completed scenes as targets. Feature injection is therefore a shared mechanism, while the learned prediction target and the network receiving the corrections differ.}

\noindent\textbf{Predicting and Aligning Representations.} I-JEPA~\citep{assran2023ijepa} predicts target-region representations from image context, while REPA~\citep{yu2025repa} aligns diffusion features with representations from a pretrained visual encoder. These objectives provide precedents for feature-level supervision. D2R instead targets the depth-encoder features of a paired completed scene, predicting them from the background, reference, and mask before decoding them with the same estimator's unchanged decoder. Its supervision acts inside the depth predictor rather than aligning the RGB generator's hidden states.

\noindent\textbf{Encoder Adaptation and RGB Conditioning.} ViT-Adapter~\citep{chen2023vitadapter} and ViT-CoMer~\citep{xia2024vitcomer} augment visual encoders for dense prediction. D2R complements these architectural mechanisms by targeting features of the completed scene while leaving the dense decoder~\citep{ranftl2021vision} unchanged. Rendering conditions can enter through dedicated branches~\citep{zhang2023adding,mou2024t2iadapter,zhao2023unicontrolnet,ju2024brushnet}, canvas patches~\citep{tan2025ominicontrol}, or image-embedding adapters~\citep{ye2023ipadapter}. D2R supplies its predicted depth as a fixed patch on the reference-guided inpainting canvas~\citep{song2026insert} and trains only renderer LoRA. Merged LoRA adds no renderer parameters.

\noindent{\textbf{Consistency Feedback for Conditional Generation.} ControlNet++~\citep{li2024controlnetpp} trains conditional generation by comparing an input control with the condition recovered from a one-step denoised image through a frozen discriminative model. D2R uses a related principle for RGB-depth consistency. Its supervision remains the paired target depth $D$, including when Stage~1 supplies the conditioning map $D_{\mathrm{in}}$, and it combines this loss with depth-patch denoising. This separates the accuracy of the training target from errors in the supplied condition (App.~\ref{sec:appx_stage2}).}

\section{Method Details}
\label{sec:appx_method}

We detail the objectives and algorithms underlying Sec.~\ref{sec:method}. Throughout, $I_{\mathrm{bg}}$, $I_{\mathrm{ref}}$, and $M$ denote the background, mask-selected reference, and 2D editable-region mask; $I$ and $\hat I$ denote the paired ground-truth and generated composites. The estimator-derived target is $D=F(I)$, and Stage~1 predicts $\hat D=\Phi_\phi(I_{\mathrm{bg}},I_{\mathrm{ref}},M)$. Neither depth is sensor-measured. Roman $E,G$ denote the depth encoder and decoder, whereas $\mathcal E,\mathcal D$ denote the RGB autoencoder. We use $\odot$ for elementwise multiplication and $\mathbf I$ for identity covariance.

\subsection{Notation and Denoising Objective}
\label{sec:appx_notation}

A frozen autoencoder $(\mathcal E,\mathcal D)$ provides latent codes for inpainting~\citep{rombach2022high}. During training, $x_0=\mathcal E(C_{\mathrm{sup}})$ encodes the supervision canvas containing the ground-truth composite, reference, and target depth. A separate input canvas $C_{\mathrm{in}}$ supplies the masked background, reference, and conditioning depth. Algorithm~\ref{alg:train} defines both explicitly. One noise tensor and timestep are sampled for the entire supervision canvas:
\begin{equation}
\begin{split}
x_t &= a(t)\,x_0 + b(t)\,\epsilon, \qquad \epsilon \sim \mathcal{N}(0, \mathbf I), \\
\bigl(a(t),\, b(t)\bigr) &= (1 - t,\; t) \quad \text{or} \quad \bigl(\sqrt{\bar{\alpha}_t},\; \sqrt{1 - \bar{\alpha}_t}\bigr),
\end{split}
\label{eq:interp}
\end{equation}
The first pair is the rectified-flow interpolant of FLUX.1-Fill and Qwen-Inpainting~\citep{wu2025qwenimage}\footnote{Official inpainting checkpoint: \href{https://huggingface.co/InstantX/Qwen-Image-ControlNet-Inpainting}{Qwen-Image-ControlNet-Inpainting}.}, following flow-based generation~\citep{lipman2022flow,liu2023flow,esser2024scaling}. The second is the variance-preserving schedule of SD2-Inpainting~\citep{rombach2022high} and SDXL-Inpainting~\citep{podell2024sdxl}, where $\bar\alpha_t$ is the cumulative signal coefficient. Normalized time runs from clean data ($t=0$) toward noise. For discrete schedules, $t=k/K$ with $k\in\{0,\ldots,K-1\}$ and $K=1000$; coefficients are read at $k$. Training samples $\operatorname{logit}(t)\sim\mathcal N(0,1)$ for rectified flow; variance-preserving backbones sample $k$ uniformly.

Under either corruption schedule, we use $\epsilon_\theta$ for the denoiser. Given $x_t$, $t$, and conditioning $c$, it predicts target $u$. For a set $\Omega'$ of latent positions,
\begin{equation}
\mathcal{L}_{\mathrm{denoise}}(\Omega') = \mathbb{E}_{x_0, \epsilon, t}\, \frac{1}{N_{\mathrm{ch}}\,|\Omega'|} \sum_{i \in \Omega'} \bigl\| \epsilon_\theta(x_t, t, c)_i - u_i \bigr\|_2^2,
\label{eq:denoise}
\end{equation}
where $i$ indexes spatial latent positions, $N_{\mathrm{ch}}$ is the channel count, and the squared Euclidean norm sums over channels. The target is velocity, $u=\epsilon-x_0$, for rectified flow and noise, $u=\epsilon$, for variance-preserving diffusion. The one-step clean-latent estimate is
\begin{equation}
\hat{x}_0 =
\begin{cases}
x_t - t\,\epsilon_\theta(x_t, t, c), & u = \epsilon - x_0,\\
\bigl(x_t - b(t)\,\epsilon_\theta(x_t, t, c)\bigr) / a(t), & u = \epsilon,
\end{cases}
\label{eq:x0hat}
\end{equation}
Equation~(\ref{eq:onestep}) specializes the first case to the deployed renderer. Both denoising losses use Eq.~(\ref{eq:denoise}); they differ only in spatial support. The consistency loss also depends on this clean-latent reconstruction and the normalized timestep weight $w(t)$.

We write $c=\operatorname{Cond}(C,m,I_{\mathrm{ref}})$ for the backbone-specific inpainting condition constructed from input canvas $C$, latent mask $m$, and reference. The binary mask marks generated positions with one. This condition includes the encoded canvas and mask, Redux reference embeddings for FLUX.1-Fill, and the ControlNet path for Qwen-Inpainting~\citep{wu2025qwenimage}; tensor construction follows each backbone's released interface. Training corrupts the entire supervision canvas $C_{\mathrm{sup}}$ while retaining the separate input condition. At inference, the reference, depth, and unmasked background remain observed inputs and are restored after every update (Eq.~(\ref{eq:clamp})). Rectified-flow sampling uses explicit Euler updates; variance-preserving backbones use their released deterministic multistep solvers. Neither adds noise after initialization.

\subsection{Depth Injector: Architecture and Objective}
\label{sec:appx_stage1}

\subsubsection{Encoder Corrections and Capacity}
\label{sec:appx_injector_arch}

The frozen model $F=G\circ E$ consists of encoder blocks $B_1,\ldots,B_N$ and dense decoder $G$~\citep{ranftl2021vision}. The set $\mathcal K$ contains the encoder taps read by the decoder. The initial sequence $\tilde h^0_{\mathrm{bg}}$ contains the masked-background tokens entering the first block. At tap $\ell$, $h^\ell_{\mathrm{bg}}=B_\ell(\tilde h^{\ell-1}_{\mathrm{bg}})$ denotes features before the current correction, including corrections propagated from earlier taps. Reference features $h^\ell_{\mathrm{ref}}$ and training-only target features $h^\ell_{\mathrm{gt}}$ come from the reference and complete composite, respectively. Mask-conditioned background queries attend to the background and reference sequences, yielding the correction
\begin{equation}
\delta^\ell = W^\ell_{\mathrm{out}}\bigl(\alpha^\ell \odot \mathrm{Attn}(Q, K, V)\bigr), \qquad Q = W_q\bigl(h^\ell_{\mathrm{bg}} + e(M)\bigr), \qquad (K,V) = W_{kv}\bigl[h^\ell_{\mathrm{bg}} ; h^\ell_{\mathrm{ref}}\bigr],
\label{eq:injector}
\end{equation}
Here $Q,K,V$ are attention queries, keys, and values, and $\mathrm{Attn}$ denotes multihead attention. The input to $W_{kv}$ concatenates background and reference token sequences; the projection produces separate keys and values. Layer indices on $Q,K,V,W_q,W_{kv}$ are suppressed. The embedding $e(M)$ places the mask on the token grid, $\alpha^\ell$ is a learned gate, and the $W$ matrices are linear projections. The zero-initialized output projection makes all corrections initially vanish, recovering the frozen estimator on the same masked-background input. Each tap has independent injector parameters; their union $\phi$ is the only trainable part of Stage~1. Subsequent frozen blocks propagate each correction to the decoder taps above it.

Width $w$ is the internal projection dimension in Eq.~(\ref{eq:injector}). For Depth Anything V2 ViT-L~\citep{yang2024depthv2}, projections map the 1024-channel encoder features to $w$ channels and back. Each tap has seven attention blocks with eight heads, with separate parameters at encoder blocks 5, 12, 18, and 24. Increasing $w$ changes injector capacity without changing the frozen encoder dimensions; parameters grow approximately quadratically with $w$.

\subsubsection{Injector Width and Cost}
\label{sec:appx_injector_width}

The adopted injector uses width $w=640$. To assess the accuracy and cost of different widths, Tab.~\ref{tab:width} varies only $w$ while fixing the taps, attention blocks, heads, gate, data, optimizer, and batch size. All trained variants use final checkpoints after 50K training iterations. Frozen denotes Depth Anything V2 ViT-L without an injector. Full-frame AbsRel is measured before RGB synthesis, and Avg is the mean of ID and OOD. Params counts trainable injector parameters; Vs.\ Frozen gives their size relative to the frozen backbone. Inference cost is measured on one RTX PRO 6000 (96\,GB) in bf16 at batch size one with $518\times518$ inputs.

Width 640 reaches ID and OOD AbsRel of 0.0462 and 0.0751 with 272.4M trainable parameters. Increasing width to 896 lowers ID error to 0.0445 but raises OOD error to 0.0755, while parameters increase to 529.2M. Latency rises from 56.5 to 67.6\,ms and inference memory from 5.21 to 6.54\,GB. Thus the additional capacity brings a small ID gain without improving OOD accuracy, supporting width 640 as the adopted balance of accuracy and cost.

\begin{table}[t]
\centering
\caption{Depth Injector Accuracy and Cost Across Widths.}
\label{tab:width}
\fitwidth{%
\begin{tabular}{lccccccc}
\toprule
\multirow{2}{*}{Width $w$} & \multirow{2}{*}{Params} & \multirow{2}{*}{Vs. Frozen} & \multirow{2}{*}{Latency $\downarrow$} & \multirow{2}{*}{VRAM $\downarrow$} & \multicolumn{2}{c}{AbsRel $\downarrow$} & \multirow{2}{*}{Avg $\downarrow$} \\
\cmidrule(lr){6-7}
 & & & & & In-Distribution & Out-of-Distribution & \\
\midrule
Frozen & 0 & --- & 25.8\,ms & 3.25\,GB & 0.1168 & 0.1506 & 0.1337 \\
\midrule
224 & 35.2M & 10.5\% & 42.4\,ms & 3.50\,GB & 0.0507 & 0.0776 & 0.0642 \\
320 & 70.1M & 20.9\% & 45.5\,ms & 3.72\,GB & 0.0463 & 0.0767 & 0.0615 \\
384 & 100.0M & 29.8\% & 45.8\,ms & 3.91\,GB & 0.0473 & 0.0765 & 0.0619 \\
448 & 135.2M & 40.3\% & 47.6\,ms & 4.26\,GB & 0.0466 & 0.0770 & 0.0618 \\
512 & 175.6M & 52.4\% & 48.9\,ms & 4.37\,GB & 0.0471 & 0.0770 & 0.0621 \\
\rowcolor{ForestGreen!12} \textbf{640 (Adopted)} & \textbf{272.4M} & \textbf{81.2\%} & \textbf{56.5\,ms} & \textbf{5.21\,GB} & \textbf{0.0462} & \textbf{0.0751} & \textbf{0.0606} \\
768 & 390.2M & 116.4\% & 59.5\,ms & 5.87\,GB & 0.0463 & 0.0778 & 0.0621 \\
896 & 529.2M & 157.8\% & 67.6\,ms & 6.54\,GB & 0.0445 & 0.0755 & 0.0600 \\
\bottomrule
\end{tabular}}
\end{table}

\subsubsection{Feature Supervision and Structural Ablations}
\label{sec:appx_injector_ablations}

The paired composite's encoder features provide a target that the frozen decoder already maps to composite depth. Feature supervision asks the injector to recover this representation from the background and reference, while retaining the pretrained feature-to-depth mapping. Decoded-depth supervision constrains only the resulting map: different feature configurations can produce similar depth, so a close output match need not recover the paired composite's encoder representation.

Equation~(\ref{eq:stage1}) compares corrected background features with detached composite features at each tap. Features have shape $B\times C\times p_h\times p_w$, where $B$ is batch size, $C$ is channel count, and $p_h,p_w$ are the spatial grid dimensions. For each sample $b$, $\sigma_b^\ell=\operatorname{std}(h^\ell_{\mathrm{gt},b})$ reduces the channel and both spatial dimensions to one scalar. The mean absolute feature error uses those same dimensions and is divided by $\mathrm{sg}(\sigma_b^\ell)+10^{-6}$. We average the normalized errors first over samples and then over decoder taps. The frozen decoder produces $\hat D$ without participating in feature-loss backpropagation.

For one sample, let $H^*=\{h^\ell_{\mathrm{gt}}\}_{\ell\in\mathcal K}$ and $\hat H=\{\tilde h^\ell_{\mathrm{bg}}\}_{\ell\in\mathcal K}$ be the target and corrected feature collections supplied to $G$. Treating each collection and depth map as a vector, assume that $G$ is $L$-Lipschitz in the $\ell_1$ norm on a region containing both collections. Then
\[
\bigl\|G(\hat H)-G(H^*)\bigr\|_1\leq L\,\bigl\|\hat H-H^*\bigr\|_1.
\]
This conditional relation motivates matching features that the frozen decoder can already interpret. We do not estimate the Lipschitz constant, and Eq.~(\ref{eq:stage1}) averages per-tap, sample-normalized errors rather than the unnormalized norm above. It therefore yields no numerical bound on test depth error or local contact accuracy and does not establish generalization. Performance across estimators and held-out categories is assessed empirically (Tab.~\ref{tab:depth_full}).

D2R applies residual corrections across the full token grid. The masked-token ablation restricts these corrections to the editable region to test whether object-local updates suffice (Sec.~\ref{sec:ablation}). Relative-depth prediction depends on the surrounding scene, so one possible explanation for the full-grid advantage is that object and background features must adapt together. This concerns their representation of relative depth; it does not imply that insertion physically moves unchanged background surfaces.

To test how scene and reference information enter the correction, we compare attention and gating variants with the frozen estimator and feature objective held fixed. Joint attention lets each background token access both scene context and reference features, following multimodal transformer designs~\citep{wu2025qwenimage}. Relative to reference-only cross-attention, it lowers full-frame AbsRel by approximately 2\% on ID and 1\% on OOD at each variant's final checkpoint. The gate and zero-initialized projection follow related conditioning mechanisms~\citep{qiu2025gated,zhang2023adding,li2023gligen,alayrac2022flamingo}. Adding the gate lowers error by a further 1\% and 3\%, respectively; the ungated variant also varies more across checkpoints.

\subsubsection{Feature Targets Versus Decoded-Depth Targets}
\label{sec:appx_supervision_comparison}

To isolate the effect of the supervision target, we start from D2R's width-640 injector on frozen Depth Anything V2 ViT-L and replace only the encoder-feature objective with a decoded-depth objective. Both variants retain the same injector architecture, frozen encoder and decoder, training data, optimizer, and batch size, and use final checkpoints after 50K training iterations (Tab.~\ref{tab:supervision_final}). The alternative uses the normalized $\ell_1$ depth discrepancy from the Stage-2 consistency objective (App.~\ref{sec:appx_stage2}): the same normalization $A$ is applied separately to predicted and target composite-depth maps, without a background-based scale-and-shift fit. We score full-frame composite depth before RGB synthesis on both paired splits.

The two objectives differ in where gradients enter the frozen estimator. Feature supervision compares decoder inputs directly. Decoded-depth supervision must propagate gradients through the decoder to the injector, retaining decoder activations for backpropagation even though the decoder weights remain fixed. Both variants deploy the same injector and depth estimator, so this difference affects training computation rather than inference cost.

Encoder-feature supervision lowers ID and OOD AbsRel by 30.9\% and 31.4\%, respectively, relative to decoded-depth-only supervision; OOD error falls from 0.1094 to 0.0751. With architecture and training iterations held fixed, the comparison supports feature targets for repurposing this frozen estimator. The gain also avoids the decoder backward pass required by the alternative. Table~\ref{tab:supervision_regions_final} reports the decoded-depth-only variant's regional errors, using the regions defined in App.~\ref{sec:appx_paired_uncertainty}.

\begin{table}[ht]
\centering
\caption{Regional AbsRel with Decoded-Depth Supervision.}
\label{tab:supervision_regions_final}
{\setlength{\tabcolsep}{16pt}%
\begin{tabular}{lccc}
\toprule
Split & Interior & Boundary Band & Background \\
\midrule
ID & 0.1081 & 0.2072 & 0.0522 \\
OOD & 0.1711 & 0.2422 & 0.0909 \\
\bottomrule
\end{tabular}}
\end{table}

A separate ablation retains the feature objective and adds decoded-depth supervision to test whether the two targets are complementary. Relative to feature-only training, the combined objective lowers ID full-frame AbsRel by about 5\% but raises OOD error by about 3\%. Retaining decoder activations for backpropagation also reduces the largest single-GPU batch by about 30\%. The ID/OOD tradeoff and memory cost favor feature-only training in this setting.

\noindent{\textbf{Effect on RGB Geometry with the Same Renderer.} We test whether the Stage-1 supervision difference remains useful after RGB synthesis. The two existing Stage-1 checkpoints in Tab.~\ref{tab:supervision_final} supply depth to the same trained FLUX.1-Fill renderer, without further training. Only the depth source changes: both conditions use the same background, reference, and mask for all 139 OOD cases, with identical sampling settings and the standard depth normalization, resizing, quantization, and canvas layout.}

\begin{table}[ht]
\centering
\caption{{From Stage-1 Supervision to Final RGB Geometry with the Same Renderer.}}
\label{tab:supervision_rgb}
\fitwidth{%
{\begin{tabular}{lcc}
\toprule
\multirow{2}{*}{Stage-1 Supervision} & \multicolumn{2}{c}{OOD AbsRel $\downarrow$} \\
\cmidrule(lr){2-3}
 & Predicted Composite Depth & Rendered RGB Geometry \\
\midrule
Decoded Depth Only & 0.1094 & 0.089 \\
\rowcolor{ForestGreen!12}
Encoder Feature Only (D2R) & \textbf{0.0751} & \textbf{0.067} \\
\bottomrule
\end{tabular}}}
\end{table}

{Table~\ref{tab:supervision_rgb} connects the predictor comparison to final compositing: the feature-supervised depth yields lower rendered-image AbsRel under the same RGB renderer. Thus, its benefit extends beyond the intermediate depth map to geometric agreement in the generated image. This comparison measures the utility of each predictor for the deployed renderer, which was trained with the default depth-supply strategy. It does not compare renderers retrained for each predictor or separate geometric accuracy from compatibility with the renderer's training distribution.}

\subsubsection{Backbone Interfaces and Parameter Sharing}
\label{sec:appx_injector_interfaces}

Depth Anything V1~\citep{yang2024depthanything}, V2~\citep{yang2024depthv2}, and V3~\citep{lin2026depth3} expose four decoder taps on one token grid. Metric-Anything~\citep{ma2026metricanything} uses a patch pyramid, requiring corrections to be broadcast before decoding. InfiniDepth~\citep{yu2026infinidepth} requires taps across token resolutions, while Depth Pro~\citep{bochkovskii2025depthpro} receives corrections in its image encoder alongside a separate patch encoder. The design and objective are shared, with tap selection, mask placement, and feature dimensions adapted per backbone.

The weight-sharing ablation replaces the default backbone's independent tap-specific injectors with shared parameters, keeping width 640 and the other settings fixed (Sec.~\ref{sec:ablation}). Sharing reduces parameter count but executes the same number of attention blocks. The adopted design uses independent parameters so each tap can learn a correction suited to its encoder layer.

\subsection{Depth-Conditioned Rendering: Training and Inference}
\label{sec:appx_stage2}

\subsubsection{Canvas and Fixed Observations}
\label{sec:appx_canvas}

{We use the dataset-provided reference-object masks to select the source object and the target masks to specify the compositing region. AnyInsertion++ supplies masks that follow the target object's silhouette, whereas the unpaired benchmarks are evaluated with coarse box-shaped target masks. The evaluations therefore cover both detailed object masks and coarse region masks. Because the datasets differ, these results do not isolate the effect of mask precision.}

Let $S$ be the composite patch's side length. The input canvas places the masked background beside a column containing the reference above the conditioning depth:
\begin{equation}
C = \bigl[\, I_{\mathrm{bg}} \odot (1 - M) \;\big|\; (I_{\mathrm{ref}} ; D_{\mathrm{in}}) \,\bigr] \in \mathbb{R}^{S \times 1.5S \times 3}, \qquad \Omega = \Omega_{\mathrm{comp}} \cup \Omega_{\mathrm{ref}} \cup \Omega_{\mathrm{depth}},
\label{eq:canvas}
\end{equation}
All evaluated compositing datasets provide a reference object mask that identifies which object to use among other objects and background pixels. We use this supplied mask to select the reference; no automatic reference selection is performed. The mask identifies source appearance, whereas $M$ marks the editable region in the target scene. The reference is tightly cropped to its object mask, centered on a white square whose side is 1.05 times the crop's longer side, and resized to its patch. Training and inference use identical padding.

In Eq.~(\ref{eq:canvas}), the vertical bar places patches side by side, while the semicolon stacks the reference above the depth. For FLUX.1-Fill, $S=1024$ gives a $1024\times1536$ canvas with $S/2\times S/2$ auxiliary patches; resizing and padding to these sizes are implicit. The single-channel relative depth $D_{\mathrm{in}}$ is replicated across three channels. The sets $\Omega_{\mathrm{comp}}$, $\Omega_{\mathrm{ref}}$, and $\Omega_{\mathrm{depth}}$ index the corresponding latent positions. The same labels identify image patches after decoding the canvas.

At inference, the latent mask $m$ maps $M$ to the composite patch's latent grid, with one marking generated positions and zero on the reference, depth, and unmasked background. Writing $x$ for the current sampled latent and $x_{\mathrm{obs}}=\mathcal E(C)$ for the encoded input canvas, the sampler restores observed positions after every update:
\begin{equation}
x \odot (1 - m) = x_{\mathrm{obs}} \odot (1 - m) \quad \text{after every step of Alg.~\ref{alg:inference}},
\label{eq:clamp}
\end{equation}

\subsubsection{Training Objectives and Depth Selection}
\label{sec:appx_render_objectives}

Stage~2 uses three equally weighted losses. Over valid latent positions $\mathcal V$ excluding square padding, the two denoising losses read separate parts of one prediction:
\begin{equation}
\begin{split}
\mathcal{L}_{\mathrm{total}} &= \mathcal{L}_{\mathrm{pixel}} + \mathcal{L}_{\mathrm{depth}} + \mathcal{L}_{\mathrm{cons}}, \\
\mathcal{L}_{\mathrm{pixel}} &= \mathcal{L}_{\mathrm{denoise}}\bigl((\Omega_{\mathrm{comp}} \cup \Omega_{\mathrm{ref}}) \cap \mathcal{V}\bigr), \qquad \mathcal{L}_{\mathrm{depth}} = \mathcal{L}_{\mathrm{denoise}}\bigl(\Omega_{\mathrm{depth}} \cap \mathcal{V}\bigr),
\end{split}
\label{eq:terms}
\end{equation}
The two denoising losses supervise different patches of the same canvas prediction. Its depth target is always $D$, regardless of which depth map conditions the input. The auxiliary depth loss trains the shared denoiser to represent composite geometry. At inference, each sampling step restores the supplied depth condition, discarding the predicted depth-patch values.

To enforce RGB-to-depth consistency, the rectified-flow prediction also yields a one-step clean estimate and its decoded composite patch:
\begin{equation}
\hat{x}_0 = x_t - t\,\epsilon_\theta(x_t, t, c), \qquad \hat{I}_{\mathrm{comp}} = \mathcal{D}(\hat{x}_0)\big|_{\Omega_{\mathrm{comp}}},
\label{eq:onestep}
\end{equation}
The full consistency loss is $\mathcal L_{\mathrm{cons}}=w(t)\|A(F(\hat I_{\mathrm{comp}}))-\mathrm{sg}(A(D))\|_1$. The target remains $D$ even when the input depth is imperfect, and $w(t)=1-t$ reduces the contribution of noisy early one-step estimates. Gradients flow through the frozen RGB decoder $\mathcal D$ and depth estimator $F$ to LoRA only.

Normalization $A$ is applied independently to each predicted and target depth map. For each sample, map $Z$ is flattened over the entire composite patch, including padding; the reference and depth-conditioning patches are excluded. We compute $\mu(Z)=\operatorname{median}(Z)$ and $r(Z)=\operatorname{mean}|Z-\mu(Z)|$, then normalize as $A(Z)=(Z-\mu(Z))/\max(r(Z),10^{-6})$. The denominator floor handles nearly constant depth. Only after normalization do we apply the validity mask when averaging the loss; it does not affect the median or mean absolute deviation. Evaluation instead fits the background alignment in Eq.~(\ref{eq:align}).

To expose the renderer to prediction errors while keeping supervision fixed, we select the conditioning depth independently for each sample and iteration:
\begin{equation}
D_{\mathrm{in}} = \begin{cases} \hat{D} & \text{with probability } 1/2, \\ D & \text{with probability } 1/2, \end{cases}
\label{eq:supply}
\end{equation}
The adopted selection policy changes the conditioning input while retaining $D$ in the supervision canvas. To test other ways of exposing the renderer to depth, the blending ablation instead supplies $D_{\mathrm{in}}=\lambda\hat D+(1-\lambda)D$, with $\lambda\sim\operatorname{Beta}(0.2,0.2)$ following mixup~\citep{zhang2018mixup}; the single-source ablations use $D_{\mathrm{in}}=D$ (GT-only) or $D_{\mathrm{in}}=\hat D$ (predicted-only). Selection supplies one intact map, whereas blending interpolates between the maps. When conditioning differs from $D$, simply reproducing the input depth need not minimize the supervised depth discrepancy. Stage-1 predictions are exported before Stage-2 training, so only the frozen consistency estimator runs inside its training loop. Appendix~\ref{sec:appx_s2} reports the comparison.

\subsubsection{Training and Inference Algorithms}
\label{sec:appx_algorithms}

Algorithm~\ref{alg:inference} specifies rectified-flow inference, and Alg.~\ref{alg:train} gives one training step using the same $\operatorname{Cond}$ notation. Stage~1 predicts $\hat D$ once; Stage~2 samples the masked composite region for $T$ steps and resizes the decoded composite to the background resolution. FLUX.1-Fill receives the reference through Redux; other backbones use the prompts in App.~\ref{sec:appx_s2}. LoRA is merged before deployment.

\begin{algorithm}[t]
\caption{D2R Inference with Fixed Composite Depth.}
\label{alg:inference}
\begin{algorithmic}[1]
\Require background $I_{\mathrm{bg}}$, mask-selected reference $I_{\mathrm{ref}}$, 2D editable-region mask $M$, frozen depth model $F = G \circ E$ with trained injector $\phi$, generator $\epsilon_\theta$ with merged LoRA, frozen autoencoder $(\mathcal{E}, \mathcal{D})$, timesteps $1 = t_1 > t_2 > \dots > t_{T+1} = 0$
\Ensure composite $\hat{I}$ at the resolution of $I_{\mathrm{bg}}$
\State $\hat{D} \gets G\bigl(\{\tilde{h}^\ell_{\mathrm{bg}}\}_{\ell \in \mathcal{K}}\bigr)$ with $\tilde{h}^\ell_{\mathrm{bg}}$ from Eq.~(\ref{eq:stage1}) \Comment{Stage 1, one forward pass}
\State $C \gets \bigl[\, I_{\mathrm{bg}} \odot (1 - M) \;\big|\; (I_{\mathrm{ref}} ; \hat{D}) \,\bigr]$ \Comment{canvas, Eq.~(\ref{eq:canvas})}
\State Construct latent mask $m$ from $M$ on $\Omega_{\mathrm{comp}}$, with zeros on the other patches
\State $x_{\mathrm{obs}} \gets \mathcal{E}(C)$, \quad $c \gets \operatorname{Cond}(C,m,I_{\mathrm{ref}})$
\State $x \gets \epsilon$ with $\epsilon \sim \mathcal{N}(0, \mathbf I)$
\For{$k = 1, \dots, T$}
\State $x \gets x + (t_{k+1} - t_k)\, \epsilon_\theta(x, t_k, c)$ \Comment{one rectified-flow step on the whole canvas}
\State $x \gets m \odot x + (1 - m) \odot x_{\mathrm{obs}}$ \Comment{clamp, Eq.~(\ref{eq:clamp})}
\EndFor
\State $\hat{C} \gets \mathcal{D}(x)$
\State \Return $\hat{I} \gets \hat{C}\big|_{\Omega_{\mathrm{comp}}}$ resized to the resolution of $I_{\mathrm{bg}}$
\end{algorithmic}
\end{algorithm}

\begin{algorithm}[t]
\caption{Stage-2 Training with RGB and Depth Supervision.}
\label{alg:train}
\begin{algorithmic}[1]
\Require paired sample $(I_{\mathrm{bg}}, I_{\mathrm{ref}}, M, I)$, target depth $D = F(I)$, exported Stage-1 depth $\hat D$, frozen $\mathcal E$, $\mathcal D$, $F$, denoiser $\epsilon_\theta$ with trainable LoRA $\theta$, configured optimizer
\State $D_{\mathrm{in}} \gets \hat{D}$ with probability $0.5$ and $D$ otherwise \Comment{per-sample selection, Eq.~(\ref{eq:supply})}
\State $C_{\mathrm{in}} \gets \bigl[\, I_{\mathrm{bg}} \odot (1 - M) \;\big|\; (I_{\mathrm{ref}} ; D_{\mathrm{in}}) \,\bigr]$ \Comment{input canvas}
\State $C_{\mathrm{sup}} \gets \bigl[\, I \;\big|\; (I_{\mathrm{ref}} ; D) \,\bigr]$ \Comment{supervision canvas, depth patch always $D$}
\State Construct latent mask $m$ from $M$ on $\Omega_{\mathrm{comp}}$, with zeros elsewhere
\State $c \gets \operatorname{Cond}(C_{\mathrm{in}},m,I_{\mathrm{ref}})$, \quad $x_0 \gets \mathcal{E}(C_{\mathrm{sup}})$
\State $\epsilon \sim \mathcal{N}(0, \mathbf I)$, \quad $\mathrm{logit}(t) \sim \mathcal{N}(0, 1)$, \quad $x_t \gets (1 - t)\, x_0 + t\, \epsilon$ \Comment{rectified flow, Eq.~(\ref{eq:interp})}
\State $v \gets \epsilon_\theta(x_t, t, c)$ \Comment{one forward pass over the whole canvas}
\State $\mathcal{L}_{\mathrm{pixel}} \gets \| v - (\epsilon - x_0) \|_2^2$ averaged on $(\Omega_{\mathrm{comp}} \cup \Omega_{\mathrm{ref}}) \cap \mathcal{V}$ \Comment{Eq.~(\ref{eq:terms})}
\State $\mathcal{L}_{\mathrm{depth}} \gets \| v - (\epsilon - x_0) \|_2^2$ averaged on $\Omega_{\mathrm{depth}} \cap \mathcal{V}$
\State $\hat{x}_0 \gets x_t - t\, v$, \quad $\hat{I}_{\mathrm{comp}} \gets \mathcal{D}(\hat{x}_0)\big|_{\Omega_{\mathrm{comp}}}$ \Comment{Eq.~(\ref{eq:onestep})}
\State $\mathcal{L}_{\mathrm{cons}} \gets (1 - t)\, \bigl\| A(F(\hat{I}_{\mathrm{comp}})) - \mathrm{sg}(A(D)) \bigr\|_1$ \Comment{Eq.~(\ref{eq:cons}), full form of App.~\ref{sec:appx_stage2}}
\State Update $\theta$ with the configured optimizer using $\nabla_\theta\mathcal L_{\mathrm{total}}$ \Comment{LoRA only; $\mathcal E,\mathcal D,F$ frozen}
\end{algorithmic}
\end{algorithm}

\subsection{Computational Cost}
\label{sec:appx_cost}

Stage~1 adds a 272.4M-parameter injector to frozen Depth Anything V2 ViT-L~\citep{yang2024depthv2} (335.3M parameters), an 81.2\% increase. Stage~2 trains rank-256 LoRA with scaling 256 while freezing the backbone, autoencoder, and Redux encoder. Adapter sizes are 927M for FLUX.1-Fill, 53M for SD2-Inpainting~\citep{rombach2022high}, 372M for SDXL-Inpainting~\citep{podell2024sdxl}, and 1{,}887M for Qwen-Inpainting~\citep{wu2025qwenimage}. Only the injector adds deployed parameters; renderer LoRA is merged into the backbone.

We measure inference on one NVIDIA RTX PRO 6000 (96\,GB), at batch size one on one OOD sample. Each row of Tab.~\ref{tab:cost} runs in a separate process with one warm-up and five timed passes. Latency is the median of those passes, excluding disk I/O; memory is peak process allocation including weights and activations. Stage-1 timing covers both the frozen estimator and its injector.

Following each released Diffusers setup, SD2 and SDXL use fp16; the depth models, Qwen, and FLUX use bf16. Stage~1 receives $518\times518$ inputs. All Stage-2 renderers use 28 steps, with canvases of $512\times768$ for SD2, $1024\times1536$ for SDXL and FLUX, and $768\times1152$ for Qwen. Guidance is 7.0 for SD2 and SDXL, 4.0 for Qwen, and distilled guidance 30 for FLUX.

Two-stage latency adds the 56.5\,ms Stage-1 pass to the measured renderer latency and is rounded to two decimal places. Inference memory peaks are reported per process; summing them would not measure the peak of a deployment with both stages resident. The table therefore separates the cost of predicting geometry from that of RGB generation.

The matched Insert-Anything$^*$~\citep{song2026insert} baseline shares D2R's backbone, canvas, Redux embedding, and LoRA configuration. D2R therefore adds one Stage-1 pass to otherwise matched RGB generation cost. This does not compare against released Insert-Anything's full-height reference canvas, whose token count differs.

\begin{table}[t]
\centering
\caption{Inference Cost on RTX PRO 6000.}
\label{tab:cost}
\fitwidth{%
\begin{tabular}{lccccc}
\toprule
Component & Latency $\downarrow$ & \makecell{Two-Stage\\Latency $\downarrow$} & \makecell{Inference\\VRAM $\downarrow$} & \makecell{Canvas\\(Height $\times$ Width)} & \makecell{Trainable\\Parameters} \\
\midrule
Frozen DAv2 \citep{yang2024depthv2} \venue{NeurIPS'24} & 25.8\,ms & --- & 3.25\,GB & $518 \times 518$ & 0 \\
\rowcolor{ForestGreen!12}
+ Depth Injector (Stage 1; Sec.~\ref{sec:stage1}) & 56.5\,ms & --- & 5.21\,GB & $518 \times 518$ & 272.4M \\
\midrule
SD2-Inpainting \citep{rombach2022high} \venue{CVPR'22} (Stage 2) & 0.81\,s & 0.87\,s & 4.39\,GB & $512 \times 768$ & 53M (LoRA) \\
SDXL-Inpainting \citep{podell2024sdxl} \venue{ICLR'24} (Stage 2) & 2.32\,s & 2.38\,s & 13.16\,GB & $1024 \times 1536$ & 372M (LoRA) \\
Qwen-Inpainting \citep{wu2025qwenimage} \venue{arXiv'25} (Stage 2) & 14.24\,s & 14.30\,s & 65.93\,GB & $768 \times 1152$ & 1{,}887M (LoRA) \\
\rowcolor{ForestGreen!12}
FLUX.1-Fill (Stage 2; Sec.~\ref{sec:stage2}) & 22.34\,s & 22.40\,s & 38.50\,GB & $1024 \times 1536$ & 927M (LoRA) \\
\bottomrule
\end{tabular}}
\end{table}

On NVIDIA RTX PRO 6000, the width-640 depth injector reaches a peak training memory of 72.4\,GiB. Stage-2 peaks per device are 81.3\,GiB for SD2, 67.8\,GiB for SDXL, 94.0\,GiB for Qwen, and 66.8\,GiB for FLUX. These device-level readings are converted from MiB to GiB and correspond to the configurations in Apps.~\ref{sec:appx_s1} and~\ref{sec:appx_s2}. Batch size and adapter settings differ across backbones, so these values document the memory needed by each training setup rather than isolate the backbone's memory efficiency.

\section{Experimental Protocols and Further Results}
\label{sec:appx_analysis}

\subsection{Datasets and Evaluation Protocols}
\label{sec:appx_bench}

\subsubsection{Paired Data and Depth Targets}
\label{sec:appx_paired_data}

AnyInsertion++ curates AnyInsertion~\citep{song2026insert} for training and paired evaluation, excluding samples with annotation errors in automatically produced backgrounds or masks. The checks below retain 52{,}918 of the 77{,}745 training samples. Evaluation expands from the original 38 test cases to 287 ID and 139 OOD cases, each with an estimator-derived depth target (Tab.~\ref{tab:anyinsertionpp}). The category split, annotation checks, and depth preparation are described below.

\begin{table}[t]
\centering
\caption{AnyInsertion++ Dataset Design.}
\label{tab:anyinsertionpp}
{\setlength{\tabcolsep}{5pt}\renewcommand{\arraystretch}{1.15}%
\begin{tabular}{>{\raggedright\arraybackslash}p{0.21\textwidth} >{\raggedright\arraybackslash}p{0.26\textwidth} >{\columncolor{ForestGreen!12}\raggedright\arraybackslash}p{0.41\textwidth}}
\toprule
Property & Original AnyInsertion \citep{song2026insert} & \textbf{AnyInsertion++ (Ours)} \\
\midrule
Paired Test Cases & 38 & 287 ID and 139 OOD \\
Held-Out Object Tags & Not separated & 21 tags, OOD split only \\
Training Pool Examined & 77{,}745 before curation & 52{,}918 kept, 24{,}827 removed \\
Ground-Truth Composite Depth & None & One per paired sample \\
Matched Comparator & None & Insert-Anything \citep{song2026insert} retrained on ours \\
\bottomrule
\end{tabular}}
\end{table}

We use Claude Opus 5\footnote{Anthropic, \href{https://www.anthropic.com/news/claude-opus-5}{Introducing Claude Opus 5}.} to check annotation validity for all 77{,}745 training samples with one fixed English prompt. It receives the background, reference image, dataset-provided reference object mask, target mask, and ground-truth composite, in that order. The checks cover target-object presence, target-mask alignment, reference-mask validity, object-instance identity, and agreement between the background and the composite with that object absent. Aesthetics, photographic quality, and difficulty are excluded, so difficult but correctly labelled samples are retained.

The model returns a confidence score from 0 to 10, a usability label, and the first failed check. Human inspection of 300 randomly sampled images per score tier informed the threshold; a sample was judged usable when at least two of three annotators agreed. Excluding scores below 8 removes 24{,}827 samples and retains 52{,}918. Appendix~\ref{sec:appx_filtering_prompt} reproduces the exact prompt.

The category split tests generalization beyond the compositing training set. The corpus contains 109 object tags grouped into categories: the 287-case ID split shares training categories, whereas the 139-case OOD split contains categories absent from training. Both include insertion and replacement. The split construction holds out 21 tags; Tab.~\ref{tab:rigid_tags} lists those with scored samples. A compositing case is the evaluation unit, but related insertion and replacement cases can share an image identifier. This gives 155 ID and 82 OOD image groups, for which App.~\ref{sec:appx_paired_uncertainty} reports image-cluster uncertainty alongside case-level uncertainty.

\noindent\textbf{Train/Test Image Overlap.} We compare all 287 ID and 139 OOD test images against the 52{,}918 training samples. An exact thumbnail match means identical SHA-1 hashes of $64\times64$ grayscale thumbnails. For near duplicates, a 64-bit DCT-based perceptual hash selects candidates with Hamming distance at most 8; normalized cross-correlation of at least 0.95 between the $64\times64$ thumbnails confirms a match. Neither test split contains a thumbnail match or a confirmed near-duplicate image under these criteria (0/287 ID; 0/139 OOD).

Ground-truth composite depth is the estimator-derived target $D=F(I)$, not sensor ground truth. The default frozen Depth Anything V2 ViT-L~\citep{yang2024depthv2} runs offline at 518-pixel input resolution. Its relative inverse depth, with larger values denoting nearer surfaces, is independently min-max normalized to $[0,1]$, resized to the original image, and stored as an 8-bit single-channel PNG. These maps supply the Stage-2 target patch and default geometric reference; Stage~1 instead learns from the same estimator's encoder features. Background and reference depth maps for the alternative provisioning strategies use the same extraction procedure.

At inference, Stage-1 predictions $\hat D$ use only the three compositing inputs. They undergo the same normalization and are rounded from a 16-bit intermediate to 8-bit PNGs for the depth patch, including on unpaired evaluation data. Thus \emph{GT Composite Depth} denotes the paired-image target, while deployed conditioning uses predicted composite depth. Scorer dependence is tested in App.~\ref{sec:appx_metrics}.

To test whether improvements extend beyond rigid objects, we group garment, accessory, person, animal, food, and plush tags as non-rigid and the remaining tags as rigid. This operational rule groups categories rather than assigning physical rigidity to every instance. Table~\ref{tab:rigid_tags} lists 146 rigid and 141 non-rigid ID cases from 42 and 45 tags, and 119 rigid and 20 non-rigid OOD cases from 14 and 6 tags. The OOD result is therefore dominated by rigid objects, with a smaller non-rigid sample.

D2R lowers AbsRel in all four groups (Tab.~\ref{tab:rigid}). Replacing predicted depth with depth from the paired composite further tests the benefit of a better condition. That comparison keeps the renderer fixed; it does not uniquely attribute the remaining error to Stage~1.

\begin{table}[t]
\centering
\caption{\textbf{Object Tags of the Two Test Splits by Rigidity Group.} Bands (a) and (b) list ID and OOD tags with scored-sample counts. Groups follow the category rule in the text.}
\label{tab:rigid_tags}
\fitwidth{\setlength{\tabcolsep}{4pt}%
\begin{tabular}{>{\raggedright\arraybackslash}p{0.17\textwidth} p{0.74\textwidth}}
\toprule
Group & Tags (Count) \\
\midrule
\multicolumn{2}{l}{\textbf{(a) In-Distribution (ID) Test Split: Categories Overlap the Training Set}} \\
Rigid ($n = 146$, 42 tags) & \footnotesize ashcan (4), balloons (2), bed (2), book (4), bottle (4), box (4), cabinet\_air\_conditioner (4), ceiling\_lamp (2), chair (2), coat\_rack (2), flashlight (4), flower (4), hair\_dryer (4), hammer (4), jug (4), ladder (4), lamp (4), laptop (4), lipstick (2), microwave (4), motorcycle (2), mug (2), paper\_box (4), percussion (4), pillow (4), pistol (4), projector (4), rifle (2), rockery (4), sink (4), soccer (4), sofa (4), spoon (4), strings (4), sunflower (2), table (4), toilet (4), toy\_figure (4), tripod (4), walking\_stick (4), wall\_lamp (2), wind\_instruments (4) \\
Non-Rigid ($n = 141$, 45 tags) & \footnotesize accessory\_bag (2), accessory\_belt (2), accessory\_bracelet (2), accessory\_cup (3), accessory\_dress (2), accessory\_glasses (2), accessory\_glove (2), accessory\_hat (2), accessory\_necklace (3), accessory\_pendant (3), accessory\_scarf (2), accessory\_shoe (2), accessory\_trousers (2), accessory\_wallet (2), accessory\_watch (2), apple (4), bagged\_snacks (2), banana (4), broccoli (4), cabbage (2), chili (4), chinese\_cabbage (4), chinese\_hot\_dishes (4), coconut (4), corn (4), cucumber (4), eggplant (4), momordica\_charantia (4), onion (4), pear (4), pineapple (4), pizza (4), plush\_toy (2), potato (4), pumpkin (4), roast\_duck (4), scarf (2), sweet\_potato (4), tomato (4), toy\_bear (4), toy\_cat (4), toy\_dog (2), watermelon (4), winter\_melon (2), person\_person (4) \\
\midrule
\multicolumn{2}{l}{\textbf{(b) Out-of-Distribution (OOD) Test Split: No Category Overlap with the Training Set}} \\
Rigid ($n = 119$, 14 tags) & \footnotesize desk\_air\_conditioner (14), fork (10), hanger (18), lamp\_post (2), mirror (6), piano (6), picture\_frame (20), refrigerator (18), remote\_control (6), ring (2), scissors (10), spanner (2), toy (3), wheelchair (2) \\
Non-Rigid ($n = 20$, 6 tags) & \footnotesize garment\_jacket (5), garment\_shirt (3), garment\_short (3), garment\_skirt (5), cat (2), dog (2) \\
\bottomrule
\end{tabular}}
\end{table}

\begin{table}[t]
\centering
\caption{\textbf{Per-Group Results by Object Rigidity.} Full-frame rendered-image AbsRel and PSNR compare D2R with the listed baselines within the groups in Tab.~\ref{tab:rigid_tags}. D2R uses predicted composite depth; the gray row replaces it with depth from the paired composite and is excluded from rankings. Bold marks the best ranked AbsRel.}
\label{tab:rigid}
\fitwidth{\setlength{\tabcolsep}{2pt}%
\begin{tabular}{lcccccccc}
\toprule
\multirow{2}{*}{Method} & \multicolumn{4}{c}{In-Distribution} & \multicolumn{4}{c}{Out-of-Distribution} \\
\cmidrule(lr){2-5}\cmidrule(lr){6-9}
 & \multicolumn{2}{c}{Rigid} & \multicolumn{2}{c}{Non-Rigid} & \multicolumn{2}{c}{Rigid} & \multicolumn{2}{c}{Non-Rigid} \\
\cmidrule(lr){2-3}\cmidrule(lr){4-5}\cmidrule(lr){6-7}\cmidrule(lr){8-9}
 & AbsRel $\downarrow$ & PSNR $\uparrow$ & AbsRel $\downarrow$ & PSNR $\uparrow$ & AbsRel $\downarrow$ & PSNR $\uparrow$ & AbsRel $\downarrow$ & PSNR $\uparrow$ \\
\midrule
\textcolor{gray}{D2R (Ground-Truth Depth)} & \textcolor{gray}{0.041} & \textcolor{gray}{21.7} & \textcolor{gray}{0.047} & \textcolor{gray}{23.7} & \textcolor{gray}{0.059} & \textcolor{gray}{25.0} & \textcolor{gray}{0.060} & \textcolor{gray}{23.5} \\
\rowcolor{ForestGreen!12}
\textbf{D2R (Predicted Depth)} & \textbf{0.045} & 21.2 & \textbf{0.050} & 23.2 & \textbf{0.067} & 24.3 & \textbf{0.071} & 22.3 \\
Insert-Anything$^{*}$ \citep{song2026insert} \venue{AAAI'26} & 0.059 & 20.1 & 0.066 & 22.3 & 0.116 & 21.8 & 0.133 & 20.4 \\
DIRECT \citep{gong2026direct} \venue{ICML'26} & 0.082 & 18.6 & 0.094 & 20.2 & 0.134 & 19.6 & 0.154 & 18.5 \\
Nano Banana 2 \venue{Tested 2026-08} & 0.192 & 15.0 & 0.158 & 17.3 & 0.254 & 15.8 & 0.136 & 20.3 \\
\bottomrule
\end{tabular}}
\end{table}

\subsubsection{Dataset Quality Control Prompt}
\label{sec:appx_filtering_prompt}

The same prompt and image order are used for every training sample. Samples scoring at least 8 are retained; the three usability labels and score intervals are defined in the prompt.

\begingroup
\fontsize{8}{10}\selectfont
\begin{verbatim}
You are auditing training samples for an object-insertion dataset. Each sample is
meant to be a matched set: a background image, a reference image containing one
object, a mask marking that object in the reference, a mask marking where the
object belongs in the target, and a ground-truth composite in which that object
has been inserted at that location. The masks and backgrounds in this release
were produced automatically, so some samples are mislabelled.

You receive five images per sample, in this order:
  1. image_src  - the background before insertion
  2. image_ref  - the image the object is taken from
  3. mask_ref   - the mask of the object inside image_ref
  4. mask_tar   - the mask of the region the object should occupy
  5. image_tar  - the ground-truth composite

Judge only whether the sample is correctly labelled. Do not judge aesthetics,
photographic quality, or difficulty. A sample that is hard but correctly
labelled is a good sample.

Check, in order:
  1. Does image_tar contain an object inside the mask_tar region? A mask over
     empty background, or over the wrong part of the scene, is an error.
  2. Does mask_tar follow that object's silhouette? A clearly different shape,
     or a mask covering a different object, is an error. A loose or non-convex
     mask that still follows the object is not an error.
  3. Does mask_ref mark a real object in image_ref rather than background or
     empty space?
  4. Are the object in image_ref and the object inside mask_tar in image_tar the
     same instance? A different object of the same category is an error.
  5. Is image_src the same scene as image_tar with that object absent? A
     different scene, or one where the object is still present, is an error.

Return one JSON object and nothing else:
  {"score": <integer 0-10>, "usable": "yes" | "partial" | "no",
   "reason": "<one short clause>"}

score is your confidence that the sample passes all five checks, where 10 means
every check passes cleanly and 0 means the sample is unusable.
usable is "no" for score 0-4, "partial" for 5-7, and "yes" for 8-10.
reason names the first check that failed, or "clean" when none did.
\end{verbatim}
\endgroup

\subsubsection{Unpaired Evaluation and Identity Margins}
\label{sec:appx_unpaired_identity}

TF-ICON~\citep{lu2023tficon}, AIComposer~\citep{li2025aicomposer}, and ComplexCompo~\citep{lu2026shine} provide 332, 272, and 300 evaluation cases without paired composites. We therefore compare generated objects with their references to measure reference retention. Predictions use target-box crops, references use dataset-provided object masks, and both are placed on white. All methods use the same complete case set for each benchmark, with scores averaged over its cases.

CLIP-I uses IsoCLIP~\citep{magistri2026isoclip} with a SigLIP2~\citep{tschannen2025siglip2} backbone; DINO-I uses the [CLS] token of DINOv3 ViT-L/16~\citep{simeoni2025dinov3}. Paired and unpaired evaluation share the encoder settings in App.~\ref{sec:appx_identity_metrics}, but use different comparison targets and crops. Object scale and residual background can affect embedding similarity, so these scores assess reference agreement rather than scene geometry. D2R uses FLUX.1-Fill with injector-predicted Depth Anything V2 ViT-L depth; Insert-Anything$^*$~\citep{song2026insert} leaves the depth patch empty.

To interpret the identity margins, let $S_b$ be the highest similarity among the fifteen baselines for a given benchmark and encoder, and $S_{\mathrm{D2R}}$ the D2R score in Tab.~\ref{tab:unpaired}. Cosine distance to the reference is $d=1-S$; its relative reduction is $100(S_{\mathrm{D2R}}-S_b)/(1-S_b)$ percent. Matched Insert-Anything$^*$ is the strongest baseline in all six comparisons.

{We aggregate benchmarks by averaging their cosine distances with equal weight, then compute $100(\bar d_b-\bar d_{\mathrm{D2R}})/\bar d_b$. Mean CLIP distance falls from 0.215 to 0.096 (55.4\%), and mean DINO distance from 0.268 to 0.208 (22.6\%); the CLIP reduction rounds to 55\% in the abstract. For example, on TF-ICON, CLIP-I rises from 0.760 to 0.902, corresponding to a distance reduction from 0.240 to 0.098 (59.1\%). Scores and distances are displayed to three decimal places, while percentages use the four-decimal evaluation records.}

DINO cosine distance decreases by 32.6\% on TF-ICON, 9.4\% on AIComposer, and 20.7\% on ComplexCompo. AIComposer gives the smallest margin: D2R scores 0.813 versus 0.794 for Insert-Anything$^*$. These percentages measure changes in embedding agreement; they do not estimate identity accuracy or the frequency of compositing failures.

\subsubsection{Baseline Evaluation Protocol}
\label{sec:appx_baselines}

Eleven open-source methods use released checkpoints and, unless specified below, official default settings: Paint-by-Example~\citep{yang2023paint}, TF-ICON~\citep{lu2023tficon}, AnyDoor~\citep{chen2024anydoor}, MimicBrush~\citep{chen2024mimicbrush}, BIFR\"OST~\citep{li2024bifrost}, DreamFuse~\citep{huang2025dreamfuse}, UniCombine~\citep{wang2025unicombine}, Personalize-Anything~\citep{feng2026personalize}, SHINE~\citep{lu2026shine}, DIRECT~\citep{gong2026direct}, and HiFi-Inpaint~\citep{liu2026hifiinpaint}.

\noindent\textbf{BIFR\"OST.} We run mask-based inpainting from the official repository's main branch with default settings, using \texttt{bifrost.ckpt} from \texttt{lingxiao2049/Bifrost}.

\noindent\textbf{DIRECT.} We adapt DIRECT~\citep{gong2026direct} to automatic evaluation by selecting the 3D proxy rotation with the highest silhouette IoU against the target mask. The dataset-provided reference mask becomes the reference image's alpha channel, and \texttt{TRELLIS-image-large} reconstructs a 3D Gaussian proxy from this RGBA input without a background-removal model. Only rotation is optimized: the proxy retains its native TRELLIS scale, with no scaling operation or scale search, and translation is reset to zero for every candidate rather than accumulated across trials.

The initial grid evaluates $12\times5\times2=120$ unique rotations: $y\in\{0^\circ,30^\circ,\ldots,330^\circ\}$, $x\in\{-60^\circ,-30^\circ,0^\circ,30^\circ,60^\circ\}$, and $z\in\{0^\circ,180^\circ\}$. Each of the three highest-IoU candidates is refined with eight random rotation proposals at each angular scale of $15^\circ$, $6^\circ$, and $2^\circ$. A refinement is retained only when its best proposal improves IoU. The highest-IoU final candidate is chosen after $120+3\times24=192$ orientation evaluations.

For the matched comparison, we retrain Insert-Anything~\citep{song2026insert} on AnyInsertion++ as Insert-Anything$^*$; its released checkpoint is excluded because the original training corpus can overlap our evaluation splits. The retrained baseline shares D2R's data, backbone, canvas, Redux embedding, and LoRA configuration, but uses an empty depth patch and only the RGB denoising loss. Both renderers use their final checkpoints after 100K training iterations and sample with 28 steps, guidance 30, and seed~0.

\noindent\textbf{Commercial Editors.} We queried \mbox{GPT-Image-1}, GPT-Image-2\footnote{OpenAI model documentation: \href{https://developers.openai.com/api/docs/models/gpt-image-1}{GPT-Image-1} and \href{https://developers.openai.com/api/docs/models/gpt-image-2}{GPT-Image-2}.}, and Nano Banana 2\footnote{Google, \href{https://blog.google/innovation-and-ai/technology/developers-tools/build-with-nano-banana-2/}{Build with Nano Banana 2}; the public model name is Gemini 3.1 Flash Image.} through public interfaces in August 2026. Each receives the same background image, reference image, and target mask with the ADD or REPLACE instruction below. Each request produces one image, and every evaluation case has one valid output. We do not choose among valid outputs, so every editor is scored on the same cases. Unspecified options use official defaults.

GPT-Image-1 and GPT-Image-2 use \texttt{quality="high"} with available output sizes of $1024\times1024$, $1536\times1024$, and $1024\times1536$. Nano Banana 2 uses \texttt{gemini-3.1-flash-image-preview} with \texttt{image\_size="1K"}. We select the available aspect ratio closest to the input scene. If the returned dimensions differ, bicubic interpolation resizes the entire output directly to the original input-scene dimensions, without cropping or padding; outputs already at the input resolution remain unchanged. This retains the full generated frame and avoids content-dependent choices of crop or padding placement. Identity scoring follows App.~\ref{sec:appx_unpaired_identity}.

The following instructions are reproduced verbatim:
\begin{quote}
\small
\textbf{ADD.} I have a reference image of a product/object, a target image, and a mask indicating the target location. Please add the product from the reference image into the masked area of the target image in a natural and realistic way. Keep the original scene, structure, and background intact. The added product should look like it naturally belongs in the scene.

\textbf{REPLACE.} I have a reference image of a product/object, a target image, and a mask indicating the area to edit. Please replace the existing object within the masked area of the target image with the product from the reference image. Keep the surrounding scene, pose/structure, background, and lighting exactly the same. Only change the object in the masked area to match the reference.
\end{quote}

\subsection{Depth Backbones and Training Protocol}
\label{sec:appx_s1}

\subsubsection{Depth Backbones and Training Settings}
\label{sec:appx_depth_training}

\begin{table}[t]
\centering
\caption{\textbf{Stage-1 Transfer Across Depth Backbones.} Each pair compares a frozen estimator with that estimator plus a trained injector. Full-frame composite depth is scored before RGB synthesis against targets from the same estimator; compare changes within each pair.}
\label{tab:depth_full}
\fitwidth{\setlength{\tabcolsep}{3pt}%
\begin{tabular}{lllcccc}
\toprule
\multirow{2}{*}{Depth Backbone} & \multirow{2}{*}{Encoder} & \multirow{2}{*}{Method} & \multicolumn{2}{c}{In-Distribution} & \multicolumn{2}{c}{Out-of-Distribution} \\
\cmidrule(lr){4-5}\cmidrule(lr){6-7}
 & & & AbsRel $\downarrow$ & $\delta_1 \uparrow$ & AbsRel $\downarrow$ & $\delta_1 \uparrow$ \\
\midrule
 \multirow{6}{*}{Depth Anything V1 \citep{yang2024depthanything} \venue{CVPR'24}} & \multirow{2}{*}{ViT-S} & Baseline & 0.126 & 0.860 & 0.183 & 0.811 \\
 & & \cellcolor{ForestGreen!12}\textbf{+ D2R} & \cellcolor{ForestGreen!12}\textbf{0.056}\,{\footnotesize($-$56\%)} & \cellcolor{ForestGreen!12}\textbf{0.972} & \cellcolor{ForestGreen!12}\textbf{0.100}\,{\footnotesize($-$45\%)} & \cellcolor{ForestGreen!12}\textbf{0.945} \\
\cmidrule(lr){2-7}
  & \multirow{2}{*}{ViT-B} & Baseline & 0.119 & 0.876 & 0.174 & 0.823 \\
 & & \cellcolor{ForestGreen!12}\textbf{+ D2R} & \cellcolor{ForestGreen!12}\textbf{0.049}\,{\footnotesize($-$58\%)} & \cellcolor{ForestGreen!12}\textbf{0.978} & \cellcolor{ForestGreen!12}\textbf{0.088}\,{\footnotesize($-$49\%)} & \cellcolor{ForestGreen!12}\textbf{0.954} \\
\cmidrule(lr){2-7}
  & \multirow{2}{*}{ViT-L} & Baseline & 0.108 & 0.883 & 0.154 & 0.845 \\
 & & \cellcolor{ForestGreen!12}\textbf{+ D2R} & \cellcolor{ForestGreen!12}\textbf{0.046}\,{\footnotesize($-$57\%)} & \cellcolor{ForestGreen!12}\textbf{0.980} & \cellcolor{ForestGreen!12}\textbf{0.083}\,{\footnotesize($-$46\%)} & \cellcolor{ForestGreen!12}\textbf{0.954} \\
\midrule
 \multirow{6}{*}{Depth Anything V2 \citep{yang2024depthv2} \venue{NeurIPS'24}} & \multirow{2}{*}{ViT-S} & Baseline & 0.124 & 0.861 & 0.166 & 0.820 \\
 & & \cellcolor{ForestGreen!12}\textbf{+ D2R} & \cellcolor{ForestGreen!12}\textbf{0.055}\,{\footnotesize($-$56\%)} & \cellcolor{ForestGreen!12}\textbf{0.971} & \cellcolor{ForestGreen!12}\textbf{0.097}\,{\footnotesize($-$42\%)} & \cellcolor{ForestGreen!12}\textbf{0.942} \\
\cmidrule(lr){2-7}
  & \multirow{2}{*}{ViT-B} & Baseline & 0.120 & 0.867 & 0.156 & 0.833 \\
 & & \cellcolor{ForestGreen!12}\textbf{+ D2R} & \cellcolor{ForestGreen!12}\textbf{0.055}\,{\footnotesize($-$54\%)} & \cellcolor{ForestGreen!12}\textbf{0.973} & \cellcolor{ForestGreen!12}\textbf{0.080}\,{\footnotesize($-$49\%)} & \cellcolor{ForestGreen!12}\textbf{0.954} \\
\cmidrule(lr){2-7}
  & \multirow{2}{*}{ViT-L} & Baseline & 0.117 & 0.869 & 0.151 & 0.842 \\
 & & \cellcolor{ForestGreen!12}\textbf{+ D2R} & \cellcolor{ForestGreen!12}\textbf{0.046}\,{\footnotesize($-$60\%)} & \cellcolor{ForestGreen!12}\textbf{0.977} & \cellcolor{ForestGreen!12}\textbf{0.075}\,{\footnotesize($-$50\%)} & \cellcolor{ForestGreen!12}\textbf{0.965} \\
\midrule
 \multirow{6}{*}{Depth Anything V3 \citep{lin2026depth3} \venue{ICLR'26}} & \multirow{2}{*}{ViT-S} & Baseline & 0.146 & 0.824 & 0.179 & 0.782 \\
 & & \cellcolor{ForestGreen!12}\textbf{+ D2R} & \cellcolor{ForestGreen!12}\textbf{0.104}\,{\footnotesize($-$29\%)} & \cellcolor{ForestGreen!12}\textbf{0.897} & \cellcolor{ForestGreen!12}\textbf{0.142}\,{\footnotesize($-$21\%)} & \cellcolor{ForestGreen!12}\textbf{0.851} \\
\cmidrule(lr){2-7}
  & \multirow{2}{*}{ViT-B} & Baseline & 0.140 & 0.838 & 0.160 & 0.811 \\
 & & \cellcolor{ForestGreen!12}\textbf{+ D2R} & \cellcolor{ForestGreen!12}\textbf{0.093}\,{\footnotesize($-$34\%)} & \cellcolor{ForestGreen!12}\textbf{0.913} & \cellcolor{ForestGreen!12}\textbf{0.132}\,{\footnotesize($-$18\%)} & \cellcolor{ForestGreen!12}\textbf{0.863} \\
\cmidrule(lr){2-7}
  & \multirow{2}{*}{ViT-L} & Baseline & 0.114 & 0.861 & 0.153 & 0.811 \\
 & & \cellcolor{ForestGreen!12}\textbf{+ D2R} & \cellcolor{ForestGreen!12}\textbf{0.073}\,{\footnotesize($-$36\%)} & \cellcolor{ForestGreen!12}\textbf{0.934} & \cellcolor{ForestGreen!12}\textbf{0.122}\,{\footnotesize($-$20\%)} & \cellcolor{ForestGreen!12}\textbf{0.889} \\
\midrule
 \multirow{2}{*}{Depth Pro \citep{bochkovskii2025depthpro} \venue{ICLR'25}} & \multirow{2}{*}{ViT-L} & Baseline & 0.080 & 0.910 & 0.161 & 0.859 \\
 & & \cellcolor{ForestGreen!12}\textbf{+ D2R} & \cellcolor{ForestGreen!12}\textbf{0.053}\,{\footnotesize($-$34\%)} & \cellcolor{ForestGreen!12}\textbf{0.962} & \cellcolor{ForestGreen!12}\textbf{0.123}\,{\footnotesize($-$24\%)} & \cellcolor{ForestGreen!12}\textbf{0.935} \\
\midrule
 \multirow{2}{*}{Metric-Anything \citep{ma2026metricanything} \venue{ECCV'26}} & \multirow{2}{*}{ViT-H/16+} & Baseline & 0.152 & 0.815 & 0.173 & 0.810 \\
 & & \cellcolor{ForestGreen!12}\textbf{+ D2R} & \cellcolor{ForestGreen!12}\textbf{0.084}\,{\footnotesize($-$45\%)} & \cellcolor{ForestGreen!12}\textbf{0.929} & \cellcolor{ForestGreen!12}\textbf{0.094}\,{\footnotesize($-$46\%)} & \cellcolor{ForestGreen!12}\textbf{0.934} \\
\midrule
 \multirow{2}{*}{InfiniDepth \citep{yu2026infinidepth} \venue{CVPR'26}} & \multirow{2}{*}{ViT-L/16} & Baseline & 0.177 & 0.834 & 0.257 & 0.779 \\
 & & \cellcolor{ForestGreen!12}\textbf{+ D2R} & \cellcolor{ForestGreen!12}\textbf{0.113}\,{\footnotesize($-$36\%)} & \cellcolor{ForestGreen!12}\textbf{0.924} & \cellcolor{ForestGreen!12}\textbf{0.159}\,{\footnotesize($-$38\%)} & \cellcolor{ForestGreen!12}\textbf{0.881} \\
\bottomrule
\end{tabular}}
\end{table}

Table~\ref{tab:depth_full} evaluates Depth Anything V1~\citep{yang2024depthanything}, V2~\citep{yang2024depthv2}, and V3~\citep{lin2026depth3} with ViT-S, ViT-B, and ViT-L; Depth Pro~\citep{bochkovskii2025depthpro} with ViT-L; Metric-Anything~\citep{ma2026metricanything} with ViT-H/16+; and InfiniDepth~\citep{yu2026infinidepth} with ViT-L/16. Each uses its released resolution, including 518 pixels for Depth Anything. Appendix~\ref{sec:appx_stage1} describes attachment interfaces. Each backbone supplies its own supervision and scoring targets; results therefore measure injection gains within that backbone.

Stage~1 trains with Eq.~(\ref{eq:stage1}) on the curated 52.9K pool in bf16 on NVIDIA RTX PRO 6000 (96\,GB) GPUs. All Stage~1 ablations use final checkpoints after 50K training iterations, including the backbone configurations in Tab.~\ref{tab:depth_full} and the supervision comparisons in App.~\ref{sec:appx_supervision_comparison}. Most configurations use batch size 24, or 1.2M sample presentations. Depth Pro~\citep{bochkovskii2025depthpro} uses batch size 2 and Metric-Anything~\citep{ma2026metricanything} batch size 1, corresponding to 100K and 50K sample presentations. Thus equal iteration counts do not imply equal sample exposure.

The default width-640 injector on Depth Anything V2 ViT-L~\citep{yang2024depthv2} uses $518\times518$ inputs, batch size 24, and Prodigy~\citep{mishchenko2024prodigy}. Its median step time is 1.25\,s, and training takes approximately 17.5\,h. Appendix~\ref{sec:appx_cost} reports training memory.

Adding the injector lowers AbsRel and raises $\delta_1$ in every within-estimator comparison in Tab.~\ref{tab:depth_full}. The approach therefore transfers across the tested pretrained depth representations. Because targets and sample exposure differ across estimators, the absolute scores do not provide a controlled architecture ranking. Appendix~\ref{sec:appx_stage1} reports the structural ablations.

\subsubsection{{Feature Correction Versus Diffusion-Based Composite Depth Generation}}
\label{sec:appx_depth_priors}

{This experiment asks whether to predict composite depth by correcting a frozen estimator's features or by directly generating a depth image with diffusion. The latter follows the depth-generation strategy used in 3DIS~\citep{zhou20253dis}, which conditions a diffusion model on text and instance layouts. We adapt that strategy to reference-based compositing with FLUX.1-Fill and Redux, conditioning on reference and background depth. The evaluated baseline is this compositing-specific implementation; it does not use 3DIS's original LDM3D backbone or layout-conditioning adapter.}

{Table~\ref{tab:depth_prior} compares this diffusion baseline with frozen Depth Anything V2 ViT-L~\citep{yang2024depthv2} and the same estimator with our width-640 injector. Both trained predictors use the same curated pairs and derive their supervision from the same paired composites and frozen depth estimator. The injector matches encoder features of the paired composite; the diffusion baseline denoises normalized $D=F(I)$ using the Stage-2 depth objective alone, without RGB or consistency losses. This tests direct depth-image generation against correction within the frozen estimator's feature space.}

{The diffusion route reuses the Stage-2 canvas layout, $1024\times1536$ resolution, and rank-256 LoRA with scaling 256 (App.~\ref{sec:appx_s2}). Background and reference depths provide conditioning, and masked depth tokens are denoised into a composite-depth map over 28 sampling steps.}

This is a Stage-1 depth-prediction comparison. We evaluate the FLUX route before adaptation and at its final checkpoint after 50K training iterations, matching the injector's iteration count. Full-frame scoring covers all 287 ID and 139 OOD cases before RGB synthesis. Predicted maps are min-max normalized to $[0,1]$, stored as 8-bit PNGs, and scored after the same least-squares alignment as other depth predictions (App.~\ref{sec:appx_geometry_metrics}). The main paired tables score depth from RGB outputs.

Fine-tuning the FLUX route lowers ID AbsRel from 0.1262 to 0.1067 and OOD AbsRel from 0.1792 to 0.1740, reductions of 15.5\% and 2.9\%. The injector reaches 0.0462 and 0.0751, reducing error relative to the adapted FLUX route by 56.7\% and 56.8\%. Its advantage is therefore present on both familiar categories and categories absent from compositing training under this protocol.

{These results favor feature correction over our diffusion-based composite-depth implementation on both splits. Architecture, conditioning, and objective change together, so this comparison evaluates the two depth-prediction strategies as implemented for compositing. The matched experiment in App.~\ref{sec:appx_supervision_comparison} isolates the supervision target within the same injector, while Tab.~\ref{tab:depth_full} tests reuse across six depth backbones.}

\noindent{\textbf{Prediction Cost.} The injector corrects features within one feed-forward evaluation of the frozen depth estimator, whereas the tested FLUX route requires 28 denoising steps followed by VAE decoding into a depth image. The complete injector-equipped predictor takes 56.5\,ms and 5.21\,GB peak inference memory on RTX PRO 6000 (App.~\ref{sec:appx_cost}). This avoids iterative sampling for composite-depth prediction. The FLUX timings in Tab.~\ref{tab:cost} measure Stage-2 RGB synthesis and are not used as timings for this depth-generation baseline.}

This finding concerns the specified FLUX.1-Fill and Redux implementation. Marigold~\citep{ke2024repurposing} shows that diffusion priors can support depth generalization, and Lotus~\citep{he2025lotus} demonstrates the importance of the objective and sampling formulation for dense prediction. Our comparison therefore does not characterize diffusion-based depth prediction as a whole. Full-frame scores combine object and scene error without localizing the gain.

\begin{table}[ht]
\centering
\caption{{Composite Depth Prediction with Feature Correction and Diffusion.}}
\label{tab:depth_prior}
{\setlength{\tabcolsep}{4pt}
\begin{tabular}{lccc}
\toprule
Depth Predictor & {Prediction} & ID AbsRel $\downarrow$ & OOD AbsRel $\downarrow$ \\
\midrule
Frozen Depth Anything V2 ViT-L \citep{yang2024depthv2} \venue{NeurIPS'24} & {Single Pass} & 0.1168 & 0.1506 \\
{Diffusion Depth (FLUX), No Fine-Tuning} & {28 Steps} & 0.1262 & 0.1792 \\
{Diffusion Depth (FLUX), Fine-Tuned} & {28 Steps} & 0.1067 & 0.1740 \\
\rowcolor{ForestGreen!12}
{\textbf{Encoder Feature Injection (D2R; Ours)}} & {Single Pass} & {\textbf{0.0462}} & {\textbf{0.0751}} \\
\bottomrule
\end{tabular}}
\end{table}

\subsection{Rendering Backbones and Ablation Protocols}
\label{sec:appx_s2}

\subsubsection{Backbones and Prompt Conditioning}
\label{sec:appx_render_conditioning}

The renderer comparison tests whether the same composite-depth interface can support SD2-Inpainting~\citep{rombach2022high}, SDXL-Inpainting~\citep{podell2024sdxl}, Qwen-Inpainting~\citep{wu2025qwenimage}, and FLUX.1-Fill. Each uses the three-patch canvas, with depth observed throughout inference, a frozen autoencoder, and only LoRA trained. SD2 and SDXL predict noise; FLUX and Qwen predict velocity. The conditioning formats below are used consistently during training and inference; App.~\ref{sec:appx_render_training} specifies training and sampling settings.

FLUX.1-Fill receives reference appearance through both the canvas and Redux. The Redux embedding occupies the text-conditioning stream, so this path constructs no text prompt. Any remaining text slots expected by the pipeline are filled with zeros to preserve its tensor layout.

SD2-Inpainting~\citep{rombach2022high}, SDXL-Inpainting~\citep{podell2024sdxl}, and Qwen-Inpainting~\citep{wu2025qwenimage} receive reference appearance through the canvas and use text instructions without a separate image-prompt encoder. Each instruction has exactly two variable slots: the object noun \{obj\}, read from the sample's file name, and the insertion or replacement phrase, denoted \{verb\} in the compressed instruction or \{action\} in the full instruction. A single space serves as the negative prompt.

The CLIP text encoders of SD2-Inpainting and SDXL-Inpainting truncate at 77 tokens. Both therefore receive the compressed instruction, which contains 42 to 56 tokens under that tokenizer across the released splits: \textit{``photorealistic composite: left, a scene with the \{obj\} \{verb\}, cast shadow, reflections, natural contact and lighting; top right, reference photo of the \{obj\}; bottom right, its depth map.''} The \{verb\} slot is \textit{inserted} for insertion and \textit{replacing the original object} for replacement.

The Qwen-Inpainting text encoder accepts 512 tokens for sampling and 1024 for training. Its full instruction contains 115 to 152 tokens under the Qwen tokenizer: \textit{``A composite image. Left half: a realistic scene where \{action\}. Top right: a clean reference photograph of a single \{obj\} on a plain background. Bottom right: a grayscale depth map matching the left half --- bright pixels for nearby surfaces, dark pixels for distant surfaces --- accurately reflecting the inserted \{obj\}'s 3D placement and contact geometry. Maintain photorealistic quality.''}

For Qwen-Inpainting, the \{action\} slot is \textit{``the \{obj\} from the top-right reference panel has been naturally inserted into the highlighted region --- with cast shadows on the ground and adjacent surfaces, surface reflections and specular highlights, and physically plausible contact with the scene, all consistent with the existing lighting direction, materials, perspective, and color temperature''} for an insertion. For a replacement, it is \textit{``the original object has been replaced with the \{obj\} from the top-right reference panel --- preserving the original cast shadows, surface reflections, and physical contact patterns of the scene while adopting the visual identity of the \{obj\}''}.

The shorter instruction omits the full instruction's explicit depth-polarity description. Both retain guidance on shadows, reflections, contact, lighting, and the canvas layout. The instructions above are reproduced verbatim.

\subsubsection{Training and Sampling Settings}
\label{sec:appx_render_training}

All trained Stage~2 ablation variants are evaluated at their final checkpoints after 100K training iterations. SD2-Inpainting, Qwen-Inpainting, and FLUX.1-Fill train rank-256 LoRA with scaling 256 using Prodigy~\citep{mishchenko2024prodigy} at a configured learning rate of 1.0. SDXL uses the AdamW settings given below. D2R renderers use exported Stage-1 predictions, the auxiliary objectives, and per-sample depth selection defined in App.~\ref{sec:appx_stage2}.

SD2-Inpainting~\citep{rombach2022high} uses a 512-pixel composite patch, a $512\times768$ canvas, batch size 12, and no gradient accumulation. White reference padding replaces the pipeline's edge-padding default. Sampling uses 28 steps, classifier-free guidance 7.0, and seed~0.

Qwen-Inpainting~\citep{wu2025qwenimage} uses a 768-pixel composite patch and a $768\times1152$ canvas to fit available training memory. Sampling uses 28 steps, true classifier-free guidance 4.0, and seed~42.

FLUX.1-Fill uses a 1024-pixel composite patch and a $1024\times1536$ canvas, with effective batch size 8 and four gradient-accumulation steps. Training takes approximately 35\,h. FLUX-based D2R and matched Insert-Anything$^*$ use Redux, 28 sampling steps, guidance 30, and seed~0. This follows the 28-step Insert-Anything~\citep{song2026insert} protocol; the FLUX.1-Fill example uses 50 steps.

SDXL-Inpainting~\citep{podell2024sdxl} uses a 1024-pixel composite patch, a $1024\times1536$ canvas, batch size 2, four gradient-accumulation steps, fp32 adapter weights, and AdamW~\citep{loshchilov2019decoupled} at $5\times10^{-5}$. Both pretrained and adapted renderers sample with guidance 7.0 and 28 steps. Reference and depth patches remain fixed under Alg.~\ref{alg:inference}. Appendix~\ref{sec:appx_cost} gives precision, latency, and training memory for the evaluated configurations.

To assess training stability, we train each stage five times under its default configuration with seeds 0, 1, 2, 3, and 4 and evaluate the final checkpoints. Relative changes across the applicable geometric, photometric, and identity metrics average approximately 1\% overall. Sampling-seed checks separately indicated stable performance.

\subsubsection{Comparison Across Diffusion Backbones}
\label{sec:appx_renderer_results}

Table~\ref{tab:generator} compares pretrained and D2R-adapted renderers on the same 139 OOD cases. Within each backbone, both receive identical Stage-1 predicted composite depth on the same canvas and use the sampling settings in App.~\ref{sec:appx_render_training}. Thus the pretrained comparator already receives depth, unlike the empty-depth Insert-Anything$^*$ baseline (Tab.~\ref{tab:main_ood}). All four adapted renderers improve AbsRel and PSNR, supporting adaptation across the tested generators. Resolution, conditioning, and training differences prevent a controlled ranking across architectures.

\subsubsection{Loss Ablations and Sampling Comparisons}
\label{sec:appx_render_ablations}

{Table~\ref{tab:loss} evaluates the renderer through successive comparisons. Rows (a) and (b) test adding composite depth under RGB-only supervision. With depth supplied, rows (c) and (d) add depth denoising or RGB-depth consistency separately; comparing them with (f) tests their combination. Rows (e) and (f) retain both objectives and change only the consistency weighting from uniform to $1-t$. All rows retain RGB supervision and matched data, backbone, LoRA, and training settings, using final checkpoints after 100K iterations. Conditioned rows (b) to (f) share per-sample 50:50 selection of predicted or target depth during training, full-canvas corruption, and Stage-1 predicted depth at evaluation. Depth-patch denoising requires a depth prediction region and its target, whereas consistency supervises depth estimated from the generated RGB patch.}

Table~\ref{tab:mix} instead changes only the training-time depth source. GT-only inputs, predicted-only inputs, and blending replace the default per-sample selection while retaining GT supervision and the other training settings. Each renderer is then evaluated with both target and predicted depth. Equation~(\ref{eq:supply}) defines selection, and App.~\ref{sec:appx_render_objectives} specifies all four inputs.

Latent Optimization (row (c) of Tab.~\ref{tab:ladder}) tests correction toward Stage-1 depth using released FLUX.1-Fill with Redux and no weight training. It applies the same correction at all 28 steps.

For this baseline, $\sigma_t$ denotes the sampler's noise level and $v_t=\mathrm{sg}(\epsilon_\theta(x_t,t,c))$ its detached velocity prediction. The clean estimate is $\hat x_0=x_t-\sigma_t v_t$. We decode its composite patch, estimate depth with frozen Depth Anything V2 ViT-L~\citep{yang2024depthv2}, and compare it with Stage-1 composite depth $\hat D$ using the normalization $A$ defined in App.~\ref{sec:appx_render_objectives}:
\begin{equation}
\begin{aligned}
\mathcal L_{\mathrm{latent}} &= \left\|A\!\left(F\!\left(\mathcal D(\hat x_0)\big|_{\Omega_{\mathrm{comp}}}\right)\right)-\mathrm{sg}\!\left(A(\hat D)\right)\right\|_1, \\
g_t &= \nabla_{x_t}\mathcal L_{\mathrm{latent}}, \qquad
x_t \leftarrow x_t-\gamma\sigma_t\|x_t\|_2\frac{g_t}{\|g_t\|_2}, \qquad \gamma=0.15.
\end{aligned}
\label{eq:latent_correction}
\end{equation}
For a nonzero gradient, the update normalizes its direction and scales the correction by the current latent norm and noise level. Thus $\gamma$ (\texttt{g\_scale}) is a relative correction factor, not a conventional learning rate; the correction shrinks as $\sigma_t$ approaches zero. Detaching the velocity prediction restricts backpropagation to the RGB decoder and depth estimator, avoiding the denoiser's backward graph. The latent correction takes approximately 2\,s per step on RTX PRO 6000.

\begin{figure}[t]
\centering
\includegraphics[width=\textwidth]{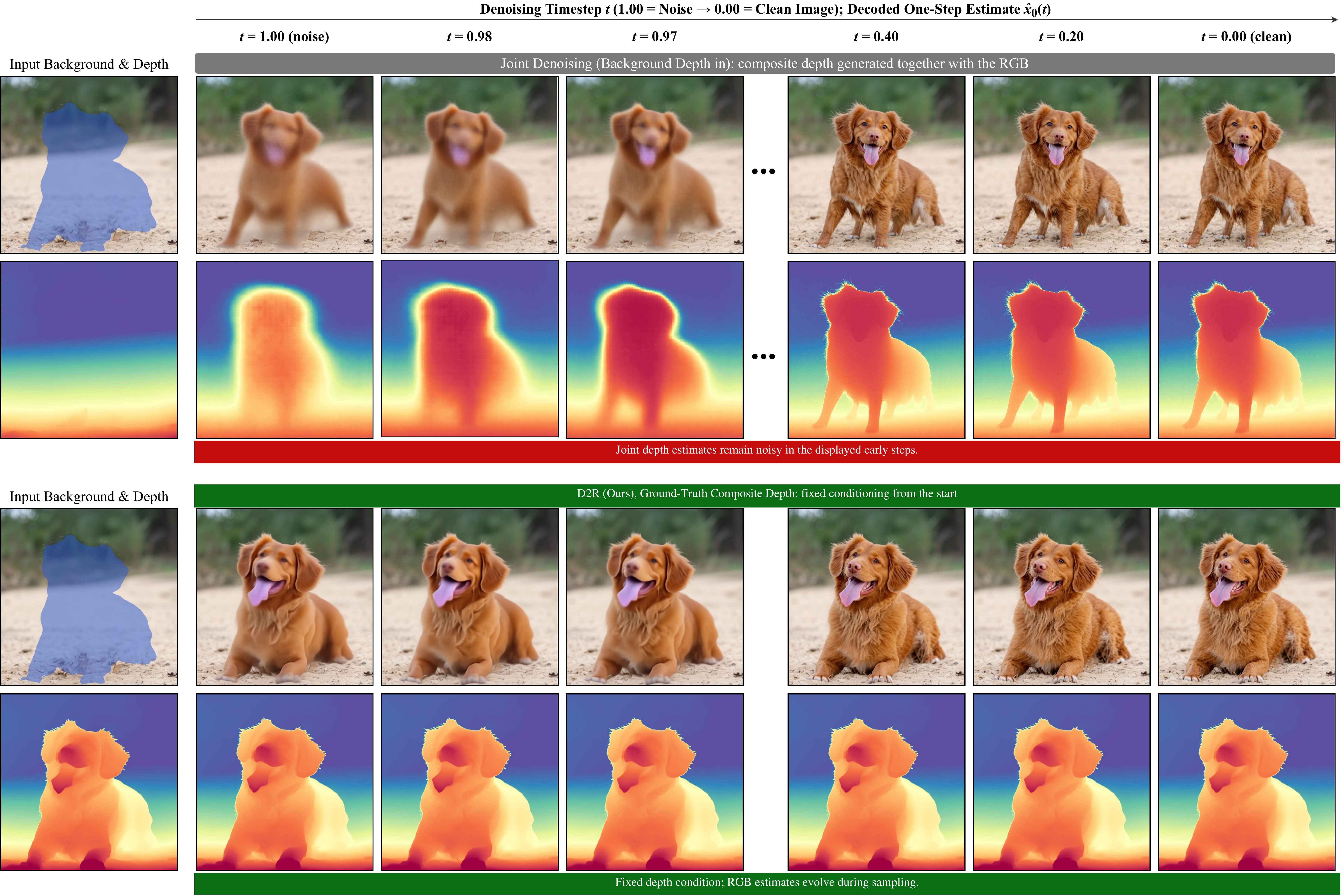}
\vspace{-13pt}
\caption{\textbf{Denoising Trajectories with Joint Prediction and Fixed Ground-Truth Depth.} The upper branch predicts depth jointly with RGB; the lower D2R branch uses GT composite depth as a fixed condition. Tiles show one-step estimates from one example. Table~\ref{tab:ladder} instead evaluates D2R using composite depth predicted by Stage 1.}
\label{fig:timeline}
\end{figure}

Joint Denoising (row (d) of Tab.~\ref{tab:ladder}) trains the same backbone and canvas for 100K iterations but generates depth alongside RGB. Its conditioning depth is $F(I_{\mathrm{bg}})$, while the targets are $D$ for depth and $I$ for RGB. The final checkpoint supplies the quantitative results in Tab.~\ref{tab:ladder}. Figure~\ref{fig:timeline} visualizes this joint trajectory alongside D2R with fixed GT composite depth; the table evaluates D2R with Stage-1 predicted depth.

The geometry-provisioning strategies in Tab.~\ref{tab:ladder} differ in both depth construction and training. Their results compare complete implementations rather than isolating conditioning time; the trajectory in Fig.~\ref{fig:timeline} illustrates the conditioning schemes qualitatively.

\subsection{Evaluation Metrics}
\label{sec:appx_metrics}

\subsubsection{Depth Alignment and Geometric Metrics}
\label{sec:appx_geometry_metrics}

We evaluate the geometry of the generated RGB image through $D_{\mathrm{out}}=F(\hat I)$, using $D=F(I)$ as the paired target and frozen Depth Anything V2 ViT-L~\citep{yang2024depthv2} as $F$. Direct Stage-1 evaluation replaces $D_{\mathrm{out}}$ with $\hat D$. Because relative-depth predictions can differ in scale and offset, we align the prediction to the target by ordinary least squares. The fitting support $\mathcal B$ consists of background pixels outside the target mask dilated with a $21\times21$ square kernel at the original evaluation resolution. If fewer than 100 background pixels remain, the fit uses the full frame:
\begin{equation}
D_{\mathrm{aligned}}=sD_{\mathrm{out}}+o, \qquad (s,o)=\arg\min_{a,b}\sum_{p\in\mathcal B}\bigl(aD_{\mathrm{out},p}+b-D_p\bigr)^2.
\label{eq:align}
\end{equation}
We solve the normal equations in closed form; degenerate fits use unit scale and a fitted offset. The aligned map is then scored over the full-frame pixel set $\mathcal R$:
\begin{equation}
\begin{aligned}
\mathrm{AbsRel} &= \frac{1}{|\mathcal{R}|} \sum_{p \in \mathcal{R}} \frac{\bigl| D_{\mathrm{aligned},p} - D_p \bigr|}{\max\bigl(D_p,\, \varepsilon_{\mathrm d}\bigr)}, \\
\mathrm{RMSE} &= \sqrt{\frac{1}{|\mathcal{R}|} \sum_{p \in \mathcal{R}} \bigl( D_{\mathrm{aligned},p} - D_p \bigr)^2},
\end{aligned}
\label{eq:absrel}
\end{equation}
\begin{equation}
\delta_k = \frac{1}{|\mathcal{R}|} \sum_{p \in \mathcal{R}} \mathbf{1}\!\left[ \max\!\left( \frac{D^+_{\mathrm{aligned},p}}{D^+_p},\; \frac{D^+_p}{D^+_{\mathrm{aligned},p}} \right) < \tau^{k} \right],
\label{eq:delta}
\end{equation}
Here $\varepsilon_{\mathrm d}=0.05$, $D^+_p=\max(D_p,\varepsilon_{\mathrm d})$, and $D^+_{\mathrm{aligned},p}=\max(D_{\mathrm{aligned},p},\varepsilon_{\mathrm d})$. AbsRel floors only its denominator; its numerator and RMSE use the original aligned and target values. For ratio accuracy, both maps are floored before taking ratios, and every pixel remains in the average. We use $\tau=1.25$~\citep{yang2024depthanything,yang2024depthv2} and report $\delta_1$. These metrics compare aligned estimator outputs rather than calibrated physical depth.

\subsubsection{Paired Uncertainty and Regional Evaluation}
\label{sec:appx_paired_uncertainty}

We test whether D2R's average geometric improvement persists across cases, rather than depending on a few large gains. D2R uses Stage-1 predicted depth in all comparisons; the rendered-image cases are the same as in the main paired evaluation. For each case, $\Delta=\mathrm{AbsRel}_{\mathrm{D2R}}-\mathrm{AbsRel}_{\mathrm{baseline}}$, so negative differences favor D2R. We compute percentile intervals from 10,000 paired bootstrap resamples with seed~0, sampling individual cases while retaining the paired method scores.

Insertion and replacement cases can share an image ID, but the primary bootstrap samples them separately. For the OOD comparison with Insert-Anything$^*$~\citep{song2026insert}, a cluster bootstrap over 82 image IDs gives $[-0.0785,-0.0286]$, while the bias-corrected and accelerated (BCa) interval is $[-0.0749,-0.0354]$. Both exclude zero, as does the case-level percentile interval in Tab.~\ref{tab:paired_robustness}. The improvement persists under both alternative bootstrap procedures.

Table~\ref{tab:paired_robustness} compares D2R with matched Insert-Anything$^*$ and MimicBrush~\citep{chen2024mimicbrush}. Wins exclude differences within the AbsRel tie tolerance $|\Delta|\leq0.001$. To measure sensitivity to the largest gains, we remove the $\operatorname{round}(0.1n)$ cases most favorable to D2R: 14 OOD or 29 ID cases. The remaining mean difference is negative in all four comparisons. Large improvements contribute to the aggregate gain, but they do not account for the entire full-frame advantage.

\begin{table}[ht]
\centering
\caption{Paired Full-Frame AbsRel Differences and Sensitivity to Large Gains.}
\label{tab:paired_robustness}
\fitwidth{\setlength{\tabcolsep}{4pt}%
\begin{tabular}{llcccc}
\toprule
Comparator & Split & Mean $\Delta$ & 95\% CI & Wins / $n$ & Trimmed $\Delta$ \\
\midrule
Insert-Anything$^*$ \citep{song2026insert} \venue{AAAI'26} & OOD & $-0.0514$ & $[-0.0717,-0.0337]$ & 103 / 139 & $-0.0218$ \\
Insert-Anything$^*$ \citep{song2026insert} \venue{AAAI'26} & ID & $-0.0147$ & $[-0.0198,-0.0101]$ & 197 / 287 & $-0.0054$ \\
MimicBrush \citep{chen2024mimicbrush} \venue{NeurIPS'24} & OOD & $-0.0231$ & $[-0.0317,-0.0148]$ & 95 / 139 & $-0.0103$ \\
MimicBrush \citep{chen2024mimicbrush} \venue{NeurIPS'24} & ID & $-0.0187$ & $[-0.0233,-0.0141]$ & 201 / 287 & $-0.0097$ \\
\bottomrule
\end{tabular}}
\end{table}

To determine where the improvement occurs, we also score the object interior, its boundary, and the background. Predicted depth is resized to the original target-depth resolution before a $21\times21$ square kernel is applied to the target mask $M$. The interior is $\operatorname{erode}(M)$; the boundary band is $\operatorname{dilate}(M)\setminus\operatorname{erode}(M)$, extending approximately ten pixels to each side of the boundary; and the background is the complement of the dilation. All regions reuse the alignment in Eq.~(\ref{eq:align}), with no additional regional fit. Empty regions are omitted from that region's mean, and the valid case count is retained. Background scores measure residual disagreement on the alignment support.

On the 139 OOD cases, D2R lowers interior and boundary-band AbsRel relative to both comparators (Tab.~\ref{tab:paired_regions}). D2R has interior AbsRel of 0.0509 and boundary-band AbsRel of 0.1491. All four paired intervals exclude zero, showing that the full-frame advantage also extends to the inserted object and nearby scene content. These differences use unrounded case scores. The result is metric-specific: the OOD interior $\delta_1$ gain over Insert-Anything$^*$ reverses after excluding the largest 10\% of improvements. The AbsRel evidence therefore supports regional gains without implying the same stability for every metric.

\begin{table}[ht]
\centering
\caption{Object and Boundary AbsRel on OOD Cases.}
\label{tab:paired_regions}
\fitwidth{\setlength{\tabcolsep}{5pt}%
\begin{tabular}{llccc}
\toprule
Comparator & Region & Baseline AbsRel & Mean $\Delta$ & 95\% CI \\
\midrule
Insert-Anything$^*$ \citep{song2026insert} \venue{AAAI'26} & Interior & 0.0734 & $-0.0225$ & $[-0.0318,-0.0138]$ \\
 & Band & 0.1887 & $-0.0395$ & $[-0.0560,-0.0230]$ \\
MimicBrush \citep{chen2024mimicbrush} \venue{NeurIPS'24} & Interior & 0.1119 & $-0.0610$ & $[-0.0747,-0.0485]$ \\
 & Band & 0.1888 & $-0.0396$ & $[-0.0562,-0.0225]$ \\
\bottomrule
\end{tabular}}
\end{table}

\subsubsection{Cross-Scorer Evaluation}
\label{sec:appx_cross_scorer}

We hold the generated images fixed and change the depth estimator used to score them. Depth Anything V1~\citep{yang2024depthanything}, V2~\citep{yang2024depthv2}, V3~\citep{lin2026depth3}, Depth Pro~\citep{bochkovskii2025depthpro}, and Metric-Anything~\citep{ma2026metricanything} each evaluate all sixteen methods on the same 287 ID and 139 OOD cases. Each scorer estimates depth from both the generated and ground-truth images, then applies Eq.~(\ref{eq:align}) and full-frame scoring. D2R retains its adopted Stage-1 predictions throughout. Depth Pro and Metric-Anything provide no training targets for this D2R configuration; their trained injector variants belong to the separate Stage-1 ablations.

This comparison uses floating-point depth maps, whereas the main protocol stores 8-bit PNGs, so its values are reported separately. Tables~\ref{tab:scorer} and~\ref{tab:scorer_ood} give AbsRel, ranks, and Spearman correlations relative to the Depth Anything V2 ordering; DAv1 to DAv3 abbreviate the corresponding Depth Anything versions. Absolute errors should be compared within each scorer, because their depth representations and scales differ.

D2R ranks first under every scorer on both splits, although the remaining methods change order. Across the four alternative scorers, Spearman correlations with the Depth Anything V2 ordering range from 0.888 to 0.985 on ID and from 0.824 to 0.976 on OOD. The leading result is therefore stable even where the broader ranking varies, supporting the geometric improvement beyond the estimator that supplied training targets.

\begin{table}[!t]
\centering
\caption{Cross-Scorer AbsRel and Ranks on the In-Distribution Split.}
\label{tab:scorer}
\fitwidth{\renewcommand{\arraystretch}{0.9}\setlength{\tabcolsep}{1.75pt}%
\begin{tabular}{lcccccccccc}
\toprule
\multirow{2}{*}{Method} & \multicolumn{2}{c}{DAv2} & \multicolumn{2}{c}{DAv1} & \multicolumn{2}{c}{DAv3} & \multicolumn{2}{c}{Depth Pro} & \multicolumn{2}{c}{Metric-Anything} \\
\cmidrule(lr){2-3} \cmidrule(lr){4-5} \cmidrule(lr){6-7} \cmidrule(lr){8-9} \cmidrule(lr){10-11}
 & AbsRel $\downarrow$ & Rank & AbsRel $\downarrow$ & Rank & AbsRel $\downarrow$ & Rank & AbsRel $\downarrow$ & Rank & AbsRel $\downarrow$ & Rank \\
\midrule
\rowcolor{ForestGreen!12}
\textbf{D2R (Ours)} & \textbf{0.048} & \textbf{1} & \textbf{0.049} & \textbf{1} & \textbf{0.057} & \textbf{1} & \textbf{0.040} & \textbf{1} & \textbf{0.090} & \textbf{1} \\
Insert-Anything$^{*}$ \citep{song2026insert} \venue{AAAI'26} & 0.064 & 2 & 0.066 & 3 & 0.074 & 3 & 0.051 & 2 & 0.099 & 2 \\
MimicBrush \citep{chen2024mimicbrush} \venue{NeurIPS'24} & 0.067 & 3 & 0.064 & 2 & 0.081 & 4 & 0.053 & 3 & 0.134 & 4 \\
UniCombine \citep{wang2025unicombine} \venue{ICCV'25} & 0.073 & 4 & 0.073 & 4 & 0.070 & 2 & 0.058 & 4 & 0.143 & 6 \\
DIRECT \citep{gong2026direct} \venue{ICML'26} & 0.089 & 5 & 0.083 & 5 & 0.089 & 5 & 0.059 & 5 & 0.124 & 3 \\
BIFR\"OST \citep{li2024bifrost} \venue{NeurIPS'24} & 0.099 & 6 & 0.096 & 6 & 0.098 & 8 & 0.064 & 7 & 0.150 & 7 \\
HiFi-Inpaint \citep{liu2026hifiinpaint} \venue{CVPR'26} & 0.105 & 7 & 0.098 & 7 & 0.091 & 6 & 0.062 & 6 & 0.140 & 5 \\
Paint-by-Example \citep{yang2023paint} \venue{CVPR'23} & 0.113 & 8 & 0.102 & 8 & 0.105 & 9 & 0.070 & 9 & 0.198 & 12 \\
DreamFuse \citep{huang2025dreamfuse} \venue{ICCV'25} & 0.129 & 9 & 0.124 & 9 & 0.098 & 7 & 0.070 & 8 & 0.157 & 8 \\
SHINE \citep{lu2026shine} \venue{ICLR'26} & 0.164 & 10 & 0.152 & 10 & 0.125 & 11 & 0.088 & 10 & 0.171 & 10 \\
AnyDoor \citep{chen2024anydoor} \venue{CVPR'24} & 0.176 & 11 & 0.162 & 12 & 0.143 & 14 & 0.097 & 12 & 0.198 & 13 \\
Personalize-Anything \citep{feng2026personalize} \venue{AAAI'26} & 0.177 & 12 & 0.162 & 13 & 0.130 & 12 & 0.111 & 14 & 0.203 & 15 \\
Nano Banana 2 \venue{Tested 2026-08} & 0.179 & 13 & 0.152 & 11 & 0.117 & 10 & 0.095 & 11 & 0.163 & 9 \\
TF-ICON \citep{lu2023tficon} \venue{ICCV'23} & 0.207 & 14 & 0.198 & 15 & 0.155 & 15 & 0.110 & 13 & 0.202 & 14 \\
GPT-Image-2 \venue{Tested 2026-08} & 0.209 & 15 & 0.197 & 14 & 0.136 & 13 & 0.112 & 15 & 0.188 & 11 \\
GPT-Image-1 \venue{Tested 2026-08} & 0.470 & 16 & 0.509 & 16 & 0.291 & 16 & 0.215 & 16 & 0.358 & 16 \\
\midrule
Spearman $\rho$ vs.\ DAv2 & \multicolumn{2}{c}{1.000} & \multicolumn{2}{c}{0.985} & \multicolumn{2}{c}{0.941} & \multicolumn{2}{c}{0.979} & \multicolumn{2}{c}{0.888} \\
\bottomrule
\end{tabular}}
\end{table}

\begin{table}[!t]
\centering
\caption{Cross-Scorer AbsRel and Ranks on the Out-of-Distribution Split.}
\label{tab:scorer_ood}
\fitwidth{\renewcommand{\arraystretch}{0.9}\setlength{\tabcolsep}{1.75pt}%
\begin{tabular}{lcccccccccc}
\toprule
\multirow{2}{*}{Method} & \multicolumn{2}{c}{DAv2} & \multicolumn{2}{c}{DAv1} & \multicolumn{2}{c}{DAv3} & \multicolumn{2}{c}{Depth Pro} & \multicolumn{2}{c}{Metric-Anything} \\
\cmidrule(lr){2-3} \cmidrule(lr){4-5} \cmidrule(lr){6-7} \cmidrule(lr){8-9} \cmidrule(lr){10-11}
 & AbsRel $\downarrow$ & Rank & AbsRel $\downarrow$ & Rank & AbsRel $\downarrow$ & Rank & AbsRel $\downarrow$ & Rank & AbsRel $\downarrow$ & Rank \\
\midrule
\rowcolor{ForestGreen!12}
\textbf{D2R (Ours)} & \textbf{0.068} & \textbf{1} & \textbf{0.072} & \textbf{1} & \textbf{0.105} & \textbf{1} & \textbf{0.103} & \textbf{1} & \textbf{0.077} & \textbf{1} \\
MimicBrush \citep{chen2024mimicbrush} \venue{NeurIPS'24} & 0.091 & 2 & 0.132 & 2 & 0.130 & 2 & 0.137 & 3 & 0.160 & 4 \\
Insert-Anything$^{*}$ \citep{song2026insert} \venue{AAAI'26} & 0.120 & 3 & 0.136 & 3 & 0.131 & 3 & 0.139 & 5 & 0.151 & 2 \\
BIFR\"OST \citep{li2024bifrost} \venue{NeurIPS'24} & 0.121 & 4 & 0.138 & 4 & 0.143 & 6 & 0.138 & 4 & 0.166 & 5 \\
HiFi-Inpaint \citep{liu2026hifiinpaint} \venue{CVPR'26} & 0.133 & 5 & 0.157 & 7 & 0.133 & 4 & 0.145 & 9 & 0.156 & 3 \\
Paint-by-Example \citep{yang2023paint} \venue{CVPR'23} & 0.137 & 6 & 0.146 & 6 & 0.156 & 9 & 0.169 & 10 & 0.216 & 10 \\
DIRECT \citep{gong2026direct} \venue{ICML'26} & 0.138 & 7 & 0.146 & 5 & 0.135 & 5 & 0.126 & 2 & 0.169 & 6 \\
Personalize-Anything \citep{feng2026personalize} \venue{AAAI'26} & 0.189 & 8 & 0.201 & 9 & 0.154 & 8 & 0.171 & 11 & 0.213 & 9 \\
UniCombine \citep{wang2025unicombine} \venue{ICCV'25} & 0.192 & 9 & 0.198 & 8 & 0.150 & 7 & 0.139 & 7 & 0.229 & 11 \\
SHINE \citep{lu2026shine} \venue{ICLR'26} & 0.220 & 10 & 0.238 & 11 & 0.169 & 12 & 0.172 & 12 & 0.209 & 8 \\
DreamFuse \citep{huang2025dreamfuse} \venue{ICCV'25} & 0.228 & 11 & 0.241 & 12 & 0.162 & 11 & 0.139 & 6 & 0.233 & 12 \\
Nano Banana 2 \venue{Tested 2026-08} & 0.238 & 12 & 0.211 & 10 & 0.160 & 10 & 0.144 & 8 & 0.207 & 7 \\
AnyDoor \citep{chen2024anydoor} \venue{CVPR'24} & 0.260 & 13 & 0.261 & 13 & 0.187 & 13 & 0.189 & 13 & 0.274 & 14 \\
GPT-Image-2 \venue{Tested 2026-08} & 0.302 & 14 & 0.303 & 14 & 0.196 & 14 & 0.199 & 14 & 0.270 & 13 \\
TF-ICON \citep{lu2023tficon} \venue{ICCV'23} & 0.360 & 15 & 0.400 & 15 & 0.240 & 15 & 0.239 & 15 & 0.342 & 15 \\
GPT-Image-1 \venue{Tested 2026-08} & 0.492 & 16 & 0.549 & 16 & 0.301 & 16 & 0.296 & 16 & 0.410 & 16 \\
\midrule
Spearman $\rho$ vs.\ DAv2 & \multicolumn{2}{c}{1.000} & \multicolumn{2}{c}{0.976} & \multicolumn{2}{c}{0.956} & \multicolumn{2}{c}{0.824} & \multicolumn{2}{c}{0.906} \\
\bottomrule
\end{tabular}}
\end{table}

\subsubsection{Photometric and Identity Metrics}
\label{sec:appx_identity_metrics}

For paired photometric evaluation, PSNR and LPIPS~\citep{zhang2018unreasonable} compare the generated composite $\hat I$ with its target $I$ after resizing both to the 512-pixel evaluation resolution.

Paired and unpaired identity evaluation share the same image encoders: IsoCLIP~\citep{magistri2026isoclip} for CLIP-I and DINOv3 ViT-L/16~\citep{simeoni2025dinov3} for DINO-I. IsoCLIP uses the SigLIP2~\citep{tschannen2025siglip2} backbone \texttt{ViT-gopt-16-}\allowbreak\texttt{SigLIP2-384} with \texttt{webli} weights. The call \texttt{IsoCLIP(model, weights, 200, 50)} removes the top 200 and bottom 50 SVD directions from the image and text projections.

DINO-I uses the pretrained \texttt{facebook/}\allowbreak\texttt{dinov3-vitl16-}\allowbreak\texttt{pretrain-lvd1689m} checkpoint with $512\times512$ inputs and its [CLS] descriptor. For both metrics, we L2-normalize the image descriptors, compute the cosine similarity for each image pair, and average over cases.

The comparison target depends on whether a paired composite is available. Paired identity uses GT and predicted composite crops from the same region defined by the GT object mask, without replacing the surrounding pixels with white. Every method uses the same 287 ID and 139 OOD cases after the common exclusions. Unpaired identity compares the reference object with the predicted composite using the crops in App.~\ref{sec:appx_unpaired_identity}. Because both the target and crop convention change, the scores are not directly comparable across protocols. Identity and paired geometry metrics serve complementary roles: embedding agreement measures similarity to the object, while depth agreement evaluates the result in the target scene.

\subsection{Limitations}
\label{sec:appx_limits}

\noindent\textbf{Errors in the Geometric Condition.} Fixed depth can pass an incorrect footprint, contact height, or occlusion order into RGB. Viewing predicted depth beside the generated image helps distinguish the supplied geometry from the renderer's response. Target-depth substitution (Tabs.~\ref{tab:main_id} and~\ref{tab:main_ood}) measures the benefit of a better condition without assigning every remaining error to one stage. Multiple geometries may fit the inputs, whereas paired evaluation uses one target. Relative depth leaves metric scale undetermined, and thin or transparent structures remain difficult.

\noindent\textbf{Color and Texture Harmonization.} A geometrically plausible insertion can still mismatch the scene's appearance. The cat in Fig.~\ref{fig:qual_aicomposer} (row four) fits the target region but retains photographic fur shading in an ink drawing; the church in Fig.~\ref{fig:qual_tficon} (row seven) retains color in a grayscale scene. D2R can reflect illustrated styles, yet larger palette and texture changes remain difficult. Depth specifies structure rather than surface appearance, so it cannot alone resolve RGB-to-grayscale harmonization. Photographic training may also favor reference retention, although its contribution is not isolated here. These limitations motivate geometry-guided harmonization and relighting as future work.

\noindent\textbf{Scope of the Evidence.} Agreement across depth scorers supports the ranking without establishing physical accuracy or local contact correctness. Regional evaluation (App.~\ref{sec:appx_paired_uncertainty}) addresses errors hidden by full-frame averages. The matched RGB baseline controls data and renderer but cannot separate the pretrained depth prior from the added 272.4M injector parameters. Holding both fixed, the feature-versus-depth comparison isolates supervision (App.~\ref{sec:appx_supervision_comparison}).

\clearpage
\section{Additional Qualitative Results}
\label{sec:appx_qual}

\begin{figure}[!ht]
\centering
\includegraphics[width=\textwidth,height=0.82\textheight,keepaspectratio]{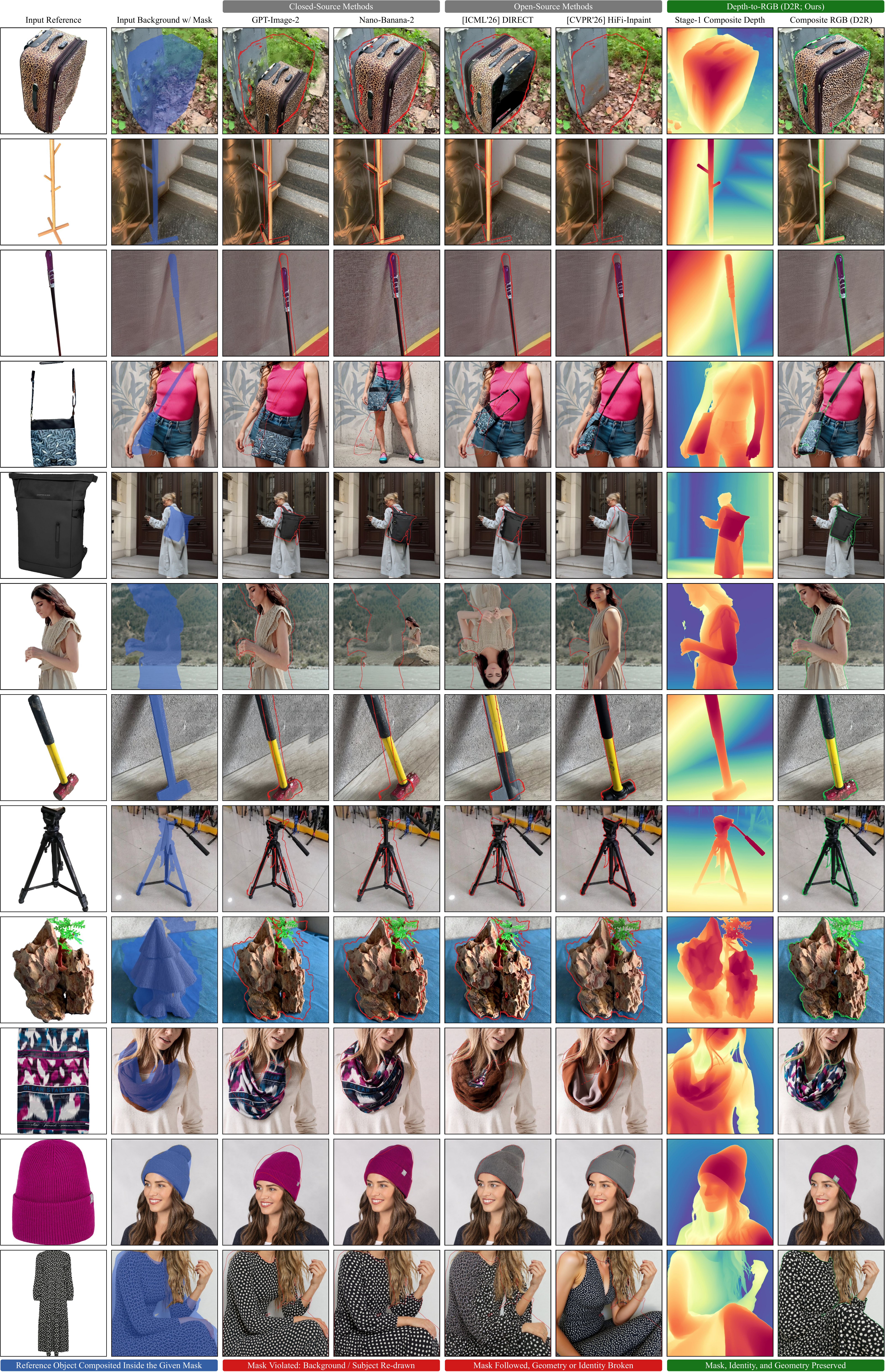}
\caption{Qualitative Results on the AnyInsertion++ In-Distribution (ID) Split.}

\label{fig:qual_id_stage1}
\end{figure}

\begin{figure}[t]
\centering
\includegraphics[width=\textwidth]{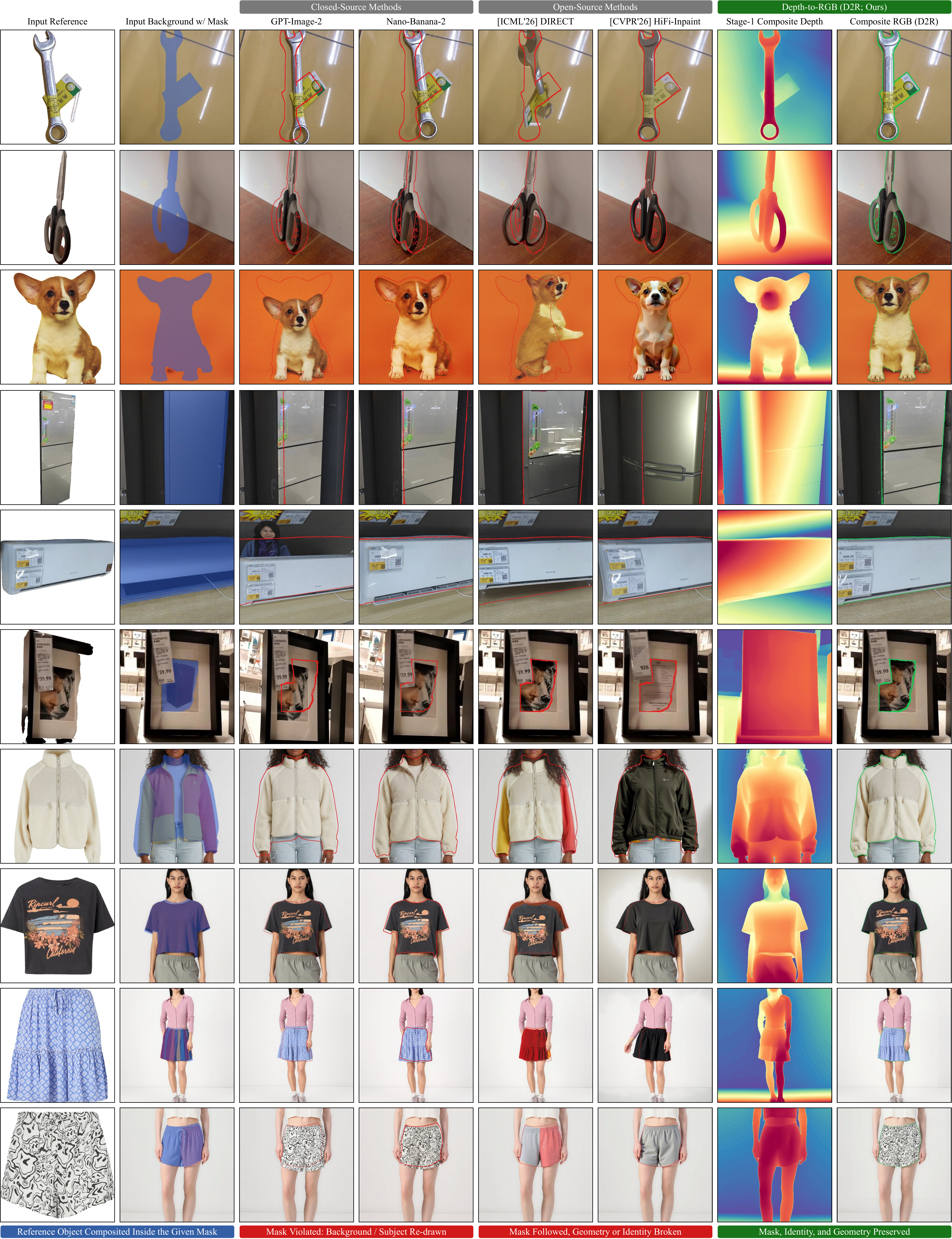}
\caption{Qualitative Results on the AnyInsertion++ Out-of-Distribution (OOD) Split.}

\label{fig:qual_ood_stage1}
\end{figure}

\begin{figure}[!ht]
\centering
\includegraphics[width=\textwidth,height=0.80\textheight,keepaspectratio]{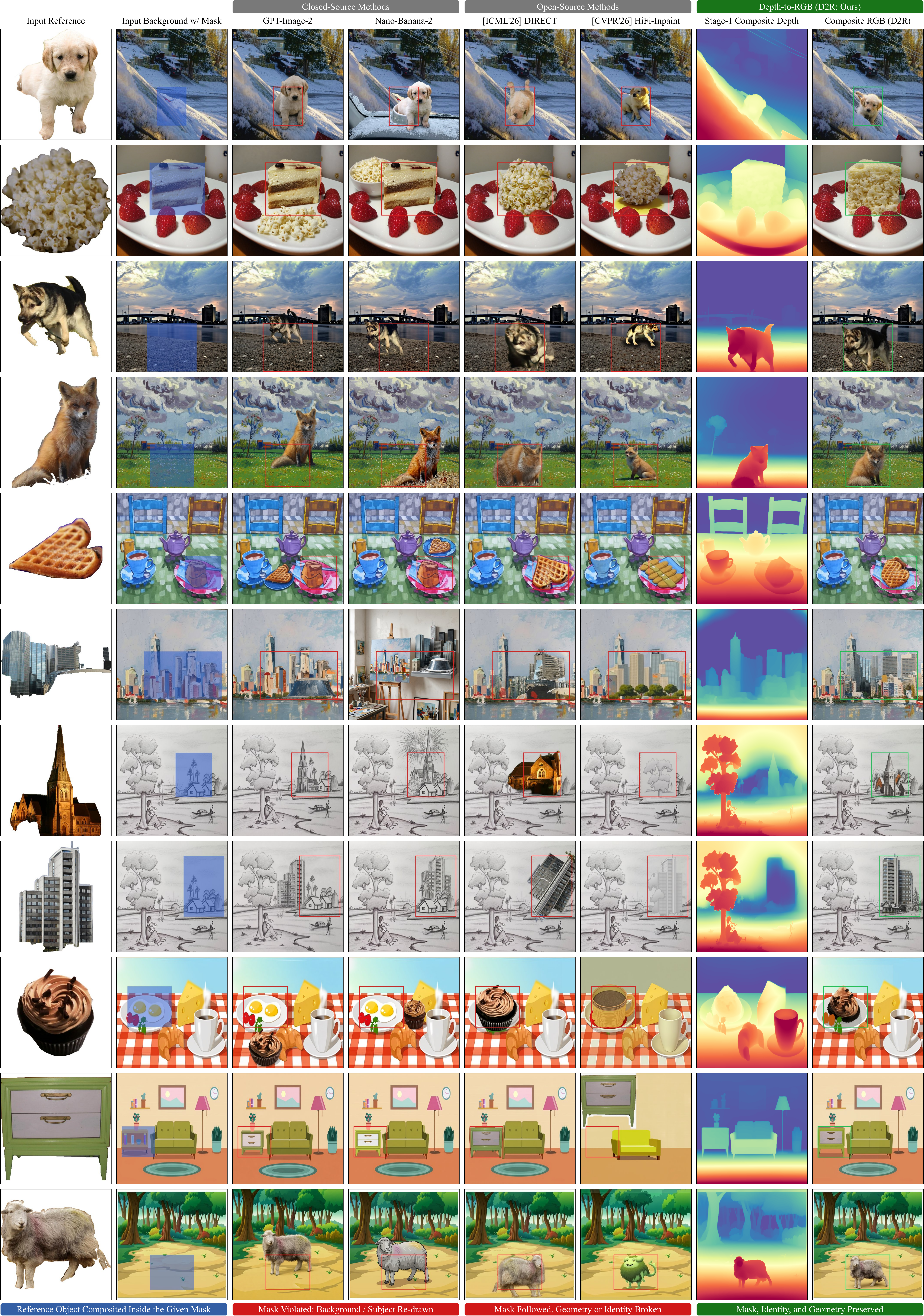}
\caption{Qualitative Results on TF-ICON~\citep{lu2023tficon}.}

\label{fig:qual_tficon}
\end{figure}

\begin{figure}[t]
\centering
\includegraphics[width=\textwidth]{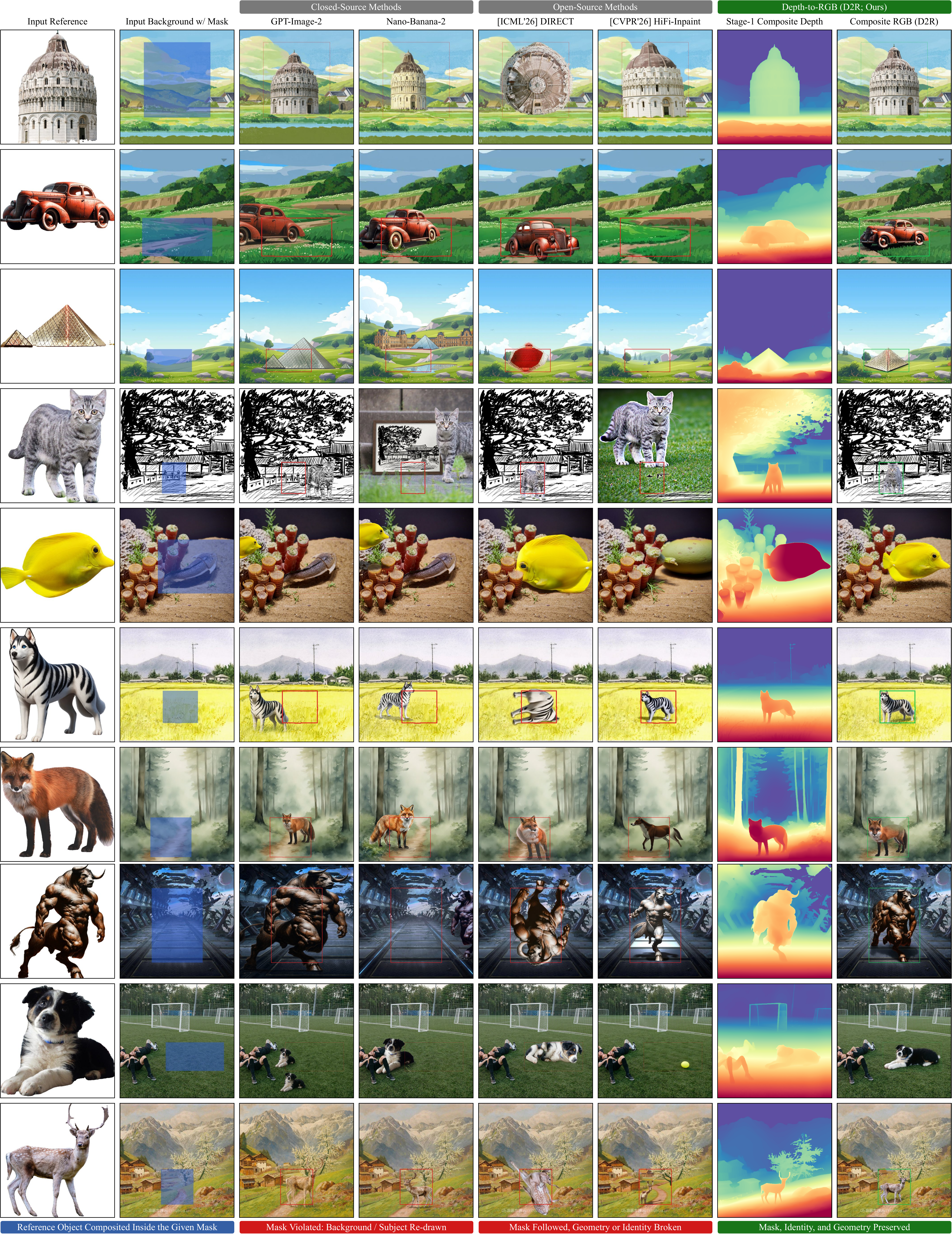}
\caption{Qualitative Results on AIComposer~\citep{li2025aicomposer}.}

\label{fig:qual_aicomposer}
\end{figure}

\begin{figure}[t]
\centering
\includegraphics[width=\textwidth]{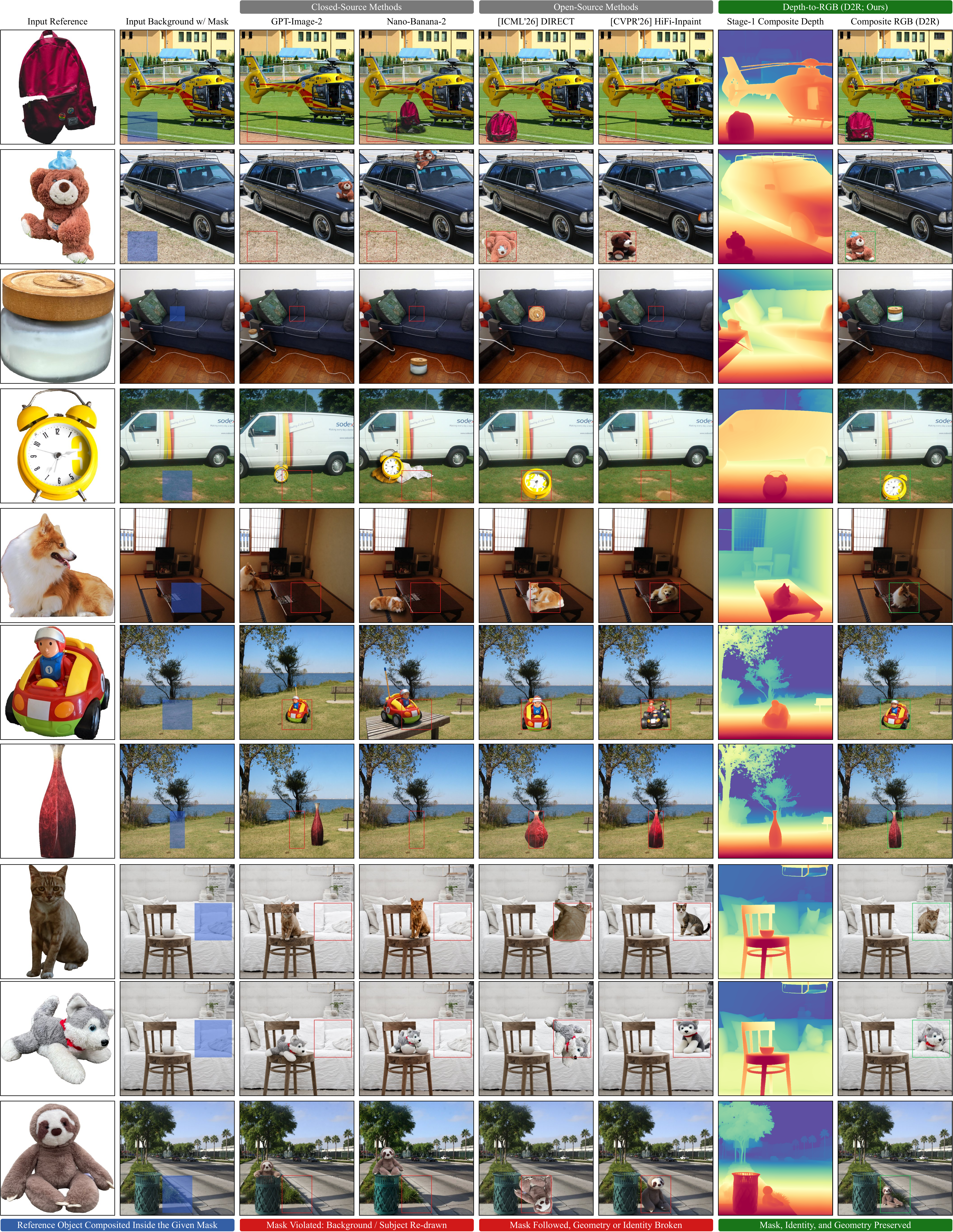}
\caption{Qualitative Results on ComplexCompo~\citep{lu2026shine}.}

\label{fig:qual_complexcompo}
\end{figure}

All displayed D2R examples use Stage-1 predicted depth. Figures~\ref{fig:qual_id_stage1}--\ref{fig:qual_complexcompo} follow Fig.~\ref{fig:teaser}: columns seven and eight show the predicted depth and resulting RGB. The OOD examples contain object tags absent from compositing training.

\clearpage

\clearpage
\noindent\textbf{Depth in Illustrated Scenes.} The bottom three rows of Fig.~\ref{fig:qual_tficon} depict cartoon breakfast-table, living-room, and forest scenes; the first four rows of Fig.~\ref{fig:qual_aicomposer} show illustrated landscapes and an ink drawing. Their depth and RGB columns show how predicted shape and placement translate into the composite. Appendix~\ref{sec:appx_limits} discusses the remaining harmonization errors.

\clearpage

\clearpage

\begin{figure}[!ht]
\centering
\includegraphics[width=\textwidth,height=0.78\textheight,keepaspectratio]{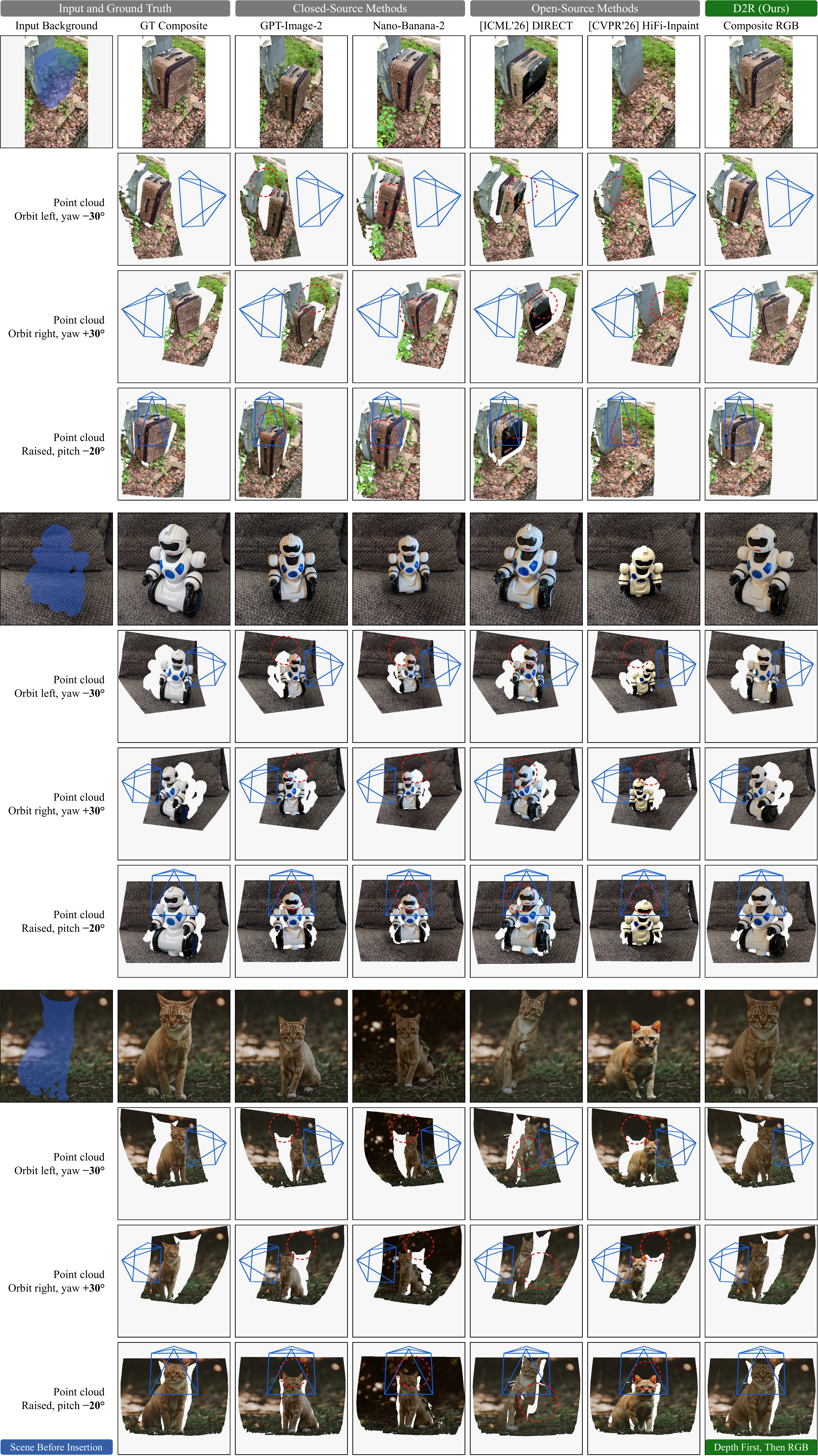}
\caption{Composites Viewed as Point Clouds.}

\label{fig:pointcloud}
\end{figure}

\clearpage
\noindent\textbf{Point-Cloud Views.} Figure~\ref{fig:pointcloud} shows each output back-projected with its own estimated depth and the background alignment in Eq.~(\ref{eq:align}). Off-axis views expose inconsistencies in these monocular predictions; they do not independently measure 3D geometry. Points crossing depth discontinuities are removed, and pale regions are disocclusions unsupported by the input view. The frustum marks the input camera. Dotted circles mark the largest inserted-object region with mean normalized disparity above $0.085$ relative to target depth, using the same rule for every method. No displayed D2R panel exceeds this criterion.

\end{document}